\newif\ificlr\iclrfalse

\documentclass{article}

\newcommand{\eg}{e.g., }

\newcommand{\ie}{i.e., }
\newcommand{\etc}{etc.}

\newcommand{\cf}{cf.\ }
\newcommand{\wrt}{w.r.t.\ }

\newcommand{\versus}{vs.\ }

\newcommand{\claude}{Claude~3.5~Sonnet}
\newcommand{\expflash}{Gemini~2.0~Flash~Experimental}
\newcommand{\flash}{Gemini~1.5~Flash}
\newcommand{\pro}{Gemini~1.5~Pro}
\newcommand{\gpt}{GPT-4o}
\newcommand{\mini}{o1-mini}
\newcommand{\preview}{o1-preview}
\newcommand{\oone}{o1}

\usepackage{amsmath}
\usepackage{amssymb}
\usepackage{amsthm}
\usepackage{booktabs}
\usepackage{caption}
\usepackage{float}
\usepackage{hyperref}
\usepackage{listings}
\usepackage{mathtools}
\usepackage{microtype}
\usepackage[newfloat,frozencache]{minted}
\usepackage{multicol}
\usepackage{multirow}
\usepackage{newfloat}
\usepackage{xcolor}
\usepackage{wrapfig}
\usepackage{numprint}
\usepackage{subcaption}
\usepackage{graphicx}
\usepackage[export]{adjustbox}

\usepackage[capitalize]{cleveref}

\usepackage[accepted]{icml2025}

\icmltitlerunning{LMAct: A Benchmark for In-Context Imitation Learning with Long Multimodal Demonstrations}

\begin{document}

    \twocolumn[
        \icmltitle{LMAct: A Benchmark for In-Context Imitation Learning\\with Long Multimodal Demonstrations}
        
        \begin{icmlauthorlist}
            \icmlauthor{Anian Ruoss}{gdm}
            \icmlauthor{Fabio Pardo}{gdm}
            \icmlauthor{Harris Chan}{gdm}
            \icmlauthor{Bonnie Li}{gdm}
            \icmlauthor{Volodymyr Mnih}{gdm}
            \icmlauthor{Tim Genewein}{gdm}
        \end{icmlauthorlist}
        
        \icmlaffiliation{gdm}{Google DeepMind}
        
        \icmlcorrespondingauthor{Anian Ruoss}{anianr@google.com}
        \icmlcorrespondingauthor{Tim Genewein}{timgen@google.com}
        
        \vskip 0.3in
    ]
    
    
    
    \printAffiliationsAndNotice{}  

    \begin{abstract}
        In this paper, we present a benchmark to pressure-test today's frontier models' multimodal decision-making capabilities in the very long-context regime (up to one million tokens) and investigate whether these models can learn from large numbers of expert demonstrations in their context.
We evaluate the performance of \claude{}, \flash{}, \pro{}, \expflash{}, \gpt{}, \mini{}, \preview{}, and \oone{} as policies across a battery of simple interactive decision-making tasks: playing tic-tac-toe, chess, and Atari, navigating grid worlds, solving crosswords, and controlling a simulated cheetah.
We study increasing amounts of expert demonstrations in the context --- from no demonstrations to 512 full episodes.
Across our tasks, models rarely manage to fully reach expert performance, and often, presenting more demonstrations has little effect.
Some models steadily improve with more demonstrations on a few tasks.
We investigate the effect of encoding observations as text or images and the impact of chain-of-thought prompting.
To help quantify the impact of other approaches and future innovations,
we open source our benchmark that covers the zero-, few-, and many-shot regimes in a unified evaluation.

    \end{abstract}
    
    \section{Introduction}
\label{sec:introduction}

The simple recipe of minimizing next-token prediction error at scale has led to large multimodal foundation models (LMs) with remarkably general capabilities~\citep{openai2023gpt4, anil2023gemini, anthropic2024claude3}.
Importantly, these capabilities come in two flavors: (i) the ability to produce outputs of high quality from a short and often underspecified prompt (\eg writing an essay about a novel topic), and (ii) the ability to learn new patterns and imitate algorithms \emph{in context}~\citep{reid2024gemini, mirchandani2023large}.
Both types of capabilities demonstrate that LMs
can manipulate learned knowledge and respond to new information in flexible and non-trivial ways.
These capabilities, both necessary for reasoning and decision-making, have led to the recent surge of using LMs as agents by sampling an action from the model~\citep{mirchandani2023large, dipalo2024keypoint}.

While LMs have been shown to be able to perform non-trivial reasoning and decision-making in some domains~\citep{ huang2022language,yao2023react,romera2024mathematical}, there are also many negative results where LMs fail to perform decision-making tasks that are very simple for humans, even when LMs arguably possess great factual knowledge of the task.
For example, LMs struggle to play legal moves, let alone beat amateur humans, in chess~\citep{carlini2023playing}.
At the same time, state-of-the-art LMs have detailed expert knowledge of chess when queried in natural language. But this \emph{declarative} knowledge fails to translate into effective decision-making, \ie ``know-how''~\citep{ryle1949concept}.
This ``knowing-doing gap'' is also observed in the recently released BALROG benchmark~\citep{paglieri2025balrog}, which evaluates zero-shot capabilities (\ie without expert demonstrations) of state-of-the-art LMs on interactive decision-making tasks in the long time horizon setting: 5 game environments from Baby AI~\citep{chevalier2019babyai} to Nethack~\citep{kuttler2020nethack}.
Overall, \citet{paglieri2025balrog} find that ``models struggle significantly with more challenging tasks'' and highlight few- and many-shot evaluation as an important potential solution.
Our benchmark (developed concurrently) addresses exactly this gap by covering the full range from zero-shot to many-shot evaluation.

\begin{figure*}[t]
    \begin{center}
        \includegraphics[width=0.8\textwidth]{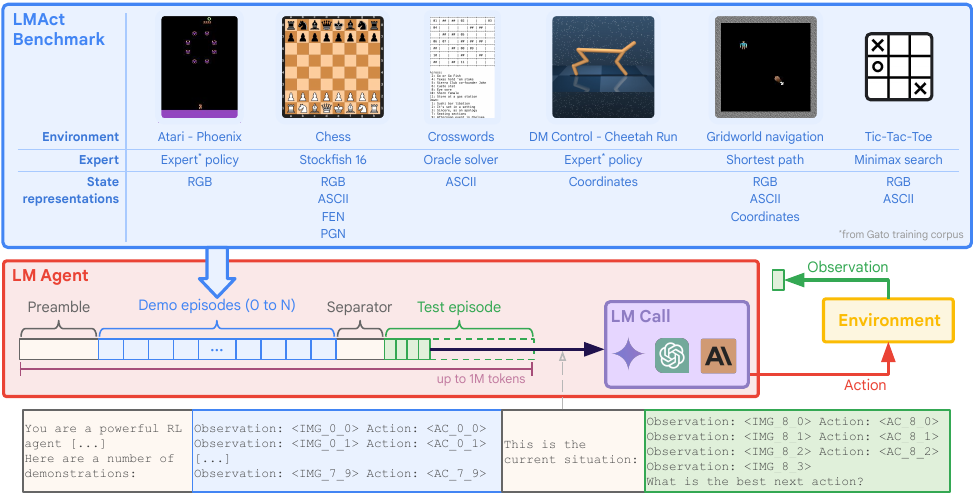}
    \end{center}
    \vspace{-8pt}
    \caption{
        LMAct overview.
        Our multimodal benchmark consists of six decision-making tasks that come with an expert policy and potentially multiple state representations.
        For evaluation, LM performance is measured on test episodes with unseen initial states.
        LMs are conditioned on a generic decision-making preamble (fixed across all tasks), followed by $0$ to $N$ demonstration episodes, and a separator that indicates the start of the current episode ($N$ can be up to $512$, with up to $100$ steps per episode, both depending on the task. The maximum context length is $1$M tokens).
        In each step of the test episode an action is generated by the LM's predicted continuation of the context.
        \ificlr
        The resulting environment interaction produces the next observation that is added to the previous state action pairs.
        \else
        The resulting environment interaction produces the next observation that is added to the growing context of state action pairs.
        \fi
    }
    \vspace{-4pt}
    \label{fig:overview}
\end{figure*}

The general question arising from these results is whether LMs in principle have the capabilities to solve interactive decision-making tasks but misunderstand the ``out-of-distribution'' specification given by the zero-shot prompt or a few examples, or whether the problem is more deeply rooted.
Accordingly, our paper focuses on whether conditioning on a \emph{large} number of expert demonstrations (state-action trajectories) helps unlock the decision-making capability of pretrained LMs.
In doing so, we test the multimodal in-context learning capabilities of modern LMs at their limits, with contexts that are up to a million tokens long (with thousands of output tokens).
Our main research question is:
\begin{quote}
    Can state-of-the-art LMs learn to act in dynamic environments by generalizing from \emph{large} numbers of \emph{multimodal} in-context expert demonstrations?
\end{quote}

\paragraph{Main Contributions}

See \Cref{fig:overview} for our tasks and methodology.
\ificlr
Our key contributions are:
\else
We make the following key contributions:
\fi

\begin{itemize}
    \item We conduct a comprehensive empirical evaluation of the multimodal in-context imitation learning capabilities of state-of-the-art LMs (\claude, \flash, \pro, \expflash, \gpt, \mini, \preview, and \oone) on a battery of interactive, potentially long-horizon, tasks: playing tic-tac-toe, chess, and Atari, navigating grid worlds, solving crosswords, and a DM~Control task.
    \item We show that --- even when optimizing the prompt (number of demonstrations, chain-of-thought prompting, \etc) for each model and task --- frontier LMs fail to reach expert performance on Atari, chess, and DM Control.
    Some models approach expert performance on crosswords, grid world, and tic-tac-toe.
    All models beat the random action baselines except on Atari.
    \item We vary the number of expert demonstration episodes in the context from $0$ up to $512$ (the limit depends on the model and the task) and find that performance is mostly independent of the number of demonstrations. In some cases we observe strong in-context learning.
    \item We run a control experiment where LLM agents need to replay the single demonstration episode in the context, where all models except for \mini{} perform well.
    \item We open-source \ificlr
    LMAct,
    \else\fi our in-context imitation learning benchmark that covers the zero-, few-, and many-shot regime in a unified manner, including all expert demonstrations and evaluation code at
    \url{https://github.com/google-deepmind/lm_act}.
\end{itemize}

    \section{Methods}
\label{sec:methods}

\ificlr
We now describe the models we evaluate~(\cref{sec:methods:models}), our benchmark environments~(\cref{sec:methods:environments}), our prompt~(\cref{sec:methods:prompt}), and our evaluation protocol~(\cref{sec:methods:evaluation-protocol}).
Full details in \cref{app:experimental-details}.
\else
We now briefly describe the models we evaluate~(\cref{sec:methods:models}), our benchmark environments~(\cref{sec:methods:environments}), how we construct the prompt~(\cref{sec:methods:prompt}), and our evaluation protocol~(\cref{sec:methods:evaluation-protocol}).
Full details are given in \cref{app:experimental-details}.
\fi

\subsection{Models}
\label{sec:methods:models}

We evaluate the current (closed-weights) frontier LMs:
\claude{} (the 2024-10-22 version)~\citep{anthropic2024claude3.5, anthropic2024introducing},
\flash{} and \pro{} (the $002$ versions)~\citep{reid2024gemini},
\expflash{}~\citep{google2024introducing},
\gpt{} (the 2024-08-06 version)~\citep{openai2024gpt4o,openai2024gpt4ocard},
\mini{} and \preview{}~\citep{openai2024o1}, and 
\oone{}~\citep{openai2024o1card}.
Except for \mini{} and \preview{}, all models process multimodal prompts, though their exact specifications differ (see \cref{app:experimental-details:models}).
We use temperature $0$ for all models (except for \mini{}, \preview{}, and \oone{}, which have a fixed temperature of $1$~\citep{openai2024reasoning}).
We set the maximum (output) sample length to $2048$ tokens for all models (except for \mini{}, \preview{}, and \oone{}), which is more than sufficient to achieve strong performance (see our ablation in \cref{fig:sample-length-ablation}).
In contrast, the performance of \mini{}, \preview{}, and \oone{} crucially depends on the number of ``reasoning tokens'' (see \cref{fig:sample-length-ablation}), so we use a maximum (output) sample length of $8192$ tokens as a good compromise between cost and performance ($2048$ tokens would lead to severe performance degradation on our tasks -- see our ablation in \cref{app:additional-results:max-sample-length-ablation}).
We post-process the model outputs by removing all the leading/trailing white spaces and only consider the text after the keyword ``Action:'', \ie we discard all (chain-of-thought) reasoning traces.

\subsection{Environments}
\label{sec:methods:environments}

We consider a battery of well-known interactive decision-making environments: the Phoenix game from Atari $2600$~\citep{bellemare2013arcade}, chess, crosswords, the cheetah run task from the DM Control suite~\citep{tassa2018deepmind}, grid world navigation, and tic-tac-toe.
We briefly describe each environment below (full details in \cref{app:experimental-details:environments}).
\ificlr
Since sampling is deterministic for most models (\ie the temperature is $0$, see \cref{sec:methods:models}), we introduce variability in the demonstration and evaluation episodes by varying the initial conditions (\eg by using different openings for chess; see below). 
\else
Since sampling is deterministic for most models (\ie the temperature is $0$, see \cref{sec:methods:models}), we introduce variability in the demonstration and evaluation episodes by varying the initial conditions (\eg via the environment seed for DM Control or by using different openings for chess; see below). 
\fi

\paragraph{Atari -- Phoenix}

We chose Phoenix as a representative Atari task that has somewhat dense rewards, which is important since we can only evaluate $400$ frames (with action repeat $4$, \ie $100$ steps), or roughly $6$ seconds of play due to context size limitations.
Phoenix also forms part of the set of $5$ games that is highly predictive of the performance on the full Atari suite~\citep{aitchison2023atari}.
We use the Arcade Learning Environment~\citep{bellemare2013arcade} version, and for expert demonstrations we use the Gato training data~\citep{reed2022generalist}.
We use the original (\ie not downsampled or grayscale) images as observations (see \cref{fig:observations-atari}).

\paragraph{Chess}

We evaluate models against the \emph{weakest-possible} version of Stockfish~$16$~\citep{romstad2008stockfish}, \ie level $0$ (which corresponds to an Elo of approx.\ $1320$).
We further weaken this version by evaluating only 1 node.
We generate expert demonstrations with the strongest (default) version of Stockfish, \ie level $20$ with a time limit of 50ms per move as the agent (note that the opponent, \ie the ``environment'', remains fixed as the weakest version of Stockfish).
We evaluate four different state representations (see \cref{fig:observations-chess}): (i) a 2D ASCII encoding of the board, (ii) the Forsyth–Edwards Notation (FEN), which encodes the board and very limited historical information as a string, (iii) the Portable Game Notation (PGN), a plain text format for recording chess games via the move history given in algebraic chess notation, and (iv) an RGB image of the board.
We always represent actions via the algebraic chess notation.
To ensure variability in the demonstration and evaluation episodes, we use the openings from the Encyclopedia of Chess Openings~\citep{matanovic1978encyclopaedia}, which we randomly sample without replacement.
\ificlr
We play for at most $100$ steps (terminating early for a win/draw/loss; the average is $38$) and assign reward $1$ to a win, $0$ to a draw, $-1$ to a loss, and $0$ else.
\else
We play all games for at most $100$ steps (terminating early in case of a win/draw/loss; the average number of steps per game is $38$) and assign a reward of $1$ to a win, $0$ to a draw, $-1$ to a loss, and $0$ to all other states.
\fi

\paragraph{Crossword}

We create a large collection of $7\times7$ crosswords using the genxword crossword generator~\citep{whitlock2011genxword} and a list of \numprint{55189} clues with the \emph{lowest difficulty rating} collected by Matthew Ginsberg (individual clues may appear in multiple crosswords with low probability).
Each episode is a distinct crossword represented as an ASCII crossword grid followed by two lists of clues, one for the ``Across'' words and one for the ``Down'' words (see \cref{fig:observations-crossword}).
A valid action consists of either ``A'' (for across) or ``D'' (for down), followed by the word's index and the word itself.
We assign a reward of $1$ to a correct word, $0$ to an incorrect word with correct length or to a correct word that has been already been placed, and we terminate the episode with a reward of $-1$ for an incorrect word.
We evaluate $25$ steps (which is sufficient to place the approx.\ $10$ words per crossword on average) and terminate early if all words are filled.
We generate the expert demonstrations by simply outputting the solution for each clue one by one in random order.

\paragraph{DM Control -- Cheetah Run}

\ificlr
We evaluate Cheetah Run from DM Control~\citep{tassa2018deepmind}.
\else
We use the Cheetah Run task from the DM Control Suite~\citep{tassa2018deepmind}.
\fi
We represent observations as string in the style of a Python dictionary of position and velocity vectors with individual values between $-1$ and $1$ (see \cref{fig:observations-dm-control}).
Each episode begins in a new, randomly initialized state.
We only evaluate the first $100$ steps of an episode (due to context length limits).
\ificlr
We use the Gato training data~\citep{reed2022generalist} to create expert demonstrations (details in \cref{app:experimental-details:environments}).
\else
As for Atari, we use the Gato training data~\citep{reed2022generalist} to create expert demonstrations (details in \cref{app:experimental-details:environments}).
\fi

\paragraph{Grid World}

We consider a $12\times12$ grid world with walls, effectively yielding a $10\times10$ grid, with a single player and a single target (no obstacles).
Actions are up, down, left, and right, and since the grid is fully observable, memorization is not required.
\ificlr
We evaluate three different state representations (see \cref{fig:observations-grid-world}): (i) 2D ASCII of the grid, (ii) the player and target coordinate tuples, and (iii) the RGB image of the grid.
\else
We evaluate three different state representations (see \cref{fig:observations-grid-world}): (i) an ASCII encoding of the 2D grid, (ii) the player and target coordinate tuples provided as plain text, and (iii) the RGB image of the grid.
\fi
The reward is $1$ if the player reaches the target and $0$ otherwise.
Episodes run for a maximum of $25$ steps (reaching any target from any initial position in a $10\times10$ grid requires at most $18$ steps).
We randomly initialize the player/target locations for every episode.
The expert demonstrations correspond to one of the shortest paths between the player and the target.

\paragraph{Tic-Tac-Toe}

We play tic-tac-toe against a \emph{random} policy that picks an empty spot uniformly at random.
For demonstration episodes the expert uses minimax search (\ie it plays an optimal action), while the environment remains fixed as a random policy.
\ificlr
We evaluate two state representations (see \cref{fig:observations-tic-tac-toe}): (i) 2D ASCII of the grid, and (ii) the RGB image of the grid.
\else
We evaluate two state representations (see \cref{fig:observations-tic-tac-toe}): (i) an ASCII encoding of the 2D grid, and (ii) the RGB image of the grid.
\fi
To ensure variability, each game starts from an opening state (similar to chess openings), which we draw uniformly across all initial game states (as a result, even the optimal minimax strategy can only win $85\%$ of the games and draws the rest).
We assign a reward of $1$ to a win, $0$ to a draw, $-1$ to a loss, and $0$ else.

\subsection{Prompt}
\label{sec:methods:prompt}

Our prompt consists of two main parts: (i) the expert demonstration episodes (\cref{code:frozen-prompt}), and (ii) the trajectory of the evaluation episode (including the episode's previous actions and environment states; see \cref{code:dynamic-prompt}).
We use a generic decision-making zero-shot preamble at the beginning of (i), and a short separator prompt at the beginning of (ii) (see \cref{fig:overview,code:frozen-prompt,code:dynamic-prompt}).
The expert demonstrations are fixed across an evaluation episode, but resampled across episodes, while the current trajectory starts with a single initial state and grows as more state-action pairs are observed (up to a maximum of $100$ steps).
We do not include an environment task description (\eg the rules of tic-tac-toe) in the preamble.
We may, however, show the available legal actions, which depend on the environment, in each step at the end of (ii) (see \cref{code:dynamic-prompt}), depending on whether it is beneficial per model and task (see our ablations in \cref{app:additional-results:ablations}).
Similarly, we may use a chain-of-thought~\citep{wei2022chain} style prompt at the very end of (ii) asking the model to provide a reasoning before proposing an action (see \cref{code:dynamic-prompt}), again, depending on whether it increases performance in the ablations (\cref{app:additional-results:ablations}).
We make both of these decisions per model and task, \ie the same model may use chain-of-thought for one task but not another.

\subsection{Evaluation Protocol}
\label{sec:methods:evaluation-protocol}

In every evaluation step, we condition the model on the current context, upon which it generates an action that we feed to the environment.
We then concatenate the resulting next environment state (prepended with the action that produced it) to the growing context.
We ablated whether to show the model's previous actions in the evaluation trajectory (see \cref{app:additional-results:past-actions-ablation}) and found that it mostly improves performance, so we always include them.
If a model fails to generate a legal action, we uniformly sample one of the legal actions (we visualize the percentage of illegal actions in \cref{app:additional-results:illegal-actions}).
Since the models have different maximum context lengths (and different text/image tokenizers), the maximum number of demonstration episodes that fit into the context depends on the model, the task, and the state representation format.
\ificlr
For example, for a given model, we can only use $16$ demonstration episodes with RGB observations but up to $256$ ASCII demonstration episodes.
\else
For example, for a given model, we may only by able to use $16$ demonstration episodes with RGB observations but up to $256$ ASCII demonstration episodes.
\fi

We always evaluate $100$ episodes with different initial conditions (each episode is evaluated individually) and report the average score. 
The maximum episode duration is $100$~steps and the score per episode is the cumulative reward over all steps (we never show reward information to the models). 
For each evaluation episode we uniformly subsample (without replacement) the demonstration episodes (for the frozen part of the prompt) from a precomputed pool of up to $1000$ distinct demonstrations.
We ensure that all evaluation episodes have initial states that differ from the demonstration episodes (except for the replay control experiments in \cref{app:additional-results:replay}).
\ificlr
We minimally postprocess (see \cref{sec:methods:models}) the LM generations to extract the action and reject incorrectly formatted but semantically correct actions, as matching the action format is important for our imitation learning benchmark.
\else
We only perform very minimal postprocessing (see \cref{sec:methods:models}) of the LM generations to obain the action and (deliberately) reject semantically correct actions that are wrongly formatted, as matching the action format is an important aspect of our imitation learning benchmark.
\fi

    \section{Results}
\label{sec:results}

We now present our comprehensive empirical evaluation of the (closed-source) frontier models on our benchmark for interactive decision-making with long multimodal context.
We investigate how performance changes when presenting more demonstration episodes in the context (see \cref{sec:methods:environments} for a task overview and \cref{app:experimental-details:environments} for details and illustrations).
For each model and task, we first ablate whether to use chain-of-thought prompting and whether to show the legal actions in the prompt (results in \cref{app:additional-results:ablations}).
We use one demonstration episode (\ie still many individual demonstration steps) for these ablations, as a compromise between lower computational demands and being representative of the in-context learning setting.
Due to the very low monthly rate limits of the Anthropic API~\citep{anthropic2024rate}, we do not sweep over the number of demonstrations for \claude{} and only run the ablation.
We also do not evaluate it on Atari since it can only process $100$ images at once (a single demonstration episode hits that limit).

\begin{figure*}[t]
    \begin{center}
        \includegraphics[width=0.85\textwidth]{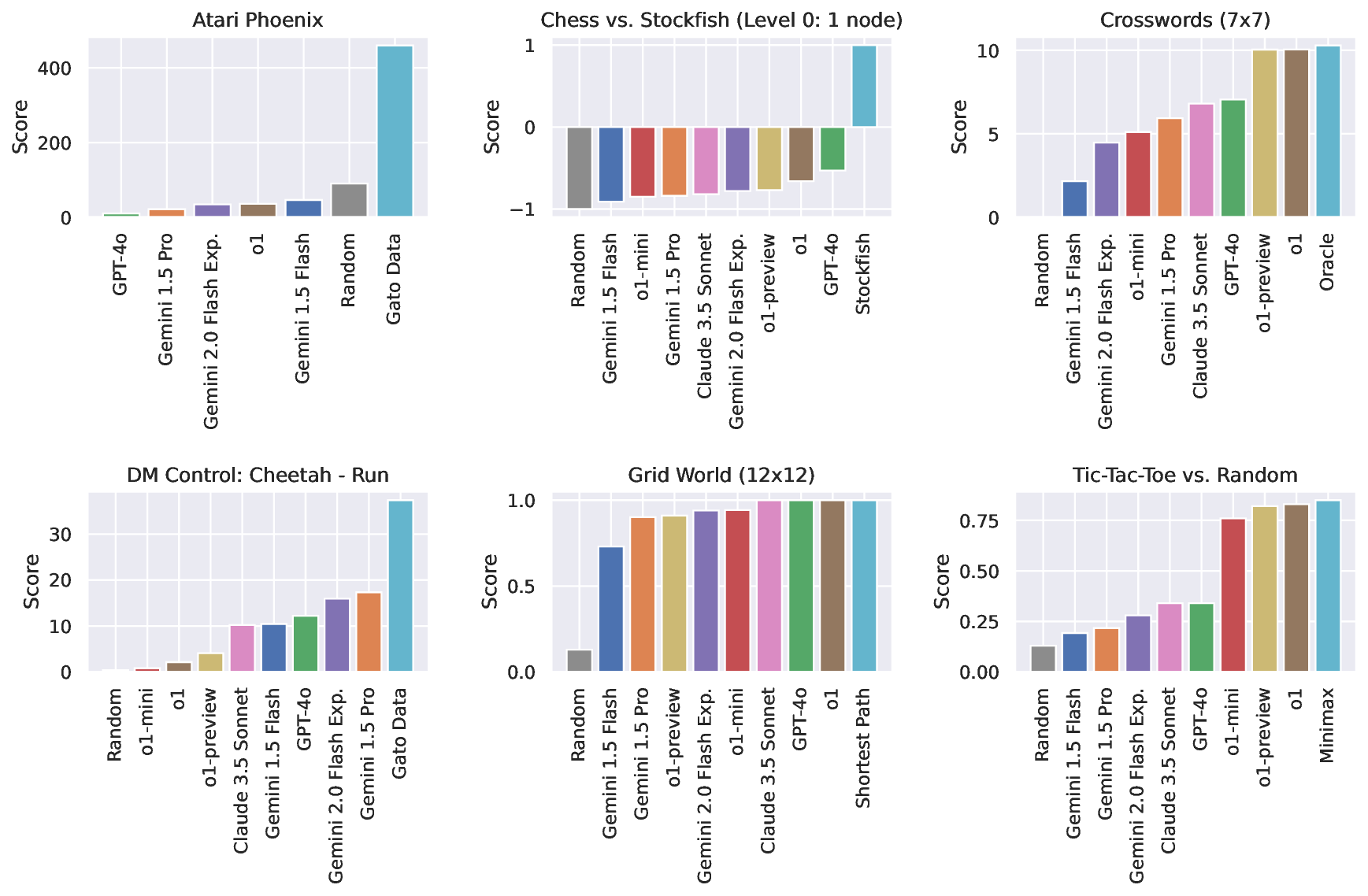}
    \end{center}
    \vspace{-12pt}
    \caption{
        Best scores per model and task across all observation formats, numbers of demonstration episodes, and ablations (chain-of-thought, showing legal actions).
        Accordingly, different bars in a panel may be based on different settings.
        The expert policy (which produced the demonstrations) is an upper baseline.
        The lower baseline randomly selects a legal action at each step.
        \claude{}, \mini{}, and \preview{} cannot be evaluated on Atari -- Phoenix because they cannot process (enough, for \claude{}) images. 
    }
    \label{fig:best-score}
    \vspace{-4pt}
\end{figure*}

\subsection{Best Scores Per Model/Task}
\label{sec:results:best-scores}

\cref{fig:best-score} shows the highest overall score per task and model across all settings (number of demonstrations, observation format, showing legal actions, chain-of-thought prompting).
Different models thus may use different settings on the same task, and the same model may use different settings across tasks. 
In \cref{sec:results:in-context-imitation-learning}, we will keep the observation format and number of demonstrations constant across all models per data point.
As stated above, results for \claude{} are from the ablation (\ie $1$ demonstration episode).
\cref{fig:best-score} shows that models often struggle to match expert scores --- even in their best setting.
Exceptions are grid worlds, which most models largely solve, tic-tac-toe, where the o1 models achieve near-optimal score (against a random opponent), and crosswords, which \preview{} and \oone{} almost solve.
On the other hand, all models always outperform the random action baseline except for Atari -- Phoenix.
LMs tend to repeat actions (see \cref{fig:atari-subsequence-length} for a detailed analysis), and in Phoenix holding down the firing button without releasing results in a single shot (no auto-fire).
In contrast, random actions produce a fair amount of (successful) shots.

\subsection{In-Context Imitation Learning}
\label{sec:results:in-context-imitation-learning}

We now investigate the in-context imitation learning capabilities of today's frontier models, by varying the number of demonstration episodes from zero-, to few-, and many-shot evaluations.
The largest number of demonstrations depends on the models' context sizes, the task, and the tasks' observation formats (images require more tokens than text).
We omit \claude{} due to its low monthly token limit.  

\paragraph{Atari -- Phoenix}

\ificlr
\begin{wrapfigure}{r}{0.5\textwidth}
    \vspace{-20pt}
\else
\begin{figure}[th!]
\fi
    \begin{center}
        \includegraphics[width=0.5\textwidth]{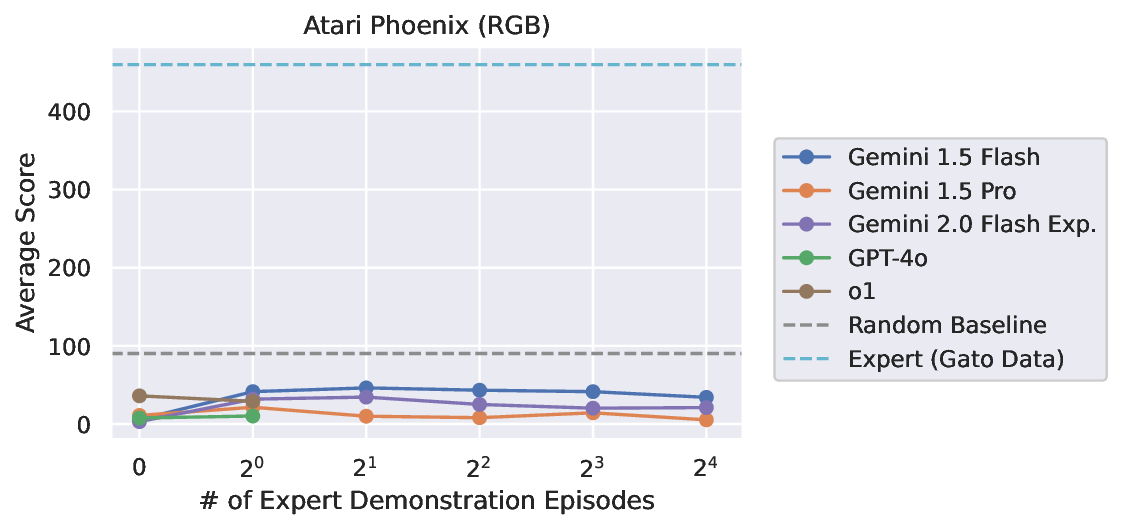}
    \end{center}
    \vspace{-12pt}
    \caption{
        In-context imitation learning on Atari -- Phoenix (RGB observations).
        Almost all models benefit (mildly) from one demonstration episode, but not from more (\gpt{} and \oone{} cannot fit multiple demonstration episodes in the context).
        While no model outperforms the random baseline, \flash{} performs best.
    }
    \label{fig:atari-phoenix}
    \vspace{-4pt}
\ificlr
\end{wrapfigure}
\else
\end{figure}
\fi

\cref{fig:atari-phoenix} shows that on Atari -- Phoenix all models except \oone{} improve slightly by having one demonstration episode, as opposed none.
However, more demonstrations (only possible for the Gemini models) do not improve the performance further.
All models struggle to outperform the random action baseline because of their tendency to repeat actions (and thus fire very little, since Phoenix has no auto-fire).
\cref{tab:atari-ablation} contains the ablation results.
Models rarely output illegal actions (see \cref{fig:atari-illegal-actions}), \ie they reliably (re)produce the correct action format.
Atari -- Phoenix is arguably the hardest task in our benchmark, since it only comes with image observations (no text representation), and has very large demands \wrt context size (high frame rate, somewhat high resolution images, etc.). 
Despite LMs' massive scale, playing Atari games well (at least by naively feeding raw images to the context) seems currently beyond reach, both in terms of capabilities, but also \wrt context size and compute demands (\cf \citet{waytowich2024atari}).

\ificlr
\begin{figure}
\else
\begin{figure}[th!]
\fi
    \begin{subfigure}[t]{\ificlr0.5\textwidth\else\columnwidth\fi}
        \begin{center}
            \includegraphics[width=\textwidth]{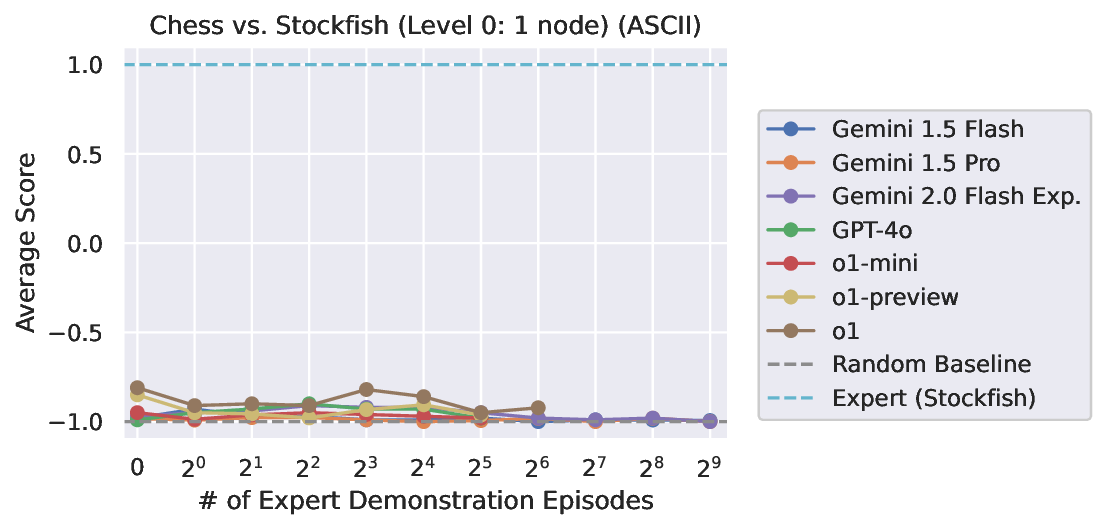}
        \end{center}
        \vspace{-0.25cm}
        \caption{ASCII observations}
    \end{subfigure}
    \begin{subfigure}[t]{\ificlr0.5\textwidth\else\columnwidth\fi}
        \begin{center}
            \includegraphics[width=\textwidth]{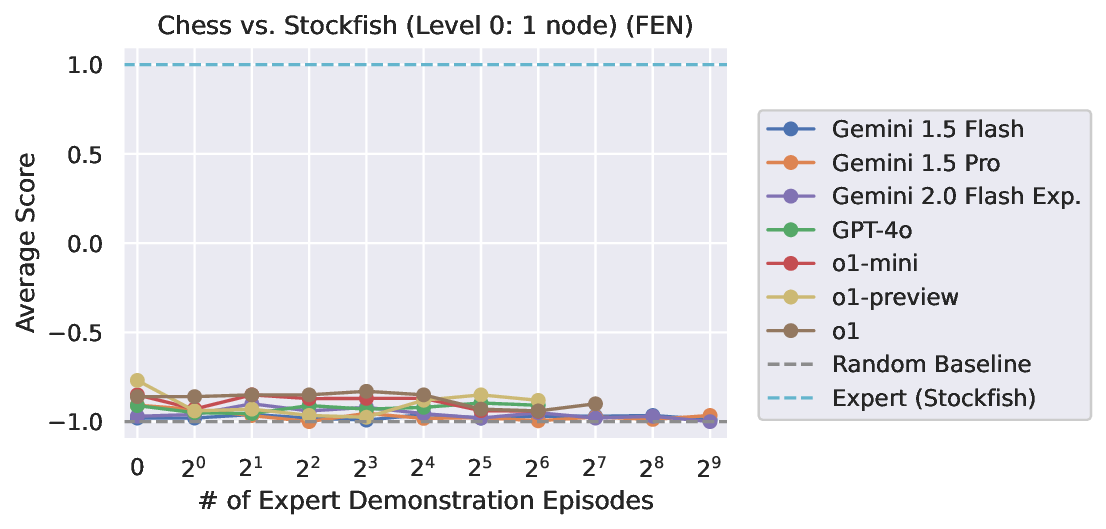}
        \end{center}
        \vspace{-0.25cm}
        \caption{FEN observations}
    \end{subfigure}
    \begin{subfigure}[t]{\ificlr0.5\textwidth\else\columnwidth\fi}
        \begin{center}
            \includegraphics[width=\textwidth]{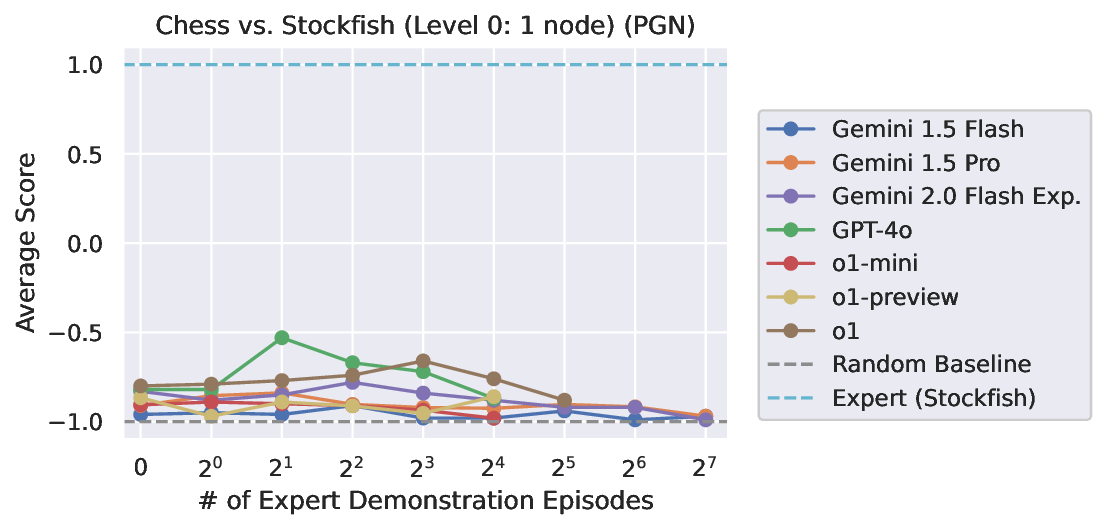}
        \end{center}
        \vspace{-0.25cm}
        \caption{PGN observations}
    \end{subfigure}
    \begin{subfigure}[t]{\ificlr0.5\textwidth\else\columnwidth\fi}
        \begin{center}
            \includegraphics[width=\textwidth]{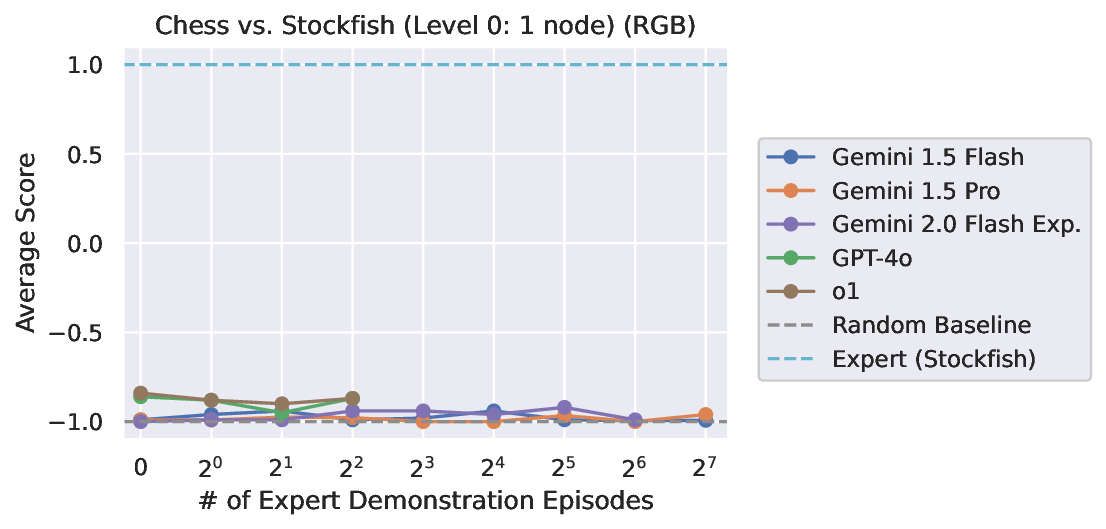}
        \end{center}
        \vspace{-0.25cm}
        \caption{RGB observations}
    \end{subfigure}
    \caption{
        In-context imitation learning on chess against the weakest variant of Stockfish (level $0$, $\approx1300$ Elo), further restricted to one node.
        The models almost always lose (\ie score $-1$) and do not benefit from more demonstrations.
        The PGN observations enable the best results, in particular for \gpt{}, which performs best but still loses majority of games against this weak opponent.
    }
    \vspace{-4pt}
    \label{fig:chess}
\end{figure}

\paragraph{Crosswords}

\ificlr
\begin{wrapfigure}{r}{0.5\textwidth}
    \vspace{-16pt}
\else
\begin{figure}[t]
\fi
   \begin{center}
        \includegraphics[width=0.5\textwidth]{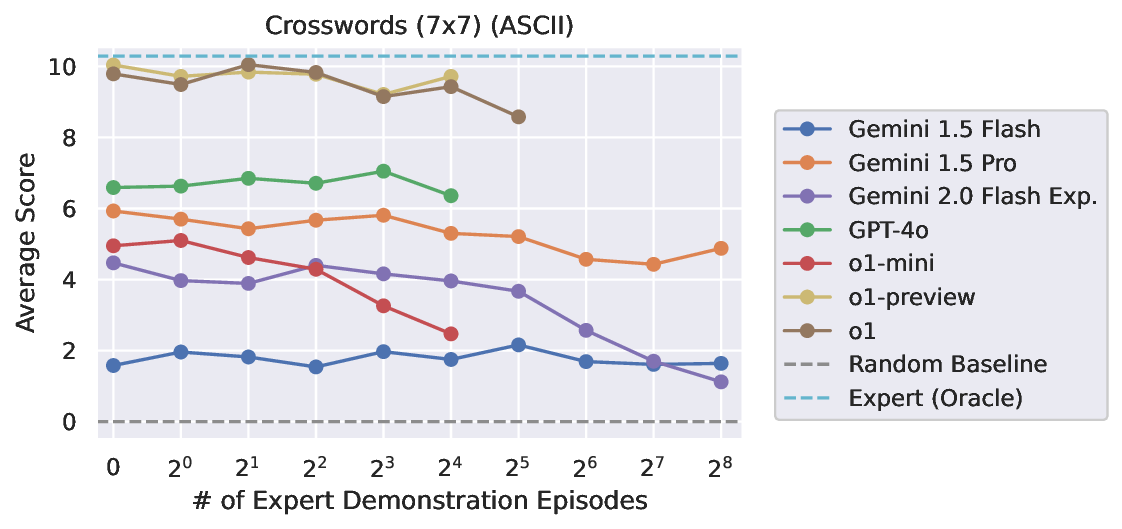}
    \end{center}
    \vspace{-8pt}
    \caption{
        In-context imitation learning on $7\times7$ crossword puzzles (using clues with the simplest rating) with ASCII observations.
        The performance of most models is largely unaffected by the number of expert demonstration episodes.
        \preview{} and \oone{} solve most crosswords, while other models struggle to varying degrees.
    }
    \label{fig:crossword}
    \vspace{-4pt}
\ificlr
\end{wrapfigure}
\else
\end{figure}
\fi

\cref{fig:crossword} shows LMs' performance on crosswords with simple clues (as rated by Matthew Ginsberg).
While the models achieve very different overall scores, their individual performance is largely independent of the number of demonstrations (except for \mini{} and \expflash{}, which degrade with more demonstrations).
Overall, \preview{} and \oone{} perform best, almost completely solving all the (simple) crosswords.
\cref{fig:crossword-illegal-actions} shows that the number of illegal actions (\ie where models either suggest a word of incorrect length or fail to respect the ``Across'' \versus ``Down'' format) is quite high for all models and roughly inversely proportional to their puzzle-solving competence.
We use chain-of-thought prompting for some but not all models (see our ablation results in \cref{tab:crossword-ablation}), but we never show the legal actions (which would be a unreasonably long list of words with correct length).

\paragraph{Chess}

\cref{fig:chess} shows LMs' chess-playing performance against the weakest version of Stockfish 16~\citep{romstad2008stockfish}
(\ie level $0$, $\approx1300$ Elo), further restricted to only evaluating a single node.
We investigate four observation formats: ASCII, FEN, PGN, and RGB (\mini{} and \preview{} cannot process images).
Showing more demonstrations has little effect on performance, and models rarely manage to beat (score $1$) or draw (score $0$) against this very weak opponent.
The results show that playing chess (without any scaffolding or fine-tuning) is still out of reach for current LMs.
\cref{fig:chess-illegal-actions} reveals that the models often output illegal actions (which also does not improve with more demonstrations) --- even though the legal actions are provided in the prompt (\cf the ablations in \cref{tab:chess-ablation-anthropic,tab:chess-ablation-gemini,tab:chess-ablation-openai}).

\paragraph{Tic-Tac-Toe}

\begin{figure}
    \begin{subfigure}[t]{0.5\textwidth}
        \begin{center}
            \includegraphics[width=\textwidth]{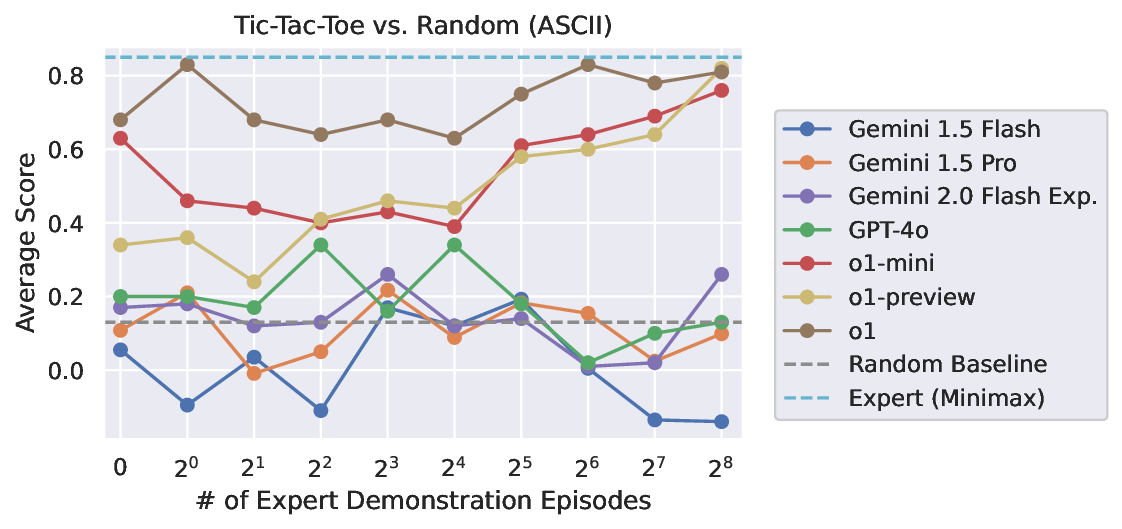}
        \end{center}
        \vspace{-0.25cm}
        \caption{ASCII observations}
    \end{subfigure}
    \begin{subfigure}[t]{0.5\textwidth}
        \begin{center}
            \includegraphics[width=\textwidth]{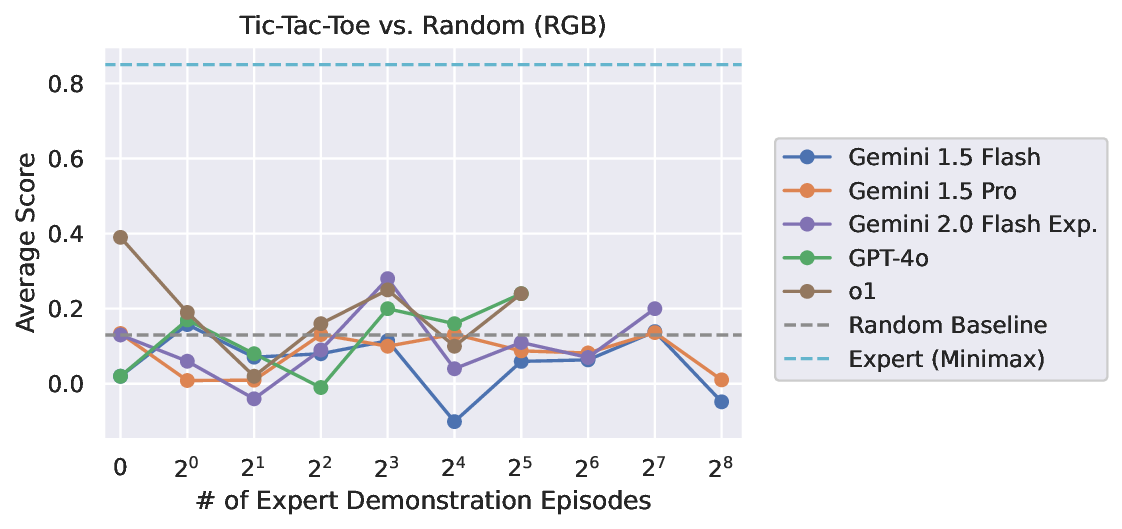}
        \end{center}
        \vspace{-0.25cm}
        \caption{RGB observations}
    \end{subfigure}
    \caption{
        In-context imitation learning on tic-tac-toe against a random action adversary.
        Apart from the o1 models on ASCII observations, all models struggle to play better than a random baseline.
        \mini{} and \preview{} improve with more demonstrations and both reach expert performance at $256$ demonstration episodes.
    }
    \vspace{-4pt}
    \label{fig:tic-tac-toe}
\end{figure}

\cref{fig:tic-tac-toe} shows the results for playing tic-tac-toe against a random-action opponent.
Since the demonstration and evaluation episodes start from different initial states, episodes begin with partially filled boards, which cannot always be won (\ie optimal score $<1$).
The o1 models reach expert performance with ASCII observations and show signs of in-context learning.
The other models struggle to outperform the random action baseline (as does \oone{} with RGB observations).
\cref{fig:tic-tac-toe-illegal-actions} shows that the models (apart from \pro{} on RGB) generate few illegal actions, implying that the models' weak performance is caused by outputting suboptimal rather than illegal actions.

\paragraph{Grid World}

\ificlr
\begin{wrapfigure}{r}{0.5\textwidth}
\vspace{-8pt}
\else
\begin{figure}[th!]
\fi
    \begin{subfigure}[t]{0.5\textwidth}
        \begin{center}
            \includegraphics[width=\textwidth]{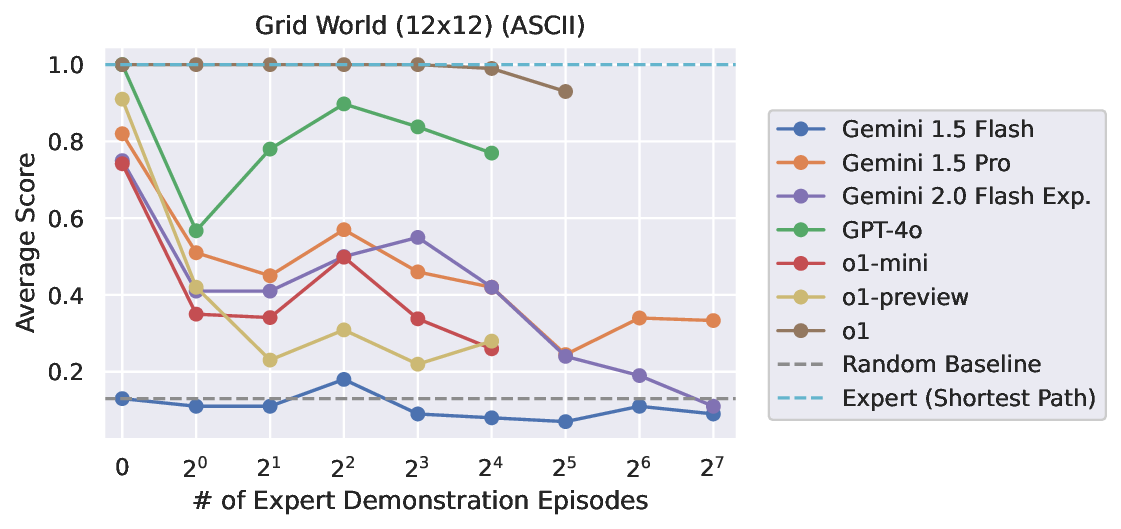}
        \end{center}
        \vspace{-0.25cm}
        \caption{ASCII observations}
    \end{subfigure}
    \begin{subfigure}[t]{0.5\textwidth}
        \begin{center}
            \includegraphics[width=\textwidth]{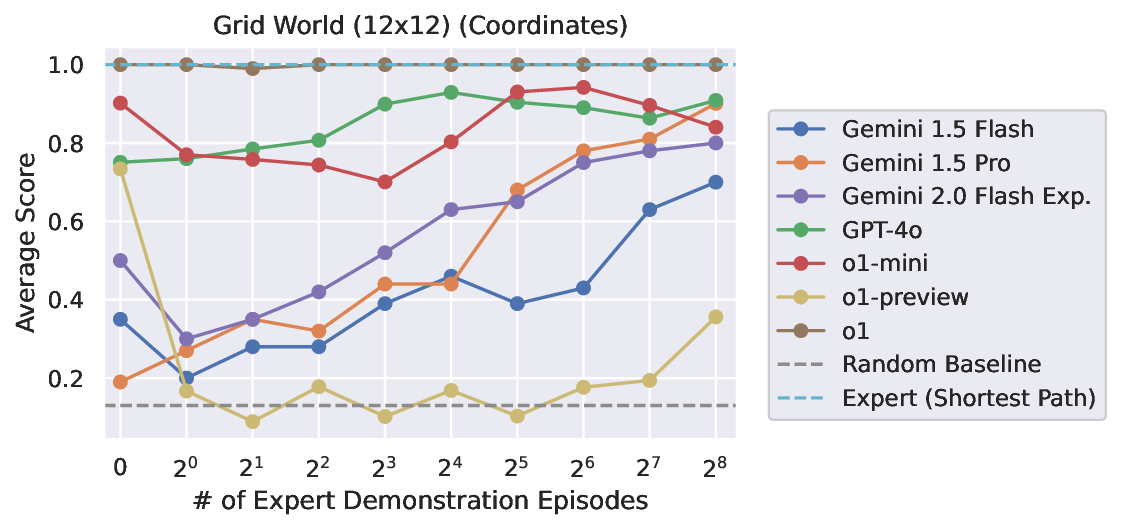}
        \end{center}
        \vspace{-0.25cm}
        \caption{Coordinate observations}
    \end{subfigure}
    \begin{subfigure}[t]{0.5\textwidth}
        \begin{center}
            \includegraphics[width=\textwidth]{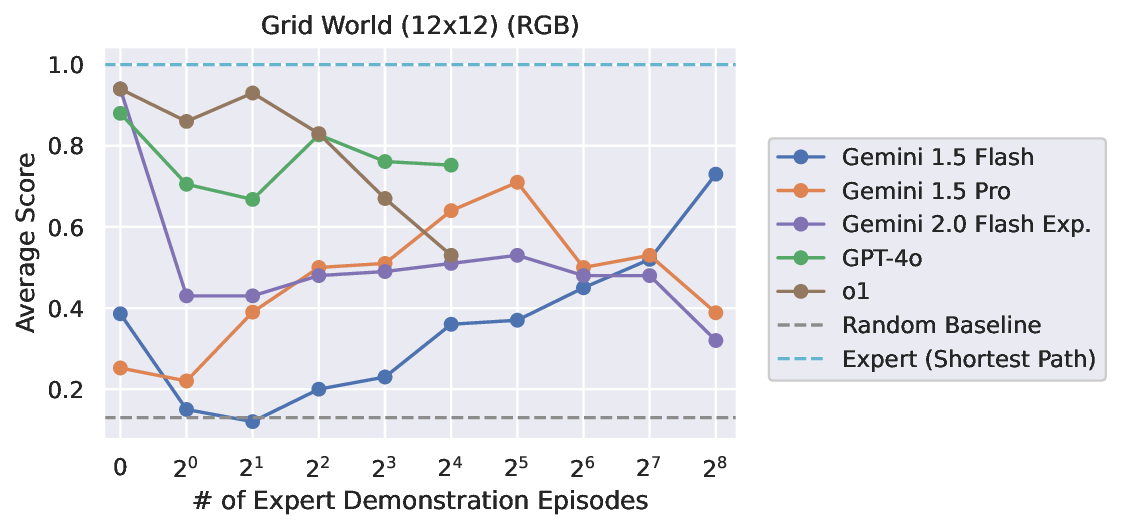}
        \end{center}
        \vspace{-0.25cm}
        \caption{RGB observations}
    \end{subfigure}
    \caption{
        In-context imitation learning for navigating to a target in a $12\times12$ grid world (see \cref{fig:observations-grid-world}) using the commands: up, down, left, right.
        On ASCII, most models deteriorate with more demonstrations.
        For coordinate observations (player and target tuple) and RGB images, Gemini improves with more demonstrations (except \expflash{} on RGB), indicating in-context learning across a very long context.
        \gpt{} (and \oone{} on RGB) shows no such a trend but already achieves high zero- and few-shot performance (\oone{} is near-perfect on ASCII and coordinates).
    }
    \label{fig:grid-world}
    \ificlr
    \vspace{-16pt}
    \else
    \vspace{-4pt}
    \fi
\ificlr
\end{wrapfigure}
\else
\end{figure}
\fi

\cref{fig:grid-world} shows how LMs perform on the task of navigating a simple grid world.
The Gemini models steadily improve with more demonstrations for coordinate and RGB image (1.5 models only) observations, demonstrating strong in-context learning with very long contexts.
While \gpt{}, \mini{}, and \preview{} do not benefit from more demonstrations, \oone{} performs very well in almost all settings.
On ASCII observations, most models (except \oone{} and \flash{}) deteriorate with more demonstrations.
\cref{fig:grid-world-illegal-actions} shows the illegal actions for each model, revealing that \preview{}'s performance drops on ASCII and coordinate observations strongly correlate with the number of illegal actions.
All other models rarely generate illegal actions after one demonstration episode.
Overall, grid world is the easiest task in our benchmark where all models perform quite well (some almost optimally) with the right combination of state representation and number of demonstrations.

\paragraph{DM Control -- Cheetah Run}

\cref{fig:dm-control} shows LMs' ability to control a simulated (half) cheetah from DM~Control~\citep{tassa2018deepmind}.
We encode observations and actions as strings (see \cref{fig:observations-dm-control}).
For all models, except \mini{}, a single demonstration episode is helpful but more than two demonstrations degrade performance.
The o1 models struggle to significantly outperform the random action baseline.
\pro{} and \expflash{} achieve the highest score, reaching roughly half the expert score.
Except for \mini{}, all models mostly generate legal actions (see \cref{fig:dm-control-illegal-actions}) --- even without any demonstration episodes (the action format can be inferred from the past actions in the evaluation trajectory since we randomly sample a legal action if the model generates an illegal one).
Therefore, the poor performance in the zero-shot regime (\ie $0$ demonstration episodes) cannot be explained by the models' potential lack of knowledge of the action format.
Instead, models (with the exception of \mini{}) manage to learn a non-trivial policy from one or two demonstration episodes (but fail to leverage more episodes to further improve performance).

    \section{Discussion \& Related Work}
\label{sec:discussion}

Many large-scale benchmarks have been developed to test the general capabilities of frontier LMs, including the Chatbot Arena~\citep{chiang2024chatbot} or LiveBench~\citep{white2025livebench}.
Other benchmarks, such as FrontierMath~\citep{glazer2024frontiermath}, TaskBench~\citep{shen2024taskbench}, GameBench~\citep{costarelli2024gamebench}, and Atari-GPT~\citep{waytowich2024atari}, specifically investigate the reasoning or interactive decision-making capabilities of (L)LMs.
Most closely related to our work, and published in parallel, the BALROG~\citep{paglieri2025balrog} benchmark evaluates frontier LMs' zero-shot reasoning on multimodal decision-making tasks in five game environments from Baby AI to NetHack.
Our work complements these benchmarks by focusing on in-context imitation learning with long multimodal context, filling a gap in the current literature. 
Our benchmark is currently in its ``easiest'' form (\eg grid world without obstacles) and can easily be made more challenging.

\ificlr
\begin{wrapfigure}{r}{0.5\textwidth}
    \vspace{-20pt}
\else
\begin{figure}
\fi
   \begin{center}
        \includegraphics[width=0.5\textwidth]{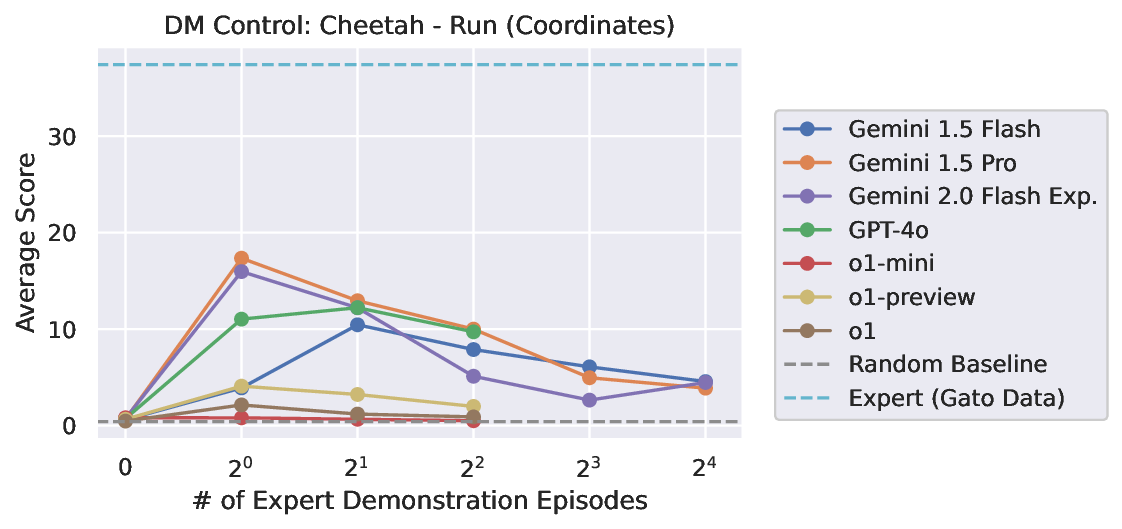}
    \end{center}
    \vspace{-8pt}
    \caption{
        In-context imitation learning on the cheetah run task from the DM Control suite using position/velocity vector observations encoded as text.
        \pro{} and \expflash{} perform best with a single demonstration episode, \gpt{} and \flash{} with two demonstration episodes, with more demonstrations degrading performance.
        \pro{} achieves the highest performance, roughly half the expert score.
    }
    \label{fig:dm-control}
\ificlr
\end{wrapfigure}
\else
\end{figure}
\fi
There are a few reasons to be optimistic about in-context imitation learning with long-context models.
Many demonstrations should, in principle, improve performance over few-shot learning (as shown by \citet{agarwal2024many, jiang2024many} on non-interactive tasks), and modern LMs have long enough contexts to test this at scale.
Additionally, pretrained LMs have a fairly general ability to recognize and imitate algorithmic patterns in their context \citep{mirchandani2023large}, which is in line with the memory-based meta-learning view on (universal) in-context prediction~\citep{ortega2019meta, grau2024learning}.
If universal enough, models would, with enough observations, recognize and correctly continue environment-agent interaction patterns.
\ificlr
We refer interested readers to \cref{app:related-work} for a more detailed discussion of additional related work.
\else
See \cref{app:related-work} for a discussion of additional related work.
\fi

While we find some cases of steady in-context learning, in the majority of cases LMs' performance is largely independent of the number of demonstrations. 
It is unclear whether in-context imitation learning is not well suited to communicate desired agentic behavior (\ie models do not ``understand'' the task purely from demonstrations), or whether models have difficulty to effectively use a dense long context.
To provide additional insight we perform a ``replay'' control experiment, where a single episode is shown in context, and the same exact episode is evaluated, such that models only need to ``copy'' actions from the demonstration episode.
We find that models (except \mini{}) perform well in this control experiment on all tasks (see \cref{app:additional-results:replay}).

\vspace{-8pt}
\paragraph{Limitations}

We perform an evaluation via closed-source APIs and thus have little control over how the data is processed and fed to the underlying models.
Since models behind the APIs can be updated at any time, it is possible that our results may not be quantitatively reproducible soon after publishing this manuscript.
Despite our best efforts in evaluating different prompt formats, we cannot rule out that even small changes to the prompt could lead to better results.
Accordingly, our current results are a lower bound on the models' performance.
We can also not guarantee that our observation formats do not cause tokenization issues across all APIs (\eg a loss of structure for a 2D grid in ASCII).
Finally, while our environments are simpler than complex real-world scenarios like robotics, they require the same fundamental agentic capabilities (long-context multimodal understanding, in-context imitation learning) that are often tested in robotics research and likely necessary for real-world success.
Accordingly, our LMAct benchmark serves as a controlled testbed to evaluate these core skills, diagnose current model limitations, and guide research toward more generally capable agents.
Although direct transfer is beyond this paper's scope, findings on our benchmark can inform future work addressing that challenge.

\vspace{-8pt}
\paragraph{Future Work}

For Cheetah Run (\cref{fig:dm-control}) and grid world navigation with ASCII demonstrations (\cref{fig:grid-world}) we observe that the models' performance deteriorates with increasing numbers of expert demonstrations, a phenomenon we refer to as ``in-context interference''.
For both tasks our analysis of the percentage of illegal actions (\cref{fig:dm-control-illegal-actions,fig:grid-world-illegal-actions}) suggests the problem is not due to an increased number of illegal actions, but, instead, due to increasingly suboptimal actions.
We have conducted a set of initial investigations into these failure modes (see \cref{app:additional-results:replay,app:additional-results:illegal-actions,app:additional-results:ablations,fig:atari-subsequence-length}), but 
a definitive answer to what causes in-context interference would require a thorough investigation, and, therefore, presents an interesting direction for future work.
Another promising avenue for future work is to focus on models capable of general in-context reinforcement learning, which is a bit different than our in-context imitation setting (in principle, all our tasks could easily be extended by providing additional reward observations). 
It also seems plausible that pretraining or finetuning with data from interactive decision-making tasks, and, in particular, in-context imitation of an expert policy, would be quite effective.
Finally, we also think that evaluating LMs' performance on partially observable tasks presents an interesting direction for future research (we primarily investigate fully observable tasks). 
Partially observable tasks require consistent integration of information across several, potentially non-adjacent, time steps, which is a great test of a model’s ability to densely attend to the information in the context.
However, under partial observability, there is a theoretical problem of self-delusion in imitation of an expert that has hidden information (\eg the hidden belief state of the agent; see \citet{ortega2021shaking}).
We want to keep our benchmark free from these complicating issues and think that the right time to move to such harder tasks is when frontier models easily solve simple, fully observable tasks at expert level (we are currently at beginner level).
Nevertheless, we believe there is great value in benchmarks on interactive decision-making tasks under partial observability (perhaps better suited for in-context reinforcement learning, which avoids the self-delusion problem), and refer to, \eg the BALROG benchmark~\citep{paglieri2025balrog}.

    \section{Conclusion}
\label{sec:conclusion}

We evaluated the multimodal in-context imitation learning capabilities of some of the world's most advanced AI models on interactive decision-making tasks --- tasks that are simple for humans but challenging for state-of-the-art LMs.
Our results show that, even with hundreds of demonstration episodes, context lengths of up to one million tokens, and thousands of output tokens, models often struggle to reach expert performance, thereby failing to translate their (factual) knowledge about the tasks' solution strategies into effective decision-making.
Solving this problem will be crucial for the next generational leap in LM capabilities towards general agents.
Despite our focus on in-context imitation learning, we believe it will be interesting to compare against other methods, including fine-tuning, retrieval-based methods, reward-conditioning, etc.
Our open-source benchmark (\url{https://github.com/google-deepmind/lm_act}) serves as a yard stick to measure progress toward that goal, and we are excited to see which innovations will be needed to solve all our simple tasks.

    \section*{Impact Statement}

The (partial) automation of intellectual labor will have significant societal and socioeconomic impact, both positive and negative.
One current hurdle is the automation of general interactive decision-making tasks, as our benchmark demonstrates.
If solved, this would lead to more autonomous AI systems and agents that could facilitate or even partly take over intellectual labor across a broad range of applications and domains.
We believe this vision warrants both enthusiasm and caution, and the scientific community and developers of AI technology bear responsibility in raising awareness of potential negative socioeconomic outcomes, and in helping develop potential mitigations. 
Having said that, our work is a benchmark that reports the current state of affairs (showing the limitations of current frontier LMs) and serves as one yard stick to measure future innovations and progress against, but does not propose innovations to improve capabilities of AI systems.
While our benchmark may contribute indirectly by helping develop more capable AI systems faster, we believe this is far outweighed by the benefit of having a clearer picture of current capabilities and progress provided by our benchmark.

    \section*{Acknowledgments}

We thank
Anna Mitenkova,
Dipanjan Das,
Jordi Grau-Moya,
Joel Veness,
Kate Olszewska,
Li Kevin Wenliang,
Laurent Orseau,
Marcus Hutter,
Matthew Aitchison,
Matthew Ginsberg,
Minmin Chen,
Orhan Firat,
Satinder Baveja,
Shaobo Hou,
Zhengdong Wang,
and the anonymous reviewers
for their helpful feedback and insightful discussions.

    \bibliography{references}
    \bibliographystyle{icml2025}
    
    \clearpage
    
    \appendix
    \onecolumn
    
    \setcounter{figure}{0}
    \renewcommand{\thefigure}{A\arabic{figure}}
    \setcounter{table}{0}
    \renewcommand{\thetable}{A\arabic{table}}
    
    \section{Related Work}
\label{app:related-work}

The emergence of strong in-context learning capabilities with increasing model and training data scale in LLMs was first observed in the GPT-2 paper~\citep{radford2019language} and even gave the title to the GPT-3 paper~\citep{brown2020language}.
Soon after, the recipe of next-token prediction at scale was applied to build agents from large sequential predictors, resulting, \eg in the decision (pretrained) transformer~\citep{chen2021decision, lee2023supervised}, or the generalist agent GATO~\citep{reed2022generalist}, and its more recent open source variant JAT~\citep{gallouedec2024jack}.
While the empirical results were certainly surprising at the time, at least in theory, in-context learning must necessarily arise as a core feature of a sequential predictor trained to minimize next token log loss over an implicit meta distribution of data~\citep{ortega2019meta, mikulik2020meta, genewein2023memory}, and could, in principle, even lead to universal in-context predictors~\citep{grau2024learning}.
An explicit application of this memory-based meta-learning principle at scale is the ``adaptive agent'' from \citet{bauer2023human}, which shows that an embodied agent in a 3D environment can adapt to novel task instances on human timescales (\ie single- or low double digit numbers of interaction episodes) purely in context and across a vast set of tasks.
The \citet{sima2024scaling} conducted another impressive large-scale application of training and fine-tuning a complex vision-language agent across a large set of environments, using instruction conditioning together with in-context learning.

Instead of pretraining separate models for general decision-making, many researchers have also attempted to directly use the knowledge and reasoning capabilities of LLMs and VLMs for building agents that interact with an environment (see \citet{xi2025rise} for a 2023 survey of LLM-based agents) --- either by fine-tuning~\citep{li2022pretrained} or purely via in-context demonstrations~\citep{dipalo2024keypoint}.
In both cases, two open question are: (i) how to best represent environment observations as tokens, and (ii) how to best elicit the decision-making capabilities of pretrained LMs.
\citet{mirchandani2023large} find that pretrained LLMs are ``general pattern machines'' that can learn to complete complex token sequences via in-context learning, including agent-environment interactions.
They even find that in many cases, randomly swapping the alphabet does not have significant impact, suggesting that LLMs may be able to deal with many different ways of translating observations into tokens.
Perhaps more important is the question whether observations should be state-action sequences (as in imitation learning and our work) or whether they should include rewards (for reward conditioning or in-context RL as in \citet{mirchandani2023large} and \citet{raparthy2024generalization}).
While this is still unclear for pretrained LMs, \citet{ruoss2024amortized} find that, when training a large transformer to play chess, performance is roughly the same for imitation learning compared to learning to predict state or action values (as long as the amount of data for all three variants is equal).
\citet{schultz2025mastering} extend these results on chess to LLMs by distilling the search proceedure into the model by linearizing the search trees.
\citet{ma2025vision} find that, when prompted appropriately, a pretrained VLM (\pro{}) can produce good value estimates for real-world robotic tasks.
The second open problem, how to best prompt LMs for decision-making, is a very active research area~\citep{li2025why,genewein2025understanding}, with a lot of focus on designing or learning zero-shot prompts, such as the famous ``Let's think about this step by step.''~\citep{kojima2022large} and ``Take a deep breath and work on this problem step-by-step.''~\citep{yang2024large}, which has led to many chain-of-thought prompting schemes~\citep{wei2022chain}.
Besides better zero-shot prompts, advanced sampling and prompt optimization schemes have been explored,
such as iterative prompt refinement where an agent starts with some demonstrations in the context, then interacts with the environment, and potentially replaces a lower performing episode with a higher performing one~\citep{mirchandani2023large, brooks2023large}. 

With the recent availability of long context models, a third possibility compared to improving zero-shot or few-shot prompts has emerged: many-shot in-context learning, \ie prompting with many demonstrations, on the order of having a full small dataset in the context.
Both \citet{agarwal2024many} and \citet{jiang2024many} show that many-shot prompts (hundreds or thousands of examples in the prompt) improve pretrained LM performance over few-shot prompts in non-interactive tasks.
Our paper explores the same direction, with potentially hundreds of full \emph{episodes} in the context, for interactive decision-making tasks.
At the time of writing, querying long context models with many tokens comes with high computational cost, but it would be interesting to investigate how specialized models that were specifically developed for long-context tasks~\citep{bulatov2022recurrent, cherepanov2024recurrent} would fare on our benchmark.
An alternative could be to virtually extend the context via retrieval based methods, \eg REGENT~\citep{sridhar2024regent}, which trains a retrieval based agent.
The retrieval problem is currently not fully solved (as also pointed out in \citet{paglieri2025balrog}), and today's largest state-of-the-art LMs do not offer retrieval via their APIs and/or have not been trained to perform retrieval.
Accordingly, placing all the data in the context, as in our work, allows estimating LM performance independent of whether retrieval works well.
Our results thus serve as an upper baseline to calibrate retrieval based methods against.

Another line of work has investigated in-context learning for control~\citep{duan2017oneshot, xu2022prompting, fu2024incontext, fang2024moka, yin2024incontext, wang2024prompt}.
While these methods focus on developing novel agents with strong in-context learning capabilities, our paper's primary goal is to benchmark the current state of existing, general-purpose frontier LMs on such tasks.
Understanding how to best leverage insights from specialized agent research to enhance these large, pre-trained models remains an open question, which underscores the importance of establishing clear benchmarks like LMAct.

Comparing capabilities of LMs that span many tasks and domains requires large-scale benchmarks, such as the Chatbot Arena (a.k.a. LMSys, \citet{chiang2024chatbot}) or LiveBench~\citep{white2025livebench}.
Other benchmarks, such as FrontierMath~\citep{glazer2024frontiermath}, Taskbench~\citep{shen2024taskbench}, Gamebench~\citep{costarelli2024gamebench}, and BABILong~\citep{kuratov2024babilong}, specifically investigate the reasoning or interactive decision-making capabilities of LLMs.
Closely related to these and to our work, and published in parallel to our work, the BALROG~\citep{paglieri2025balrog} benchmark evaluates state-of-the-art LMs (the same as in our work with the exception of \mini{} and \preview{} and the addition of the Llama 3 models, which we do not evaluate) on multimodal reasoning and decision-making tasks by using a set of 5 increasingly harder game environments from Baby AI to Nethack.
Like our work, the authors find that state-of-the-art LMs struggle significantly in challenging game environments.
The authors also make the observation that models arguably possess a lot of knowledge about their tasks when queried appropriately, but that state-of-the-art LMs have a ``knowing-doing'' gap.
Unlike our work, the BALROG paper performs zero-shot evaluation and lists few-shot and many-shot evaluations as an important open research question (the released codebase supports few-shot evaluations, but these evaluations have not been performed at the time of writing).
Similar to BALROG, \citet{waytowich2024atari} evaluate the zero-shot game-playing capabilities of frontier LLMs (GPT-4V Turbo, \gpt, \flash{}, and Claude~$3$~Haiku) on $8$ different Atari games (but not the Phoenix game that we consider).
Compared to these previous benchmarks, ours is the only one that puts the emphasis on imitation learning with long context in multimodal interactive environments --- a regime that pushes against the limits of modern LMs' in-context learning and reasoning capabilities, both from an engineering and a capabilities perspective.
Furthermore, our benchmark covers the zero-shot, few-shot, and many-shot setting in a unified and thus easily comparable evaluation.

    \section{Experimental Details}
\label{app:experimental-details}

In this section, we provide additional details on our experimental setup.

\subsection{Models}
\label{app:experimental-details:models}

\paragraph{\claude{}}

\claude{} has a maximum context of $200$k tokens, but can only process $100$ images at a time. In our setting, a single episode can already consist of up to $100$ images.
As a result, we can, \eg not evaluate in-context learning for \claude{} on Atari.
\claude{} uses approximately $(\mathrm{width~px} \cdot \mathrm{height~px}) / 750$ tokens per image~\citep{anthropic2024vision}.
Finally, note that the monthly token limits for \claude{} are very low compared to the other models (\$$5000$ spend limit per month~\citep{anthropic2024rate}), which is why we can only conduct a somewhat limited evaluation of this model.

\paragraph{\flash{}, \pro{}, and \expflash{}}

The Gemini~$1.5$ models have a maximum context window of $10$ million tokens.
In practice, we restrict ourselves to $1$ million tokens due to the prohibitive cost of evaluating even longer prompts.
The Gemini API considers images to be of fixed size, meaning that they consume a fixed number of tokens (currently $258$), regardless of their file size~\citep{google2024understand}.
\expflash{} supports up to $1$M input tokens and $8$k output tokens~\citep{google2024flash}.

\paragraph{\gpt{}}

The \gpt{} model has a context length of $128$k tokens and can process up to $250$ images per prompt~\citep{openai2024models}.
We use the ``auto'' resolution setting to process the images, which automatically determines whether to use a ``low'' or ``high'' resolution based on the image size~\citep{openai2024vision}.
In the ``low'' resolution mode, the model represents an image with $85$ tokens.
In the ``high'' resolution mode, the model first consumes the low-resolution image (using $85$ tokens) and then creates additional detailed crops of the image using $170$ tokens per $512\mathrm{px} \cdot 512\mathrm{px}$ tile to cover the whole image.

\paragraph{\mini{}, \preview{}, and \oone{}}

\mini{} and \preview{} have a context window of $128$k tokens with up to $65$k output tokens but cannot process images~\citep{openai2024reasoning} so we only evaluate them in the text-based environments (\eg not on Atari).
The \oone{} model has a context window of $200$k tokens and up to $100$k output tokens.

\subsection{Environments}
\label{app:experimental-details:environments}

\paragraph{Visualizing State Representations}

\begin{figure}[t]
    \begin{minipage}{0.33\textwidth}
        \begin{center}
            \includegraphics[width=0.5\textwidth]{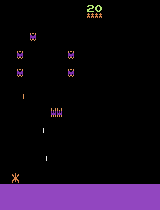}
        \end{center}
        \captionof{figure}{An RGB observation for the Phoenix game for Atari 2600.}
        \label{fig:observations-atari}
    \end{minipage}
    \hfill
    \begin{minipage}{0.33\textwidth}
        \begin{center}
            \subfloat[ASCII]{\input{figures/observations/tic_tac_toe_ascii}}
            \hfill
            \subfloat[RGB]{
                \includegraphics[width=0.35\textwidth,valign=t]{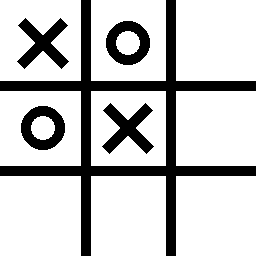}
            }
        \end{center}
        \caption{Sample observations for each state representation format from our tic-tac-toe environment.}
        \label{fig:observations-tic-tac-toe}
    \end{minipage}
\end{figure}

\begin{figure}
    \begin{center}
        \subfloat[ASCII]{
            \adjustbox{valign=t}{
                \resizebox{0.25\textwidth}{!}{\begin{tabular}{@{}cccccccc@{}}
    r & n & b & q & k & b & n & r \\[1ex]
    p & p & p & p & p & p & p & p \\[1ex]
    . & . & . & . & . & . & . & . \\[1ex]
    . & . & . & . & . & . & . & . \\[1ex]
    . & . & . & . & . & . & . & . \\[1ex]
    . & . & . & . & . & . & . & . \\[1ex]
    P & P & P & P & P & P & P & P \\[1ex]
    R & N & B & Q & K & B & N & R \\[1ex]
\end{tabular}
}
            }
            \label{fig:observation-chess-ascii}
        }
        \hfill
        \subfloat[PGN]{
            \input{figures/observations/chess_pgn}
            \label{fig:observation-chess-pgn}
        }
        \hfill
        \subfloat[RGB]{
            \includegraphics[width=0.25\linewidth,valign=t]{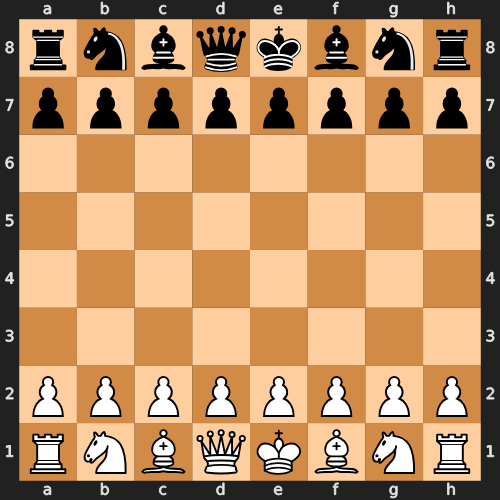}
            \label{fig:observation-chess-rgb}
        }
    \end{center}
    \par\bigskip
    \begin{center}
        \subfloat[FEN]{           
            \input{figures/observations/chess_fen}
            \label{fig:observation-chess-fen}
        }
    \end{center}
   \caption{
        Sample observations for each state representation format from our chess environment, all of which we generate with the python-chess library~\citep{fiekas2012python}.
        Note that, unlike the ASCII, FEN, and RGB, which show the opening board state, the PGN corresponds to a more advanced position to visualize the move list (which would be empty for the opening board state).
    }
   \label{fig:observations-chess}
\end{figure}

\begin{figure}[t]
    \begin{center}
        \subfloat[ASCII]{\input{figures/observations/grid_world_ascii}}
        \par\bigskip
        \subfloat[Coordinates]{\input{figures/observations/grid_world_coords}}
        \hfill
        \subfloat[RGB]{
            \includegraphics[width=0.25\linewidth,valign=t]{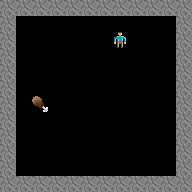}
            \label{fig:observations-grid-world-rgb}
        }
    \end{center}
   \caption{
        Sample observations from our grid world environment for all three state representation formats.
        For the RGB observations (\cref{fig:observations-grid-world-rgb}), we use the sprites from Crafter~\citep{hafner2022benchmarking} for the walls, the player, and the target (the floor is a black square).
    }
   \label{fig:observations-grid-world}
\end{figure}

To evaluate \emph{multimodal} in-context imitation learning, we consider multiple state representations for our environments --- which representations exactly depends on the environment.
For example, for chess, we evaluate four different formats: (i) ASCII, (ii) FEN, (iii) PGN, and (iv) RGB images.
We visualize all formats for each environment in \cref{fig:observations-atari,fig:observations-chess,fig:observations-crossword,fig:observations-dm-control,fig:observations-grid-world,fig:observations-tic-tac-toe}.

\paragraph{Atari -- Phoenix}

Unlike chess - where a superhuman expert policy (\ie Stockfish) is publicly available, or tic-tac-toe - where the optimal policy can be described with a closed-form algorithm, the best performance on Atari 2600 games is generally obtained by strong reinforcement learning (RL) agents~\citep{mnih2013playing, hessel2021muesli}.
However, rather than training an RL policy from scratch (which can be finicky), we make use of the training data corpus of the Gato project~\citep{reed2022generalist}.
Concretely, Gato trained a Muesli~\citep{hessel2021muesli} agent for $200$M steps and randomly recorded roughly $20$k episodes generated by the agent during training.
As a result, the dataset also contains trajectories from the beginning of training where the agent does not yet perform well, so we only consider the last $2048$ trajectories (\ie the final stages of training).
We further subsample these trajectories by only keeping the $256$ highest-scoring as our expert demonstrations.
Since we only evaluate the first $100$ steps (\ie $400$ frames with an action repeat of $4$), we subsample \wrt the cumulative reward in the first $400$ frames and not \wrt the entire episode.
Overall, we obtain a collection of demonstration episodes with a high average return of $459.9$ in the first $400$ frames. In our experiments we match the Gato setting~\citep{reed2022generalist}, \ie sticky actions and no uncontrolled random initial no-ops. To ensure variability in the evaluation episodes, we manually perform different numbers of no-ops (based on the random seed) at the beginning of every episode before starting the evaluation, which changes the initial state from which the agent has control (since enemies in Phoenix keep moving during this time).

\begin{figure}[t]
    \begin{center}
        \subfloat[Initial crossword]{
            \adjustbox{valign=t}{\input{figures/observations/crossword_empty}}
            \label{fig:observations-crossword-initial}
        }
        \hspace{64pt}
        \subfloat[Solved crossword]{
            \adjustbox{valign=t}{\input{figures/observations/crossword_solved}}
            \label{fig:observations-crossword-solved}
        }
    \end{center}
   \caption{
        Sample observations from our crossword environment.
        \cref{fig:observations-crossword-initial} shows the initial crossword and \cref{fig:observations-crossword-solved} shows the solved crossword after all the words have been placed in their corresponding slots.
        We create the crosswords of size $7\times7$ using the genxword crossword generator~\citep{whitlock2011genxword} and a list of \numprint{55189} clues collected by Matthew Ginsberg (we only use the clues with the lowest difficulty rating; the full list contains \numprint{236615} clues).
   }
   \label{fig:observations-crossword}
\end{figure}

\paragraph{DM Control -- Cheetah Run}

\begin{figure}[t]
    \begin{center}
        \subfloat[Observation]{
            \adjustbox{valign=t}{\input{figures/observations/dm_control_observation}}
        }
        \par\bigskip
        \subfloat[Action]{
            \adjustbox{valign=t}{\input{figures/observations/dm_control_action}}
        }
    \end{center}
   \caption{
        A sample observation and action from the cheetah run task from the DM Control suite~\citep{tassa2018deepmind}.
        Note that the observation is actually presented to the model on a single line (we use the multi-line representation here for ease of visualization).
        Moreover, we only show the truncated position and velocity vectors (represented by the ellipsis).
        The full position vector contains $8$ elements, and the full velocity vector contains $9$ elements.
        All elements are \texttt{float64} converted to string.
    }
   \label{fig:observations-dm-control}
\end{figure}

There is generally no closed-form or publicly available expert policy for the tasks from the DM Control suite.
Thus, we also leverage the Gato training corpus to generate our expert demonstration episodes for cheetah run.
For this task the Gato project trained a D4PG~\citep{barth2018distributed} agent, and, like for Atari, we only consider the last $10$k episodes (since those correspond to the later stages of training and thus better performance). 
Ideally we would want to subsample the highest-scoring trajectories as expert demonstrations, but, unfortunately, the underlying MuJoCo~\citep{todorov2012mujoco} physics have changed since the time of the Gato data collection, which means that the rewards in the dataset no longer match the observation-actions pairs in our DM Control environment (\ie replaying the actions from the same initial state does not yield the same return).
Thus, we first replay the actions for every trajectory in our collection (setting the initial state based on the first observation) and use the new returns to subsample the $1000$ highest-scoring trajectories (again only considering the cumulative reward in the first $100$ steps).
Overall, we obtain a collection of demonstration episodes with a high average return of $37.4$ in the first $100$ steps.

\subsection{Prompts}
\label{app:experimental-details:prompts}

The frozen part and the dynamic part of the evaluation prompt are illustrated in \cref{code:frozen-prompt} and \cref{code:dynamic-prompt}, respectively.

\begin{listing}
    \caption{
        The frozen part of the evaluation prompt, which contains the expert demonstration episodes and stays constant throughout an evaluation episode.
        In this example, we have 8 demonstration episodes with 10 steps each and RGB observations.
        Before feeding the prompt to the model, we replace the observation and action placeholders with the actual observations (\ie images in this case) and action strings.
    }
    \label{code:frozen-prompt}
   \begin{minted}[mathescape,
               fontsize=\footnotesize,
               linenos,
               numbersep=5pt,
               gobble=7,
               frame=lines,
               framesep=2mm]{python}
        demonstration_prompt = '''
        You are a powerful reinforcement learning agent. You can effectively identify a policy
        exposed by demonstrations and reproduce it in a new situation.
        
        Here are a number of demonstrations:
        
        Observation: <IMG_0_0> Action: <AC_0_0>
        Observation: <IMG_0_1> Action: <AC_0_1>
        ...
        Observation: <IMG_0_9> Action: <AC_0_9>
        
        Observation: <IMG_1_0> Action: <AC_1_0>
        Observation: <IMG_1_1> Action: <AC_1_1>
        ...
        Observation: <IMG_1_9> Action: <AC_1_9>
        
        ...
        
        Observation: <IMG_7_0> Action: <AC_7_0>
        Observation: <IMG_7_1> Action: <AC_7_1>
        ...
        Observation: <IMG_7_9> Action: <AC_7_9>
        '''
    \end{minted}
\end{listing}

\begin{listing}
    \caption{
        The dynamic part of the evaluation prompt containing the evaluation trajectory.
        While stepping through an environment, we append this prompt to the one in \cref{code:frozen-prompt} in every evaluation step (\eg for the 3rd step here), again replacing the observation and action placeholders with the actual observations and actions.
        This example also shows the legal moves (lines $8$ to $10$) and uses chain-of-thought prompting (lines $12$ to $17$, \citet{wei2022chain}), both of which may be omitted depending on our ablations in \cref{app:additional-results:ablations} (which can vary for each model-task combination).
    }
    \label{code:dynamic-prompt}
    \begin{minted}[mathescape,
               fontsize=\footnotesize,
               linenos,
               numbersep=5pt,
               gobble=7,
               frame=lines,
               framesep=2mm]{python}
        evaluation_prompt = '''
        This is the current situation:
        
        Observation: <IMG_8_0> Action: <AC_8_0>
        Observation: <IMG_8_1> Action: <AC_8_1>
        Observation: <IMG_8_2>
        
        In this situation, this is the list of all the moves that are legal:
        
        no action, jump left, left
        
        Given the demonstrations and the current situation, you should infer the next logical
        action. Check that the chosen action is in the set of legal moves. Think step by step
        and very briefly explain your reasoning for choosing this  action. You must answer with
        the reasoning followed by the action in the following format:
        Reasoning: ...
        Action: ...
        '''
    \end{minted}
\end{listing}

    \section{Additional Results}
\label{app:additional-results}

In this section, we present additional results and ablations from our experimental evaluation.

\subsection{Ablating the Maximum Sample Length}
\label{app:additional-results:max-sample-length-ablation}

\begin{figure}
    \begin{subfigure}[t]{\textwidth}
        \begin{center}
            \includegraphics[width=\textwidth]{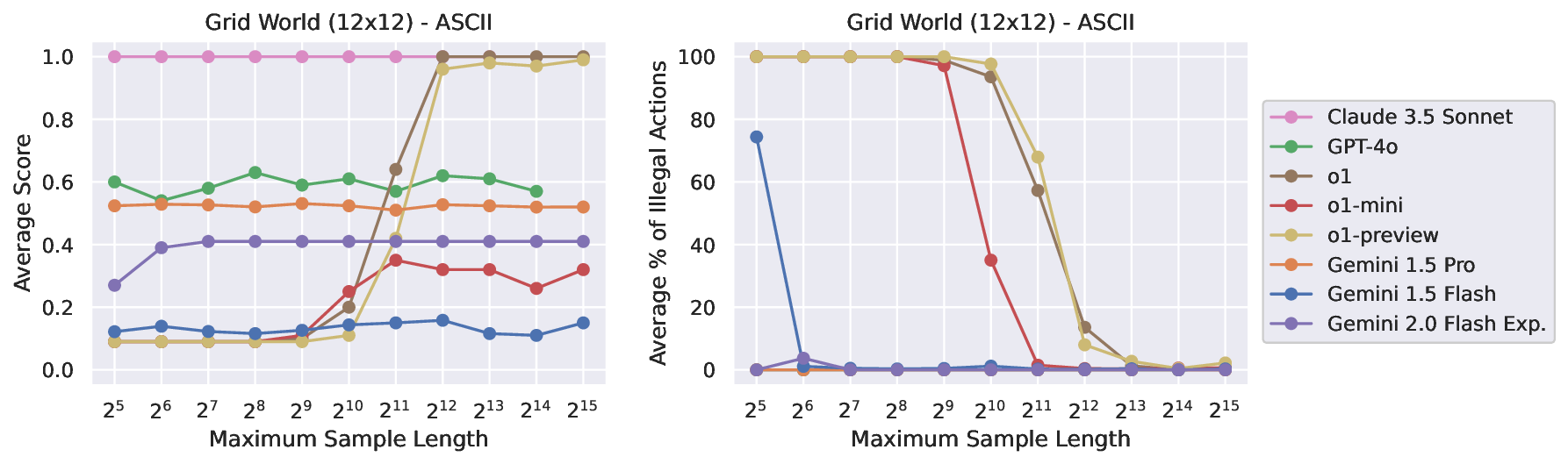}
        \end{center}
        \caption{ASCII observations}
    \end{subfigure}
    \par\bigskip
    \begin{subfigure}[t]{\textwidth}
        \begin{center}
            \includegraphics[width=\textwidth]{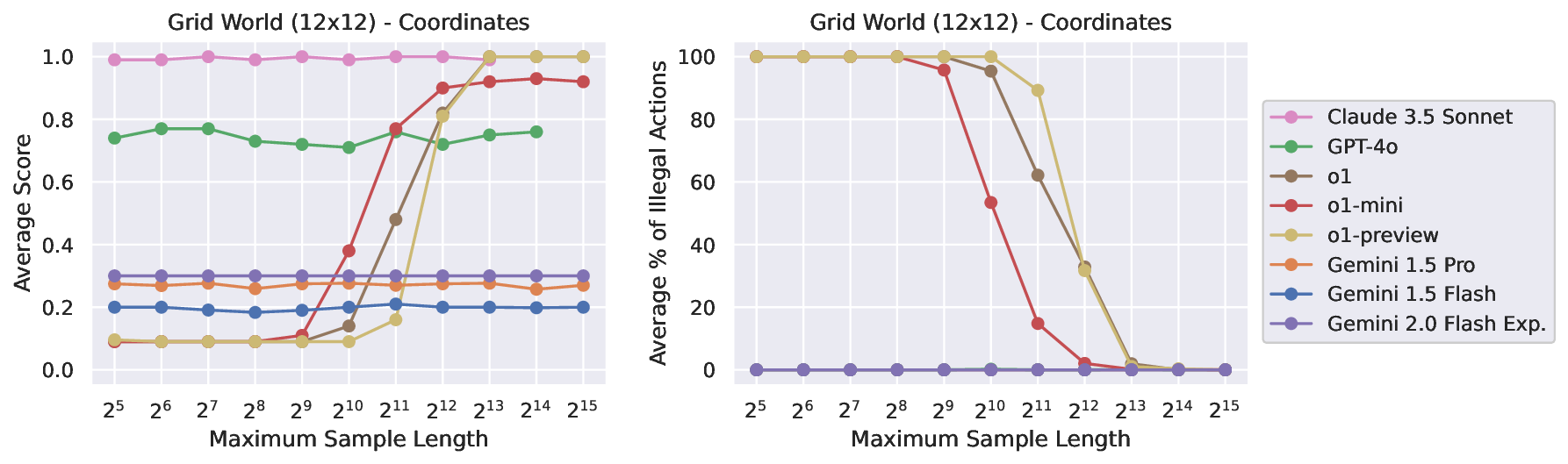}
        \end{center}
        \caption{Coordinate observations}
    \end{subfigure}
    \par\bigskip
    \begin{subfigure}[t]{\textwidth}
        \begin{center}
            \includegraphics[width=\textwidth]{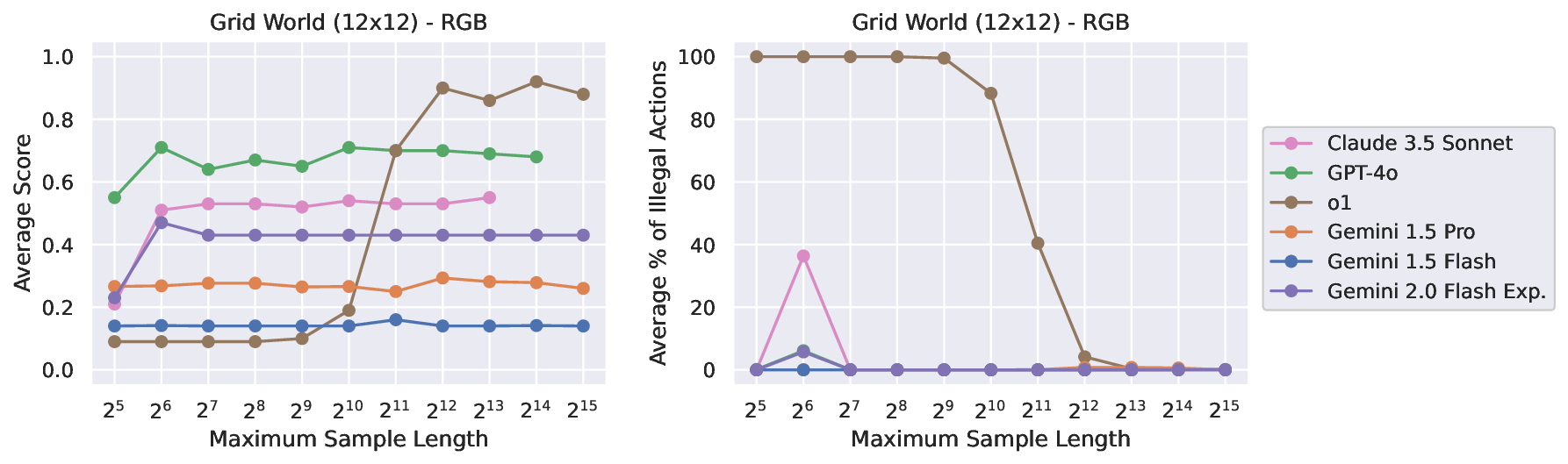}
        \end{center}
        \caption{RGB observations}
    \end{subfigure}
    \caption{
        Ablating the maximum sample length for all three observation types (ASCII, coordinates, and RGB images) of the grid world navigation task with $1$ demonstration episode (the left panels show the average score, the right panels show the percentage of illegal actions).
        As expected, $32$ output tokens are sufficient for \claude{}, \gpt{}, \flash{}, and \pro{} to achieve their best performance on the task.
        In contrast, \mini{} and \preview{} require between $4096$ and $8192$ output tokens to achieve their best performance.
        With less than $8192$ output tokens, the o1 models do not have enough internal ``reasoning tokens'' at their disposal to produce an output, which can be seen by the sharp increase in the percentage of illegal actions in the right panels.
        Note that \mini{} and \preview{} are text-only models and therefore cannot process RGB image observations.
    }
    \label{fig:sample-length-ablation}
\end{figure}

The o1 family of models tends to generate (long) internal ``reasoning traces'' before returning an output.
Thus, if the maximum sample length is not large enough, these models may not have enough ``reasoning tokens'' and therefore do not produce an output (in which case the API returns an empty sample).
Since we want to report each model's best performance on our benchmark, we therefore ablate the maximum number of sample tokens and choose the configuration that trades off good performance and low cost (since, after a certain point, more tokens generally do not improve performance but only increase the cost).
The maximum sample length also has an impact on the other models, particularly when using chain-of-thought reasoning, but a relatively low sample length typically suffices for our tasks (unlike the o1 models which require large maximal sample length).

To that end, \cref{fig:sample-length-ablation} shows our ablation of the maximum sample length for \claude{}, \gpt{}, \mini{}, \preview{}, \flash{}, and \pro{} on all three observation types (ASCII, coordinates and RGB) for the grid world navigation task with $1$ demonstration episode.
\cref{fig:sample-length-ablation} shows the average score over $100$ evaluation episodes for maximum sample lengths from $32$ to $32768$ tokens (\claude{} only supports $8192$ output tokens and \gpt{} only supports $16384$ output tokens).
Unsurprisingly, the performance of \claude{}, \gpt{}, \flash{}, and \pro{} is largely unaffected by the number of sample tokens, \ie even $32$ tokens are sufficient to achieve their best performance (with the exception of \claude{} and \gpt{} on RGB observations, where they benefit from having $64$ tokens).
This is in stark contrast with \mini{} and \preview{}, which require between $4096$ and $8192$ output tokens to achieve their optimal performance (with a very steep degradation below $4096$).
To verify whether this sharp rise in performance is actually due to the model not generating a valid action, \cref{fig:sample-length-ablation} also visualizes the average percentage of illegal actions per episode over the maximum sample length, which (inversely) correlates very strongly with performance for the o1 models.
We therefore set the maximum sample length to $8192$ for the o1 models as a tradeoff between cost and performance (the maximum for \preview{} is $32768$, the maximum for \mini{} is $65536$~\citep{openai2024reasoning}).
This should enable the o1 models to achieve their best performance on our benchmark --- even though they do so at a much higher (computational) cost than the other models.

\subsection{Replaying a Demonstration Episode}
\label{app:additional-results:replay}

In-context imitation learning requires several different skills, one of which is being able to locate and retrieve the relevant demonstration(s) from the context.
We therefore conduct a ``sanity check'' experiment where we provide a single demonstration episode and ``replay'' the exact same episode for evaluation (akin to a multimodal sequence copying task).
Thus, for every step, the model only has to find the correct location in the demonstration episode in its context and reproduce its corresponding action.
Accordingly, we slightly change the evaluation setup since the next observation (both action and next state) in the evaluation trajectory is always determined by the demonstration episode and not by the action generated by the model (\ie we perform teacher forcing rather than a dynamic evaluation).
As a result, we are not interested in a model's return but in how often it manages to match the action from the demonstration episode.
Like in our other experiments, we evaluate $100$ different episodes.
For the sake of simplicity, we do not use chain-of-thought prompting, do not show the legal actions in the prompt, and only consider a single demonstration episode.
Other than that, we leave the prompt the same as in our default experimental setup (\ie including the same preamble and separator).

\begin{figure}
    \includegraphics[width=\textwidth]{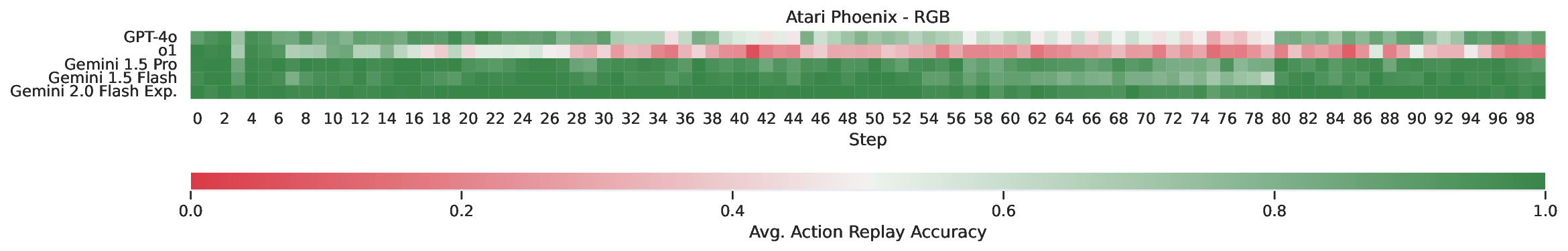}
    \caption{
        Replaying the demonstration episode for the Phoenix game (with RGB observations) from Atari $2600$.
        The color visualizes the models' average accuracy when attempting to replay the action for a given step.
        The Gemini 1.5 models generally perform well at replaying the demonstrations, regardless of the step, though they show a slight degradation in performance towards the $80$-step mark, from which they immediately recover again.
        The performance of \gpt{} shows a slightly higher degradation towards step $80$, but also immediately recovers thereafter.
        Note that \claude{} cannot process more than $100$ images at a time, so we cannot evaluate it on this task.
        Similarly, \mini{} and \preview{} are text-only and, therefore, cannot process RGB observations.
    }
    \label{fig:replay-atari}
\end{figure}

\begin{figure}
    \begin{subfigure}[t]{\textwidth}
        \begin{center}
            \includegraphics[width=0.9\textwidth]{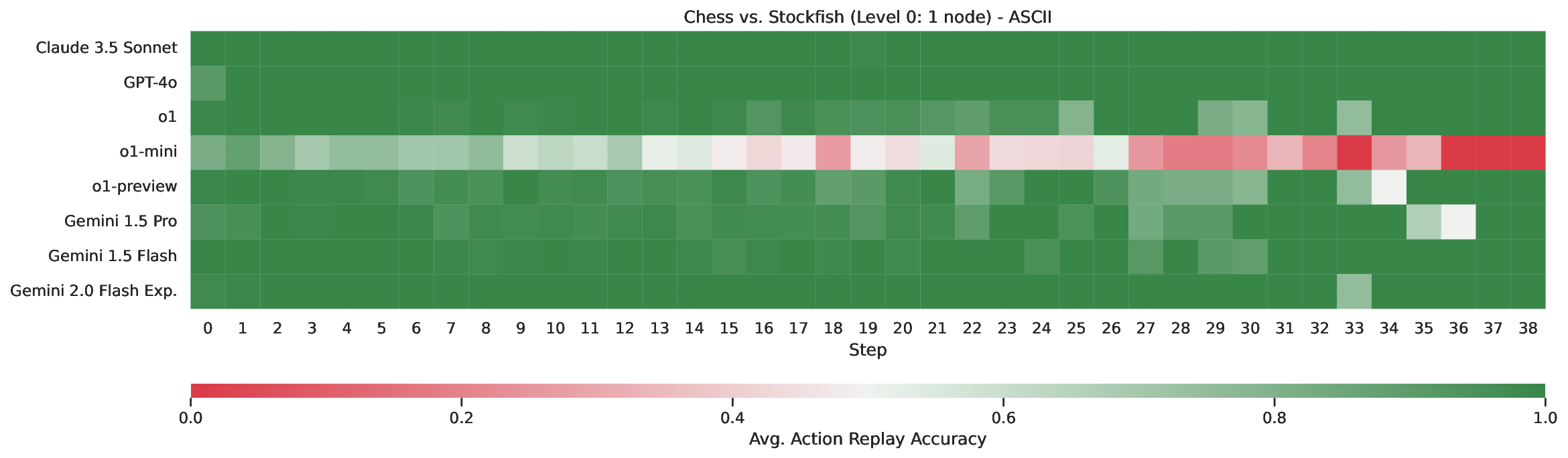}
        \end{center}
        \vspace{-8pt}
        \caption{ASCII observations}
    \end{subfigure}
    \par\bigskip
    \begin{subfigure}[t]{\textwidth}
        \begin{center}
            \includegraphics[width=0.9\textwidth]{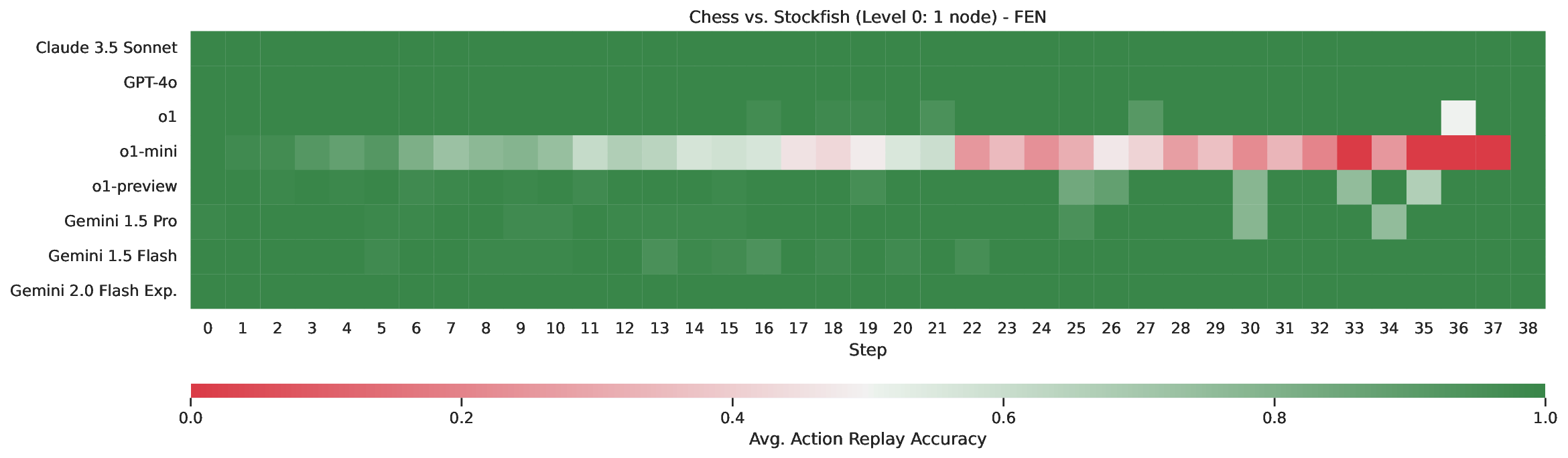}
        \end{center}
        \vspace{-8pt}
        \caption{FEN observations}
    \end{subfigure}
    \par\bigskip
    \begin{subfigure}[t]{\textwidth}
        \begin{center}
            \includegraphics[width=0.9\textwidth]{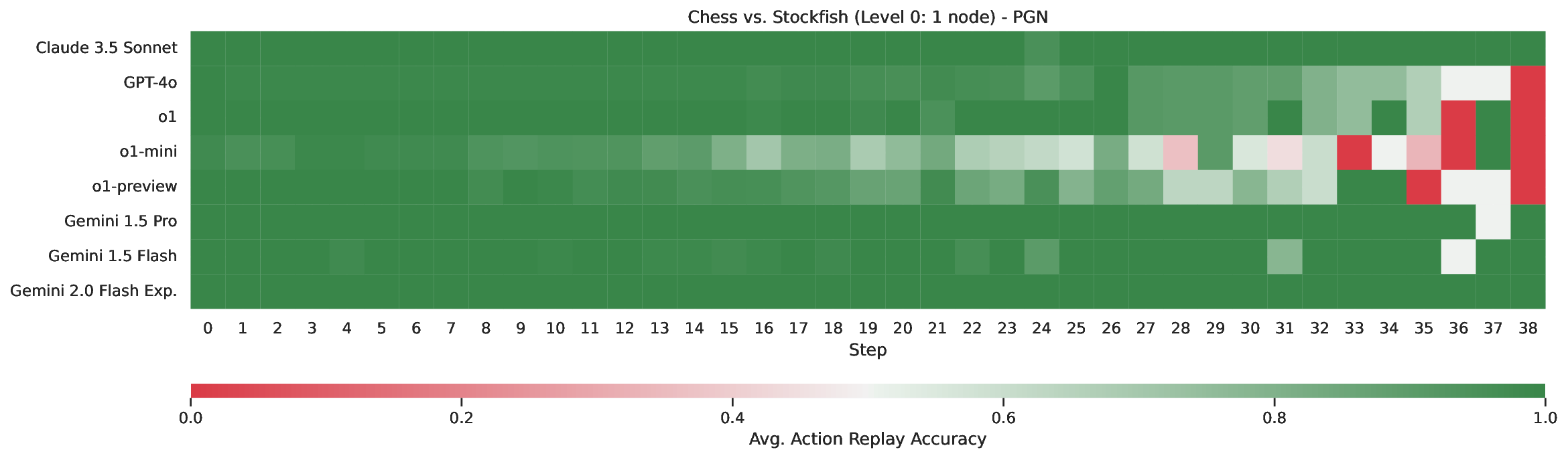}
        \end{center}
        \vspace{-8pt}
        \caption{PGN observations}
    \end{subfigure}
    \par\bigskip
    \begin{subfigure}[t]{\textwidth}
        \begin{center}
            \includegraphics[width=0.9\textwidth]{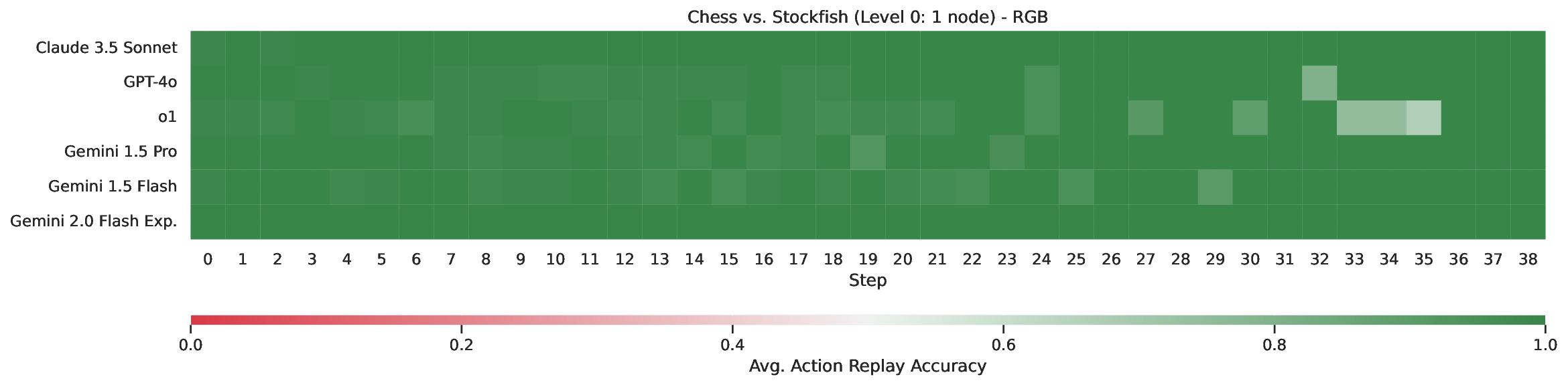}
        \end{center}
        \vspace{-8pt}
        \caption{RGB observations}
    \end{subfigure}
    \caption{
        Replaying the demonstration episode for the different observation types from our chess environments.
        The color visualizes the models' average accuracy when attempting to replay the action for a given step.
        All models generally perform well across all observation types, except for \mini{}, which shows a strong performance degradation towards the end of the episode across all three observation text observation types (recall that \mini{} and \preview{} are text-only models, and, therefore, cannot process RGB images).
        With the exception of \claude{}, all models struggle (to varying degrees) to replay the last few actions with PGN observations.
    }
    \label{fig:replay-chess}
\end{figure}

\begin{figure}
    \begin{center}
        \includegraphics[width=0.6\textwidth]{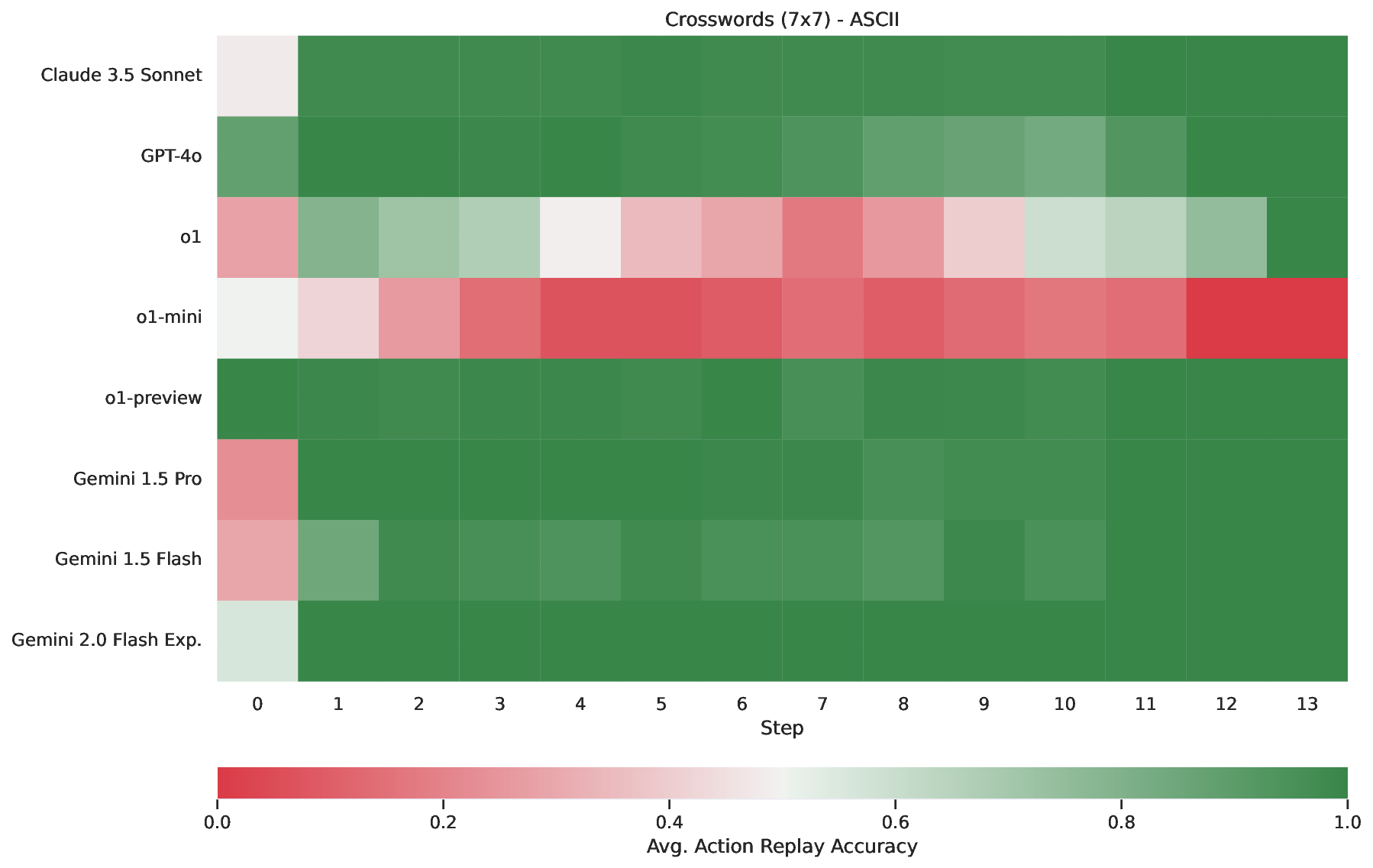}
    \end{center}
    \caption{
        Replaying the demonstration episode for our crossword environment with ASCII observations.
        The color visualizes the models' average accuracy when attempting to replay the action for a given step.
        All models generally perform well, except for \mini{}, which completely fails at this task.
        Moreover, \claude{}, \flash{}, and \pro{} struggle to replay the first step.
    }
    \label{fig:replay-crossword}
\end{figure}

\begin{figure}
    \begin{center}
        \includegraphics[width=\textwidth]{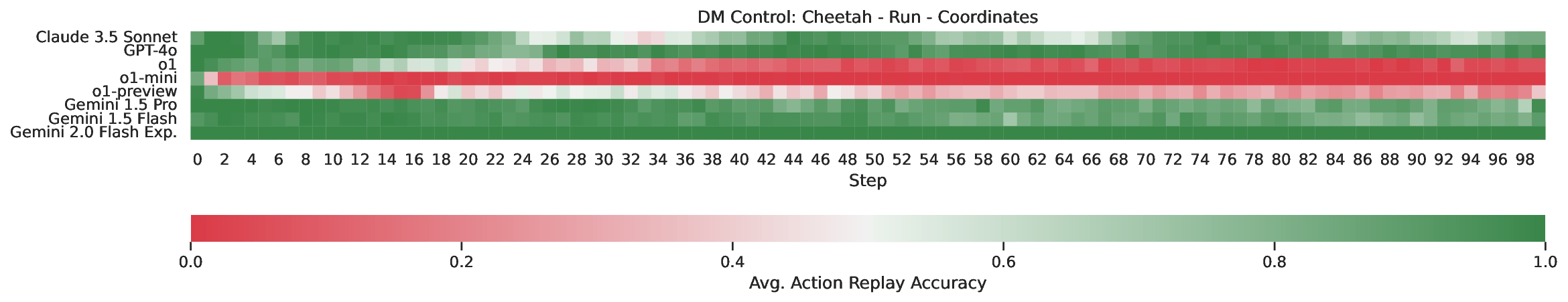}
    \end{center}
    \caption{
        Replaying the demonstration episode for the cheetah run task from the DM Control suite (with coordinate observations).
        The color visualizes the models' average accuracy when attempting to replay the action for a given step.
        \gpt{}, \flash{}, and \pro{} generally perform well.
        In contrast, \mini{} and \preview{} significantly struggle with this task.
        \claude{} shows patches of slightly poorer performance around steps $32$, $64$, and $96$, but performs well otherwise.
    }
    \label{fig:replay-dm-control}
\end{figure}

\begin{figure}
    \begin{subfigure}[t]{0.5\textwidth}
        \begin{center}
            \includegraphics[width=\textwidth]{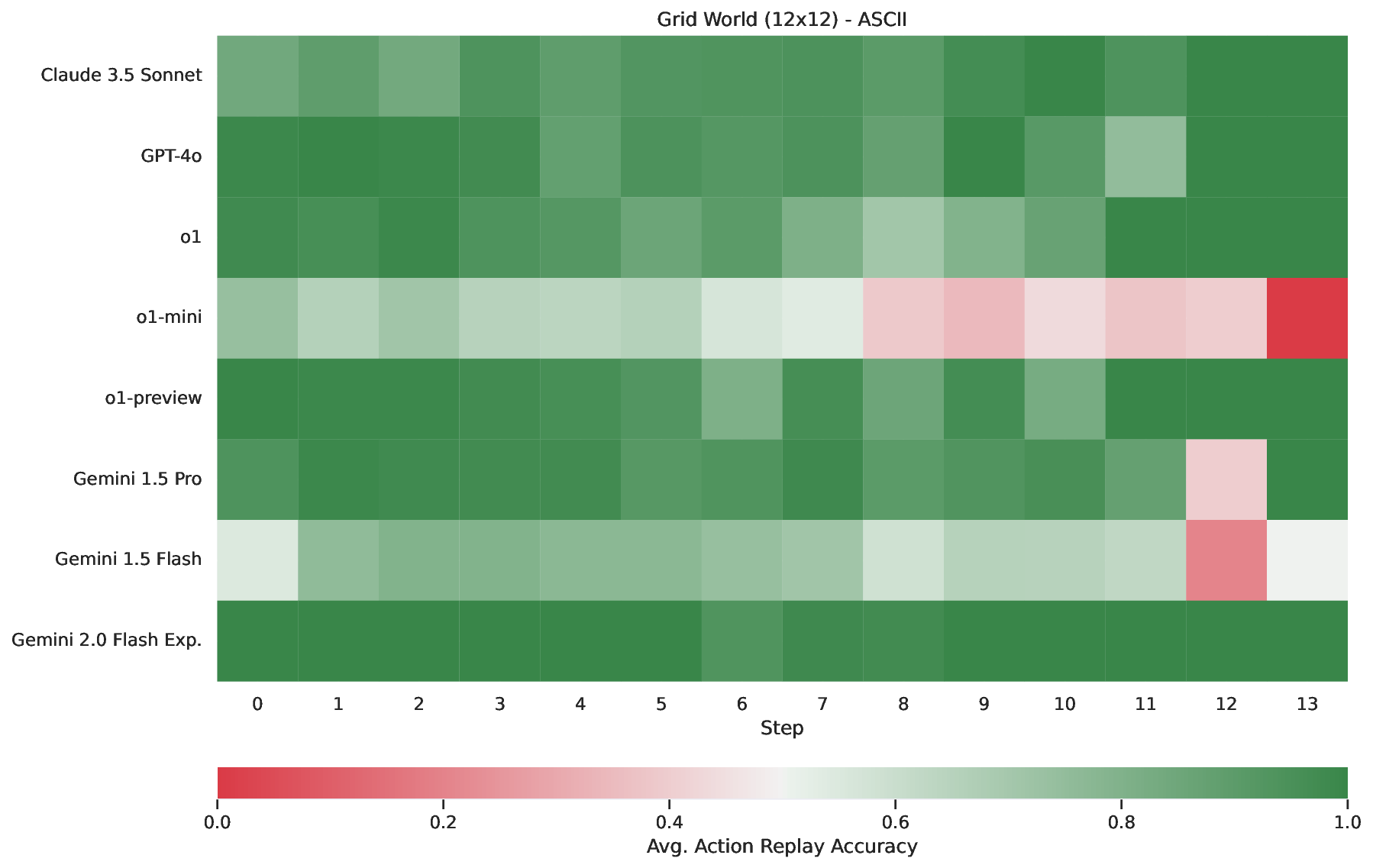}
        \end{center}
       \caption{ASCII observations}
    \end{subfigure}
    \begin{subfigure}[t]{0.5\textwidth}
        \begin{center}
            \includegraphics[width=\textwidth]{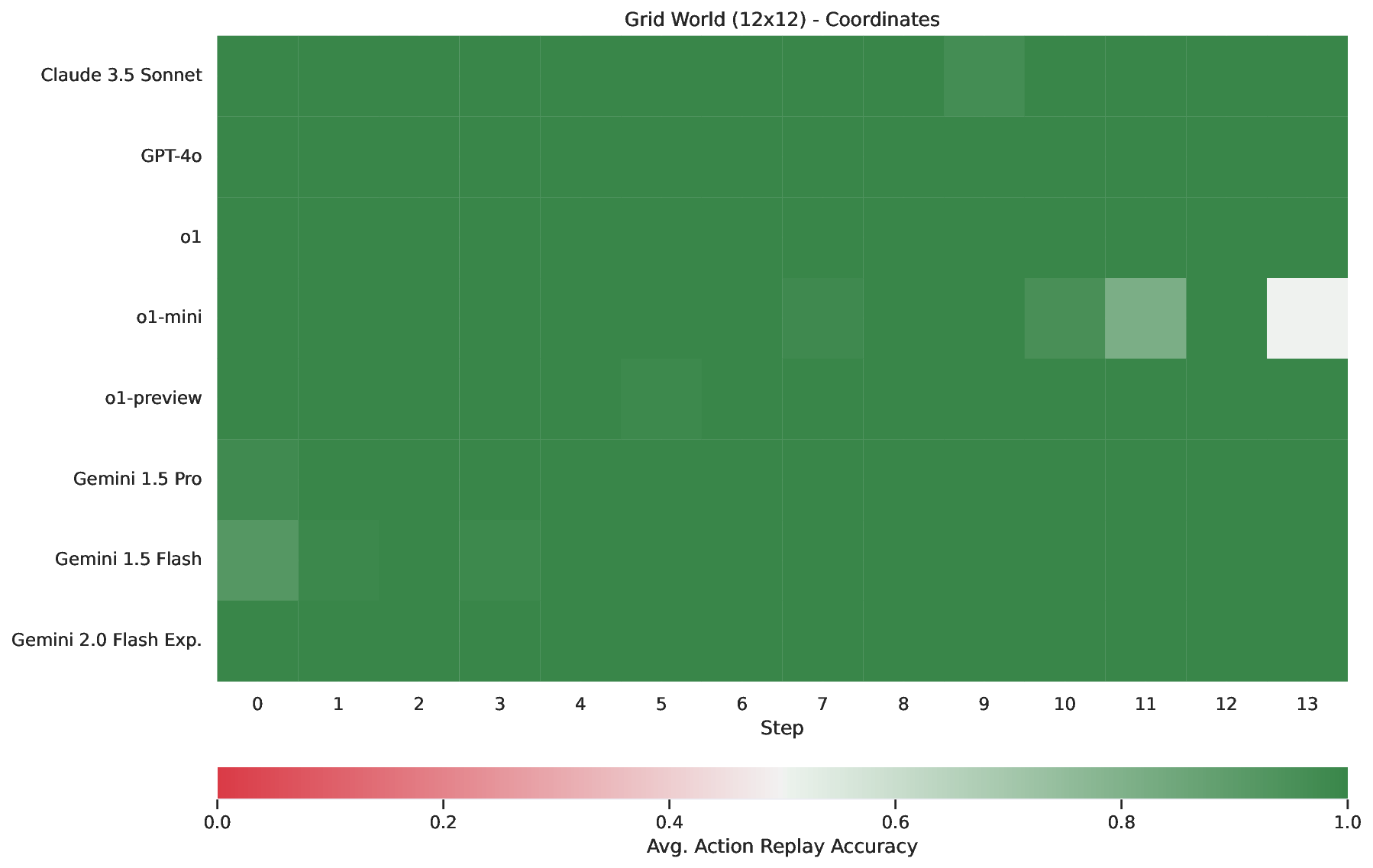}
        \end{center}
        \caption{Coordinate observations}
    \end{subfigure}
    \par\bigskip
    \begin{subfigure}[t]{0.5\textwidth}
        \begin{center}
            \includegraphics[width=\textwidth]{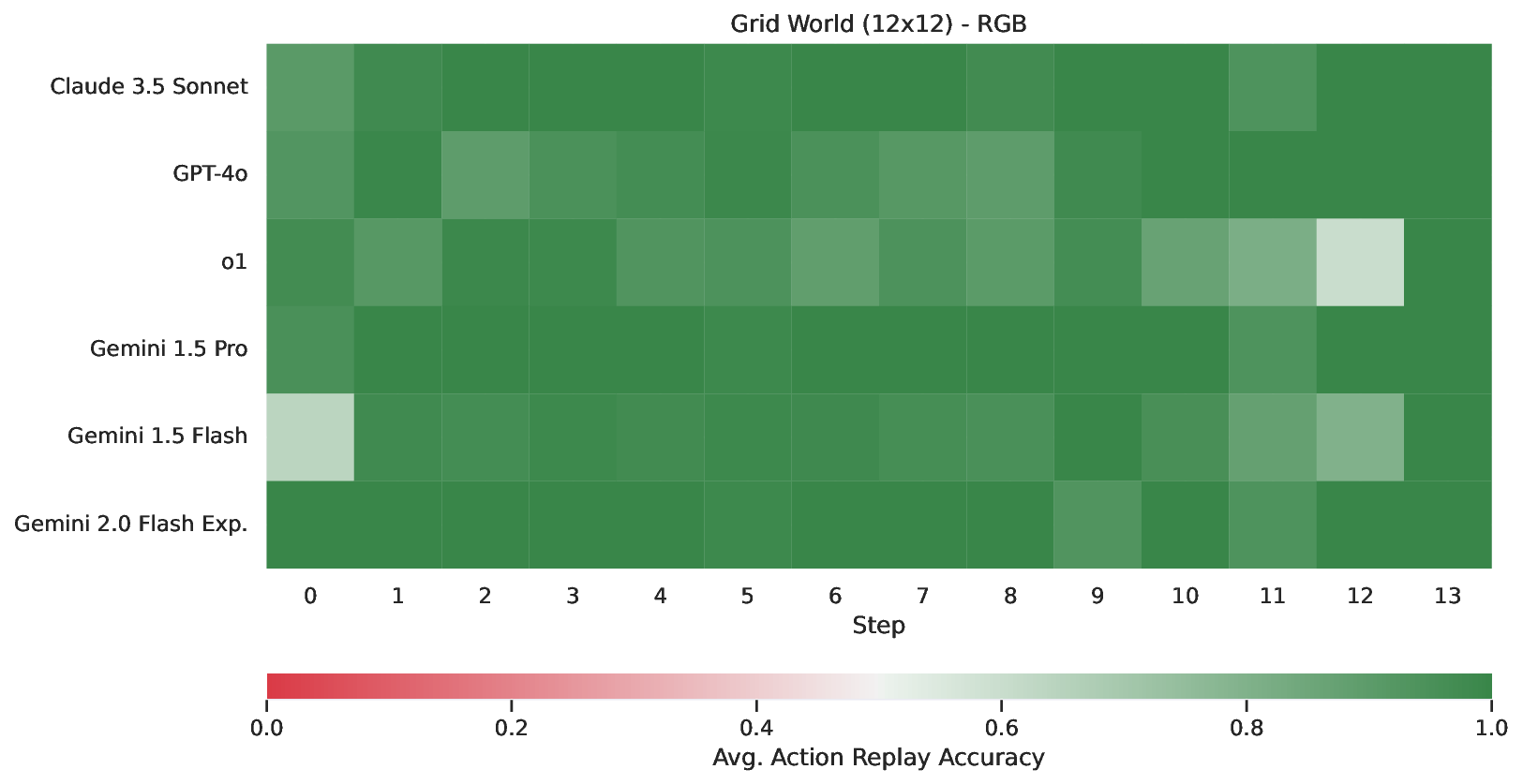}
        \end{center}
        \caption{RGB observations}
    \end{subfigure}
    \caption{
        Replaying the demonstration episode for the all observation types from our grid world navigation task.
        The color visualizes the models' average accuracy when attempting to replay the action for a given step.
        All models generally perform well across all observation types, except for \mini{} on ASCII observations.
        Moreover, the Gemini 1.5 models struggle on step $12$ with ASCII observations, and \mini{} struggles with the last step for coordinate observations.
        Note that \mini{} and \preview{} are text-only models and, therefore, cannot process RGB observations.
    }
    \label{fig:replay-grid-world}
\end{figure}

\begin{figure}
    \null\hfill
    \begin{subfigure}[t]{0.25\textwidth}
        \begin{center}
            \includegraphics[width=\textwidth]{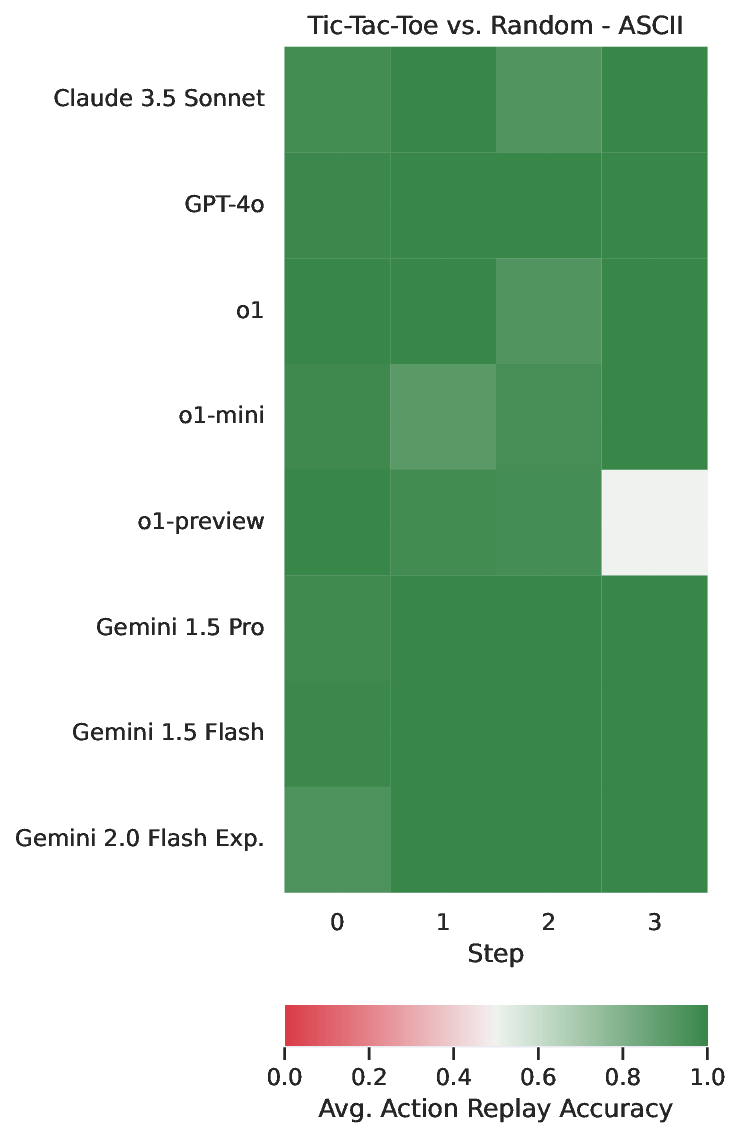}
        \end{center}
        \caption{ASCII observations}
    \end{subfigure}
    \hfill
    \begin{subfigure}[t]{0.25\textwidth}
        \begin{center}
            \includegraphics[width=\textwidth]{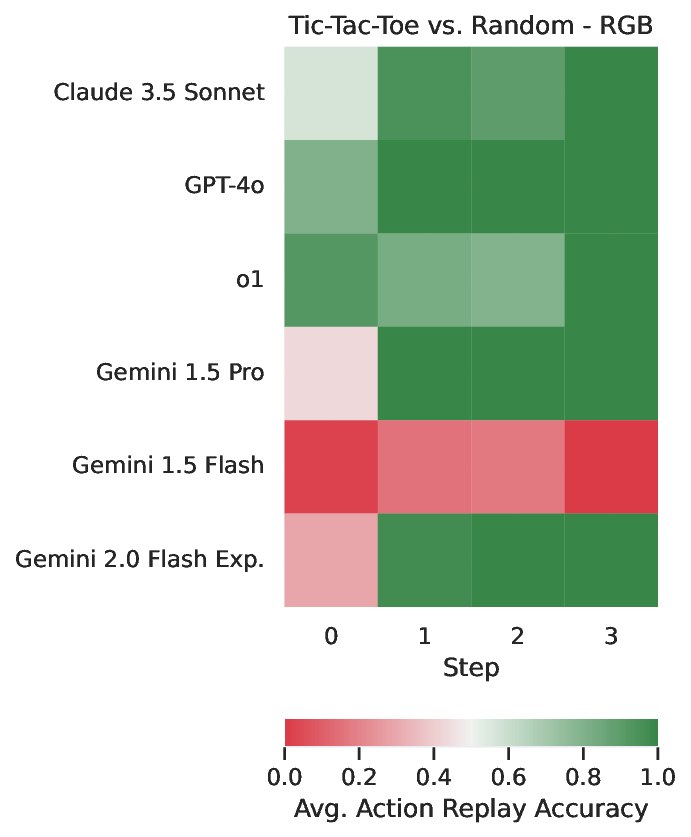}
        \end{center}
        \caption{RGB observations}
    \end{subfigure}
    \hfill\null
    \caption{
        Replaying the demonstration episode for the all observation types from our tic-tac-toe environment.
        The color visualizes the models' average accuracy when attempting to replay the action for a given step.
        All models generally perform well across all observation types except for \flash{} on RGB images, where it fails completely.
        \claude{} and \pro{} also struggle (to varying degrees) with the first step (step $0$) for RGB observations.
        Note that \mini{} and \preview{} are text-only models and, therefore, cannot process RGB image observations.
    }
    \label{fig:replay-tic-tac-toe}
\end{figure}

\cref{fig:replay-atari,fig:replay-chess,fig:replay-crossword,fig:replay-dm-control,fig:replay-grid-world,fig:replay-tic-tac-toe} show the average action replay accuracy per evaluation step for all frontier models on all six environments (for all state representation formats separately).
We observe that models are generally capable of replaying the demonstration episodes, with slightly lower performance deeper into the episode.
The significant exception is \mini{} which struggles with the replay task in most settings.
Similarly, \flash{} fails to replay the actions for tic-tac-toe with RGB observations.
Note that, in theory, the observation type is irrelevant for this task since the models could just count the number of observations in the evaluation trajectory and select the corresponding action in the demonstration trajectory.

\begin{figure}
    \begin{subfigure}[t]{0.32\textwidth}
        \begin{center}
            \includegraphics[width=\textwidth]{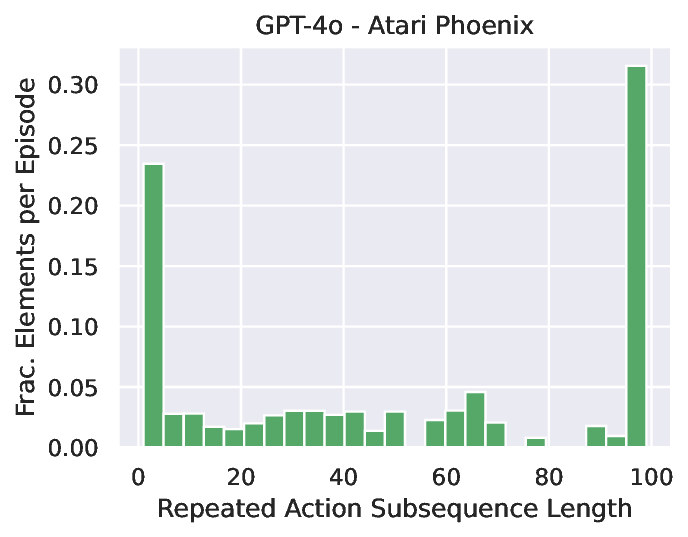}
        \end{center}
       \caption{\gpt{}}
    \end{subfigure}
    \hfill
    \begin{subfigure}[t]{0.32\textwidth}
        \begin{center}
            \includegraphics[width=\textwidth]{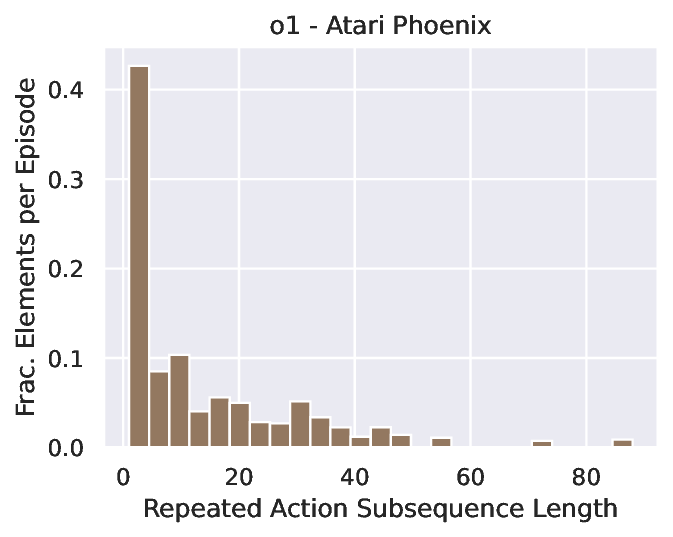}
        \end{center}
       \caption{o1}
    \end{subfigure}
    
    \begin{subfigure}[t]{0.32\textwidth}
        \begin{center}
            \includegraphics[width=\textwidth]{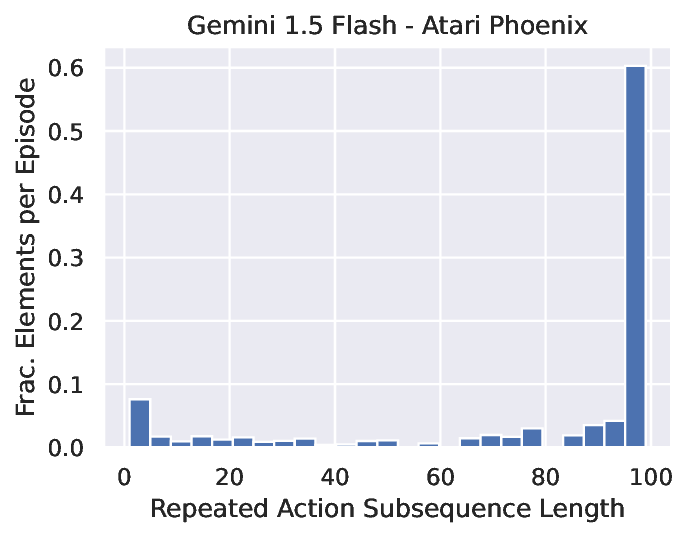}
        \end{center}
        \caption{\flash{}}
    \end{subfigure}
    \hfill
    \begin{subfigure}[t]{0.32\textwidth}
        \begin{center}
            \includegraphics[width=\textwidth]{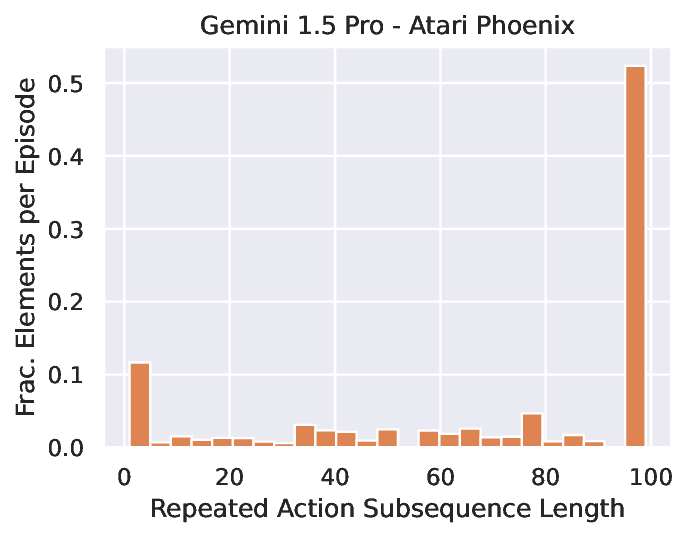}
        \end{center}
        \caption{\pro{}}
    \end{subfigure}
    \hfill
    \begin{subfigure}[t]{0.32\textwidth}
        \begin{center}
            \includegraphics[width=\textwidth]{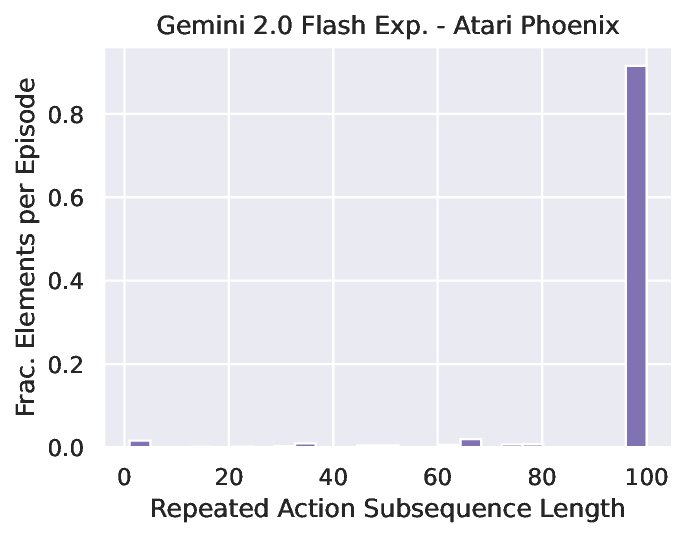}
        \end{center}
        \caption{\expflash{}}
    \end{subfigure}
    \caption{
        Fraction of elements per repeated action subsequence length in the Phoenix game from Atari $2600$ (without demonstration episodes).
        To compute these results, we partition each evaluation episode into segments consisting of the same action generated consecutively.
        The segment length is on the x-axis, and the height of each bar is the fraction of actions (out of the total number of actions) across all time steps in all evaluation episodes that fall into each segment length.
        All three models have many episodes (\eg more than $60$\% for \flash{}) where they repeat the same action throughout almost the entire episode.
        Accordingly, they fire very rarely and thus achieve a low score (\cf \cref{fig:atari-phoenix}), since, in the Phoenix game, in order to fire repeatedly, the firing button needs to be pressed and released repeatedly --- constantly holding it down (\ie repeating the previous action) only results in a single shot.
    }
    \label{fig:atari-subsequence-length}
\end{figure}

\subsection{Illegal Actions}
\label{app:additional-results:illegal-actions}

\begin{figure}
    \begin{center}
        \includegraphics[width=0.5\textwidth]{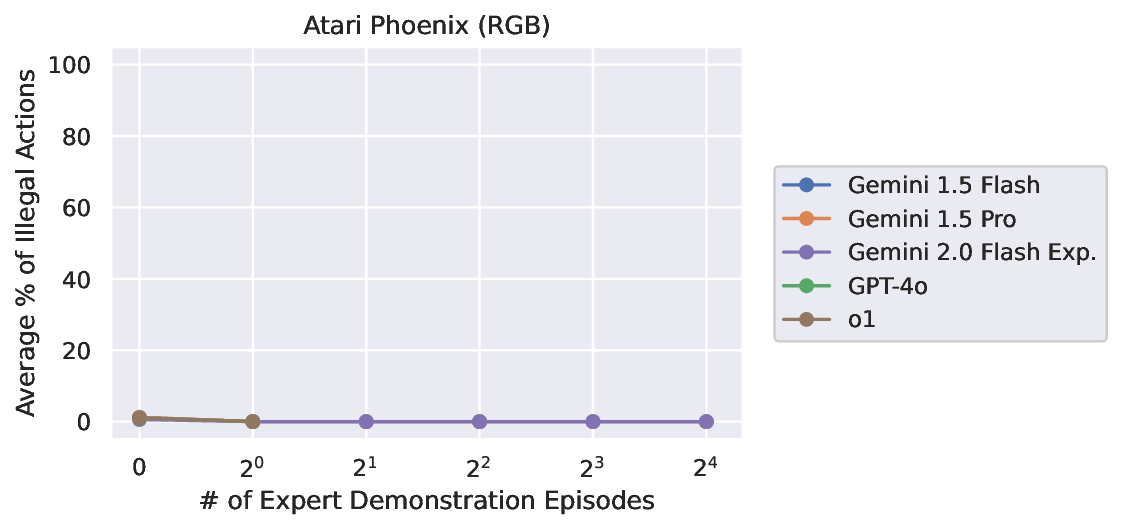}
    \end{center}
    \caption{
        Average percentage of illegal actions per episode over the number of expert demonstrations for the Phoenix game (RGB observations) from Atari 2600.
        All models mostly generate legal actions.
        Note that \mini{} and \preview{} are text-only and, thus, cannot process RGB observations.
    }
    \label{fig:atari-illegal-actions}
\end{figure}

\begin{figure}
    \begin{subfigure}[t]{0.5\textwidth}
        \begin{center}
            \includegraphics[width=\textwidth]{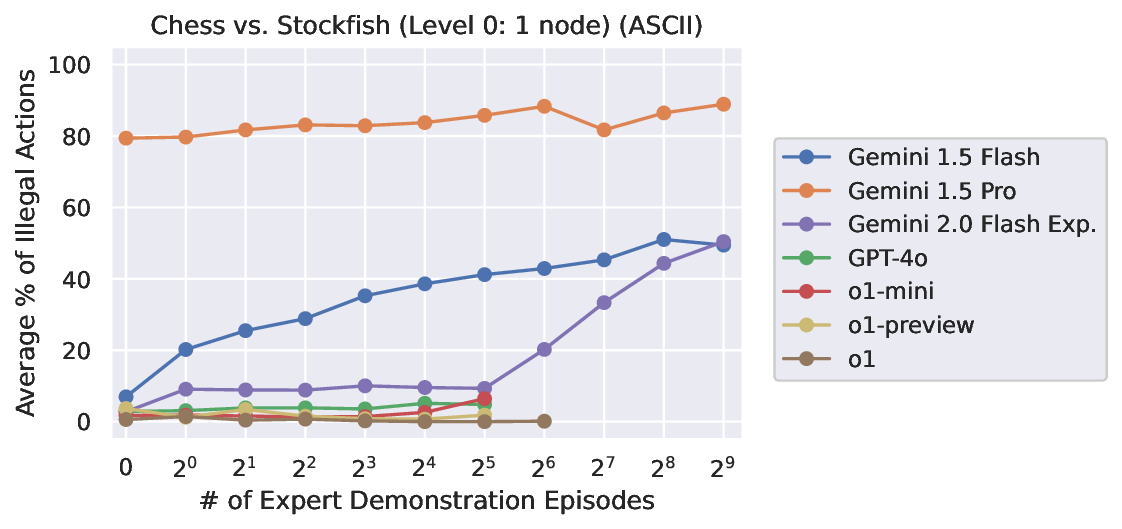}
        \end{center}
        \caption{ASCII observations}
    \end{subfigure}
    \begin{subfigure}[t]{0.5\textwidth}
        \begin{center}
            \includegraphics[width=\textwidth]{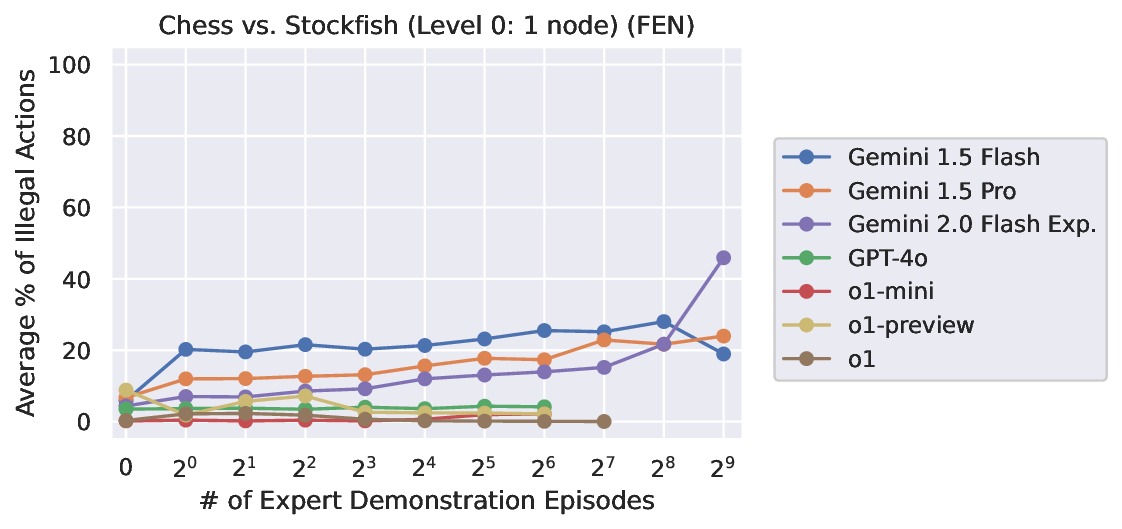}
        \end{center}
        \caption{FEN observations}
    \end{subfigure}
    \par\bigskip
    \begin{subfigure}[t]{0.5\textwidth}
        \begin{center}
            \includegraphics[width=\textwidth]{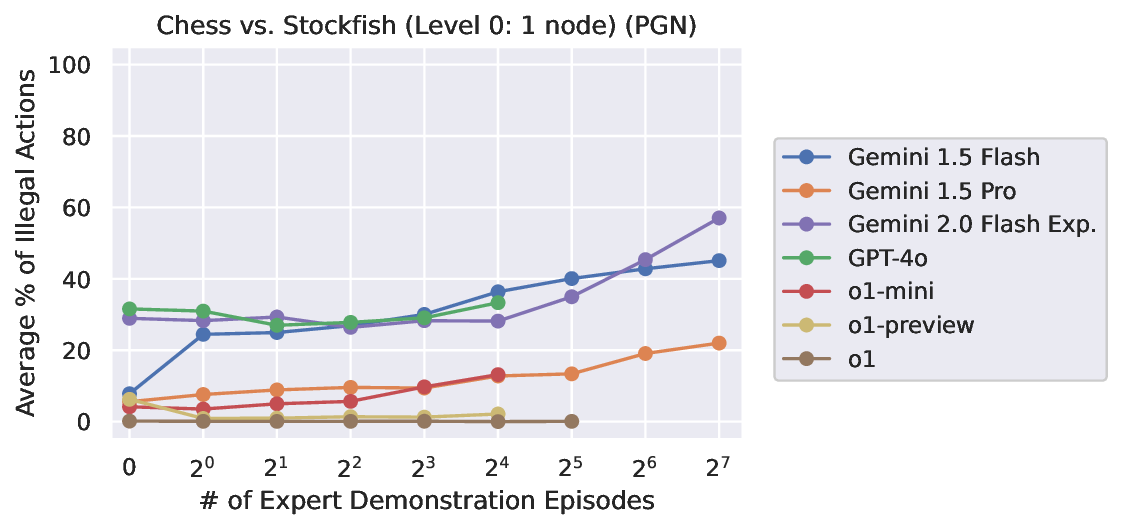}
        \end{center}
        \caption{PGN observations}
    \end{subfigure}
    \begin{subfigure}[t]{0.5\textwidth}
        \begin{center}
            \includegraphics[width=\textwidth]{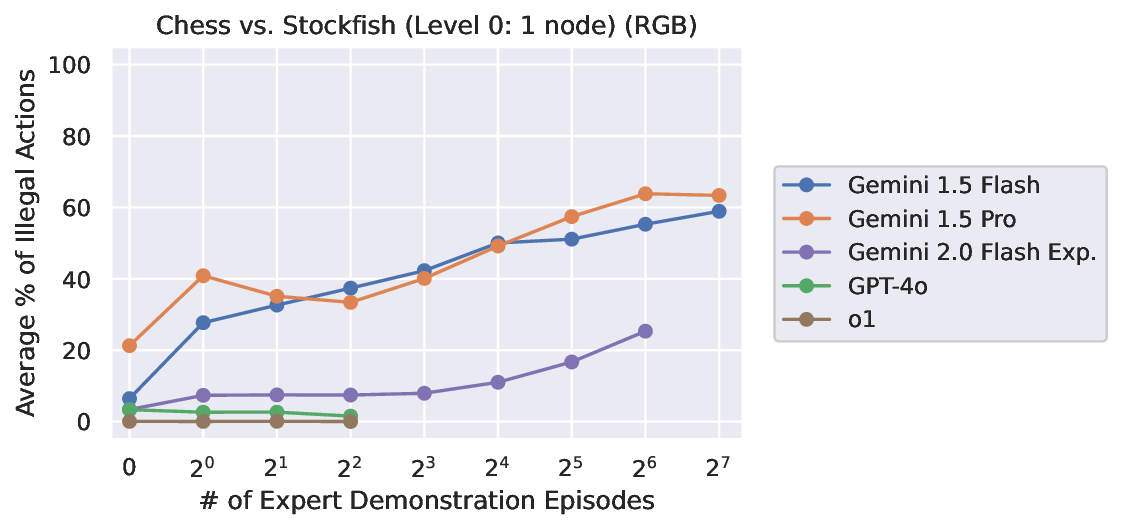}
        \end{center}
        \caption{RGB observations}
    \end{subfigure}
    \caption{
        Average percentage of illegal actions per episode over the number of expert demonstrations for all state representation formats from our chess environment.
        \gpt{}, \mini{}, and \preview{} rarely generate illegal actions (except for \gpt{} with PGN observations).
        In contrast, the Gemini 1.5 models consistently struggle to generate legal actions across all state representation formats and with higher rates of illegal actions with increasing numbers of expert demonstrations in the context.
        Note that \mini{} and \preview{} are text-only models and therefore cannot process RGB image observations.
    }
    \label{fig:chess-illegal-actions}
\end{figure}

\begin{figure}
    \begin{minipage}{0.475\textwidth}
            \begin{center}
                \includegraphics[width=\textwidth]{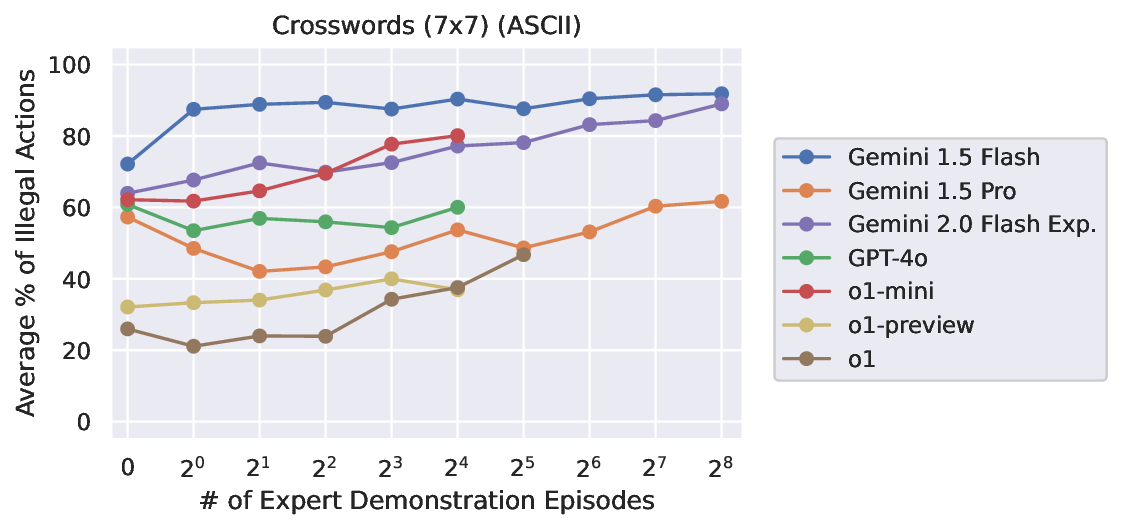}
            \end{center}
            \captionof{figure}{
                Average percentage of illegal actions per episode over the number of expert demonstrations for our crossword environment with ASCII observations.
                All models produce a high percentage of illegal actions (between $\sim40$\% and $\sim90$\% of all the actions in an episode).
            }
            \label{fig:crossword-illegal-actions}
    \end{minipage}
    \hfill
    \begin{minipage}{0.475\textwidth}
            \begin{center}
                \includegraphics[width=\textwidth]{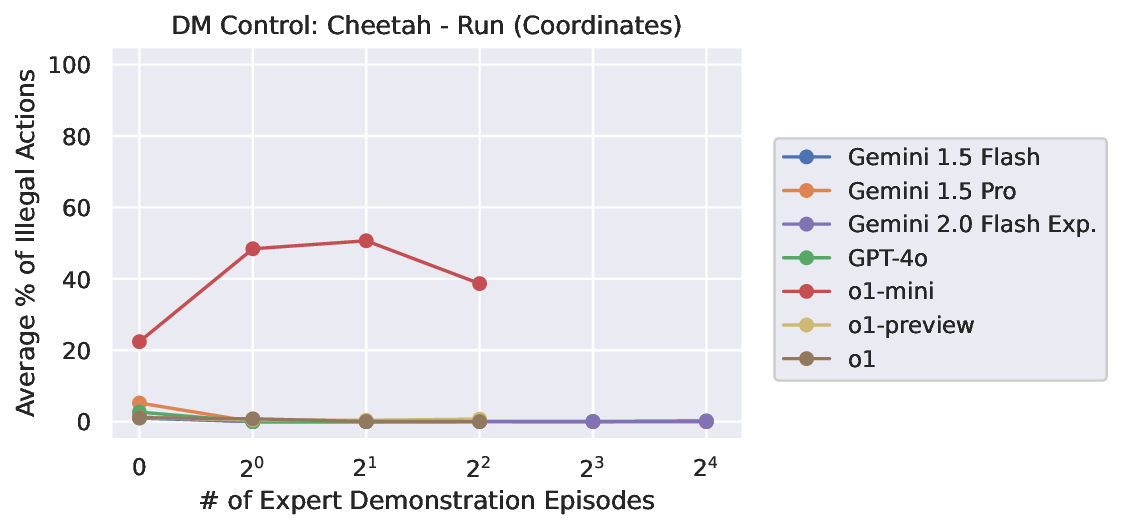}
            \end{center}
            \captionof{figure}{
                Average percentage of illegal actions per episode over the number of expert demonstrations for the cheetah run task from the DM Control suite with coordinate observations.
                All models, except \mini{}, consistently produce mostly legal actions.
            }
            \label{fig:dm-control-illegal-actions}
    \end{minipage}
\end{figure}

\begin{figure}
    \begin{subfigure}[t]{0.5\textwidth}
        \begin{center}
            \includegraphics[width=\textwidth]{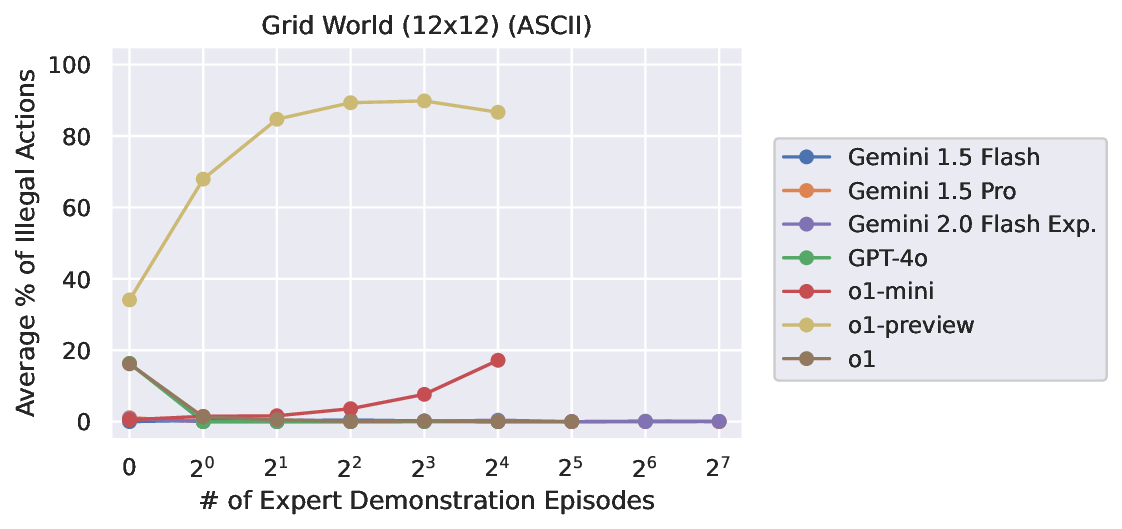}
        \end{center}
        \caption{ASCII observations}
    \end{subfigure}
    \begin{subfigure}[t]{0.5\textwidth}
        \begin{center}
            \includegraphics[width=\textwidth]{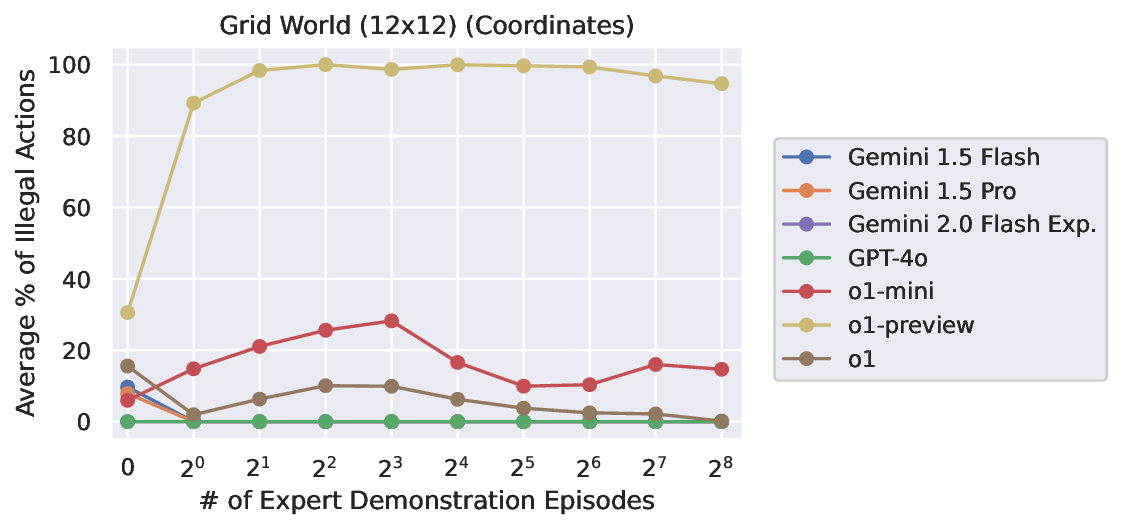}
        \end{center}
        \caption{Coordinate observations}
    \end{subfigure}
    \par\bigskip
    \begin{subfigure}[t]{0.5\textwidth}
        \begin{center}
            \includegraphics[width=\textwidth]{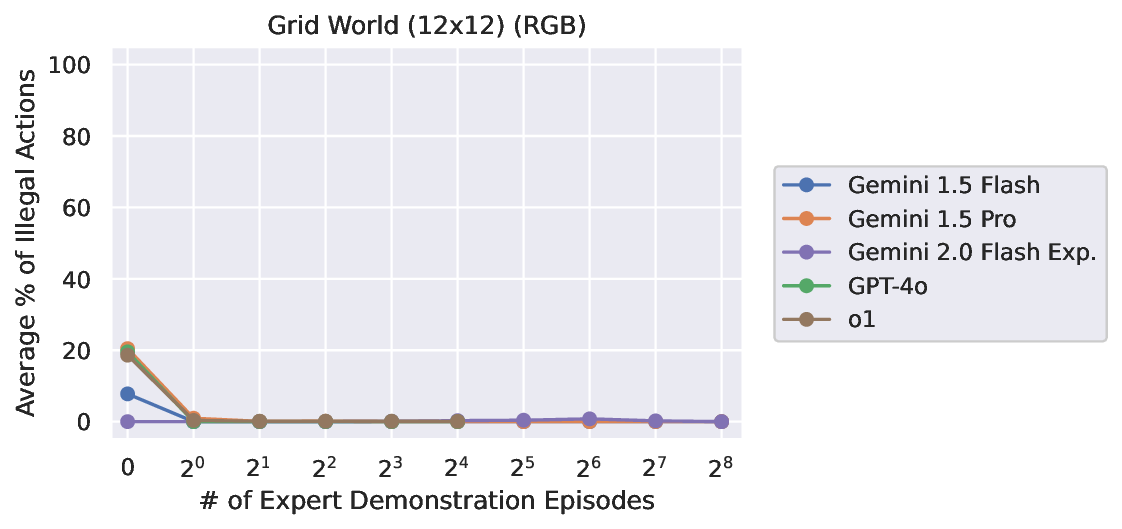}
        \end{center}
        \caption{RGB observations}
    \end{subfigure}
    \caption{
        Average percentage of illegal actions per episode over the number of expert demonstrations for all state representation formats form our grid world navigation task.
        The o1 models increasingly struggle to produce legal actions with increasing numbers of expert demonstrations in the context, with \preview{} having a very high percentage of illegal actions.
        In contrast, \gpt{}, \flash{}, and \pro{} consistently produce legal actions.
        Note that \mini{} and \preview{} are text-only models and therefore cannot process RGB image observations.
    }
    \label{fig:grid-world-illegal-actions}
\end{figure}

\begin{figure}
    \begin{subfigure}[t]{0.5\textwidth}
        \begin{center}
            \includegraphics[width=\textwidth]{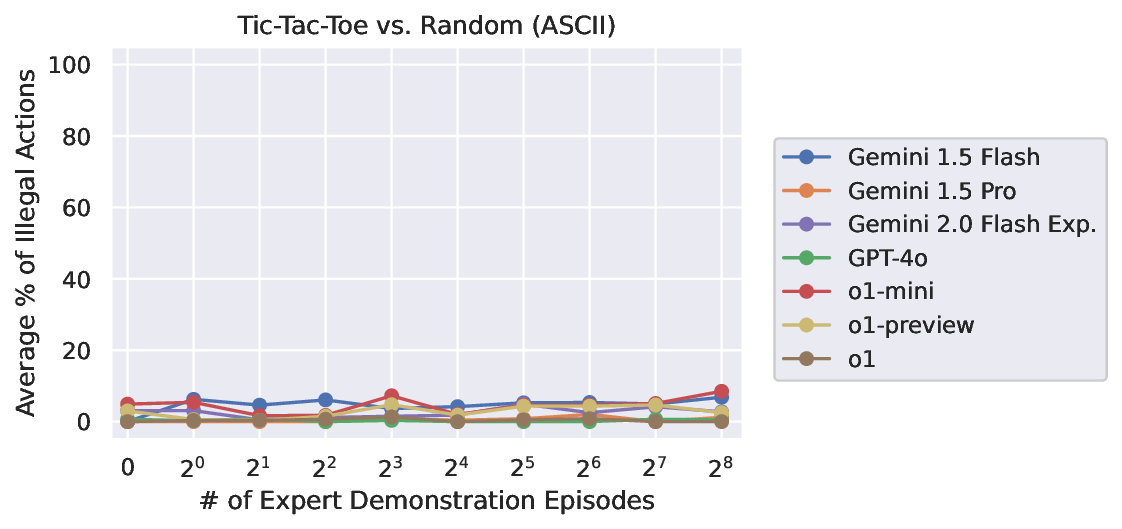}
        \end{center}
        \caption{ASCII observations}
    \end{subfigure}
    \begin{subfigure}[t]{0.5\textwidth}
        \begin{center}
            \includegraphics[width=\textwidth]{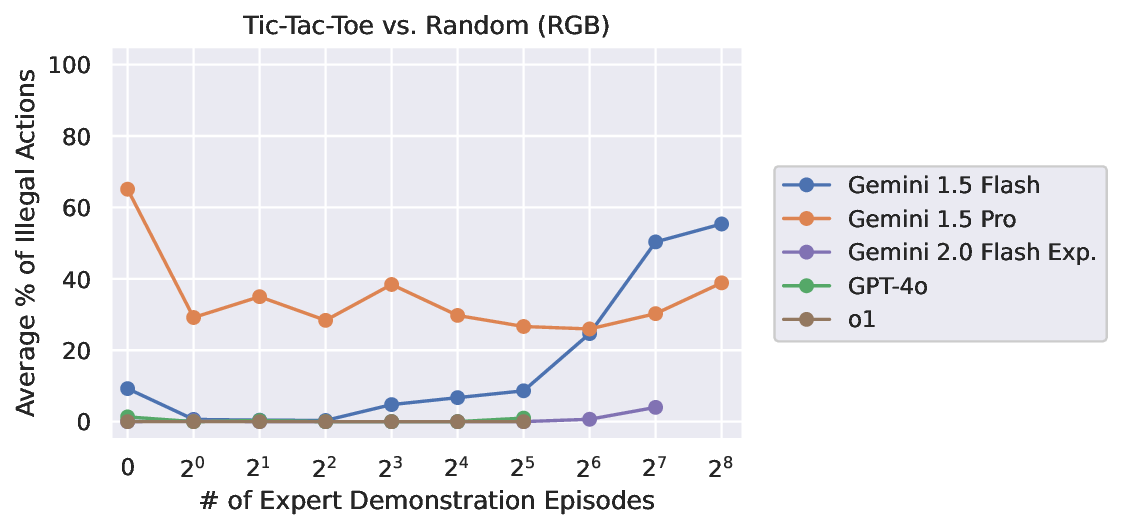}
        \end{center}
        \caption{RGB observations.}
    \end{subfigure}
    \caption{
        Average percentage of illegal actions per episode over the number of expert demonstrations for all state representation formats from our tic-tac-toe environment.
        With ASCII observations, all models mostly produce legal actions.
        In contrast, for RGB image observations, the Gemini 1.5 models struggle to produce legal actions (\pro{} independently of the number of expert demonstrations and \flash{} increasingly with the number of expert demonstrations).
        Note that \mini{} and \preview{} are text-only models and therefore cannot process RGB image observations.
    }
    \label{fig:tic-tac-toe-illegal-actions}
\end{figure}

As described in \cref{sec:methods:evaluation-protocol}, we do not terminate the evaluation episode early if a model does not produce a valid action (except on the crossword task) but instead randomly sample one of the actions that are legal in the current state and continue the evaluation with the observation produced by that action.
To differentiate illegal actions from acting randomly (over legal actions), we compute how often models actually propose an illegal action and visualize the percentage of illegal actions per episode over the number of expert demonstrations in \cref{fig:atari-illegal-actions,fig:chess-illegal-actions,fig:crossword-illegal-actions,fig:dm-control-illegal-actions,fig:grid-world-illegal-actions,fig:tic-tac-toe-illegal-actions} (analogous to \cref{fig:atari-phoenix,fig:chess,fig:crossword,fig:dm-control,fig:grid-world,fig:tic-tac-toe}).
Note that, since we perform an ablation over \emph{showing the legal moves in the prompt} (see \cref{app:additional-results:ablations}), in theory, all models could have the necessary information to sample a legal action at all times (whether we actually show this or not in our main experiments depends on whether it improves model performance in the ablations).
Nevertheless, we observe that models do produce illegal actions actions across most environments.
For example, on the grid world navigation task with coordinate observations \preview{} produces almost $100$\% illegal actions with $2$ or more expert demonstrations in the context.
Interestingly, the ability to produce legal actions also depends on the observation type: For tic-tac-toe, all models mostly produce legal actions with ASCII observations, but the Gemini 1.5 models generate up to $\sim60$\% illegal actions for RGB observations from the same environment.

\subsection{Hyperparameter Ablations}
\label{app:additional-results:ablations}

As mentioned in \cref{sec:results}, we ablate the use of chain-of-though prompting~\citep{wei2022chain} and whether or not to include the list of actions that is legal given the current observation in the prompt for each model-task combination since we want to report the best-possible performance the models can attain on our benchmark.
For completeness, we present all the ablation results in  \cref{tab:atari-ablation,tab:chess-ablation-anthropic,tab:chess-ablation-gemini,tab:chess-ablation-openai,tab:crossword-ablation,tab:dm-control-ablation,tab:grid-world-ablation-anthropic,tab:grid-world-ablation-gemini,tab:grid-world-ablation-openai,tab:tic-tac-toe-ablation}.
We highlight the best score for each model-observation pair in bold and use the corresponding hyperparameter combination for our sweeps over the number of expert demonstration episodes in \cref{sec:results:in-context-imitation-learning}.

\begin{table}
    \caption{
        Ablating the use of chain-of-thought prompting and whether or not to show legal actions in the prompt for the Phoenix game from Atari 2600.
        For this task, all models perform best without legal actions and without chain-of-thought.
    }
    \begin{center}
        \resizebox{0.55\textwidth}{!}{\begin{tabular}{@{}llllr@{}}
    \toprule
    \textbf{Model}                                                      & \textbf{Observation}      & \textbf{Legal Actions}    & \textbf{Chain-of-Thought} & \textbf{Average Score} \\
    \midrule
    \multirow[c]{4}{*}{Gemini 1.5 Flash}                                & \multirow[c]{4}{*}{RGB}   & \multirow[c]{2}{*}{False} & False                     & \textbf{42.83} \\
                                                                        &                           &                           & True                      & 6.16 \\
                                                                        &                           & \multirow[c]{2}{*}{True}  & False                     & 8.07 \\
                                                                        &                           &                           & True                      & 6.46 \\
    \midrule                                                                                                                                                   
    \multirow[c]{4}{*}{Gemini 1.5 Pro}                                  & \multirow[c]{4}{*}{RGB}   & \multirow[c]{2}{*}{False} & False                     & \textbf{21.86} \\
                                                                        &                           &                           & True                      & 8.68 \\
                                                                        &                           & \multirow[c]{2}{*}{True}  & False                     & 6.33 \\
                                                                        &                           &                           & True                      & 6.21 \\
    \midrule                                
    \multirow[c]{4}{*}{Gemini 2.0 Flash Exp.}                           & \multirow[c]{4}{*}{RGB}   & \multirow[c]{2}{*}{False} & False                     & \textbf{31.80} \\
                                                                        &                           &                           & True                      & 11.40 \\
                                                                        &                           & \multirow[c]{2}{*}{True}  & False                     & 15.20 \\
                                                                        &                           &                           & True                      & 10.20 \\
    \midrule                                
    \multirow[c]{4}{*}{GPT-4o}                                          & \multirow[c]{4}{*}{RGB}   & \multirow[c]{2}{*}{False} & False                     & \textbf{9.95} \\
                                                                        &                           &                           & True                      & 5.40 \\
                                                                        &                           & \multirow[c]{2}{*}{True}  & False                     & 6.40 \\
                                                                        &                           &                           & True                      & 5.40 \\
    \midrule                                
    \multirow[c]{4}{*}{o1}                                              & \multirow[c]{4}{*}{RGB}   & \multirow[c]{2}{*}{False} & False                     & \textbf{29.00} \\
                                                                        &                           &                           & True                      & 19.20 \\
                                                                        &                           & \multirow[c]{2}{*}{True}  & False                     & 23.00 \\
                                                                        &                           &                           & True                      & 16.40 \\
   \bottomrule
\end{tabular}
}
    \end{center}
    \label{tab:atari-ablation}
\end{table}

\begin{table}
    \caption{
        Ablating the use of chain-of-thought prompting and whether or not to show legal actions in the prompt for \claude{} on chess.
        For this task, \claude{} always profits from having the legal actions in the prompt.
        Chain-of-thought prompting improves performance on FEN and RGB observations, but not on ASCII and PGN observations.
    }
    \begin{center}
        \resizebox{0.55\textwidth}{!}{\begin{tabular}{@{}llllr@{}}
    \toprule
    \textbf{Model}                                                      & \textbf{Observation}      & \textbf{Legal Actions}    & \textbf{Chain-of-Thought} & \textbf{Average Score} \\
    \midrule
    \multirow[c]{17}{*}{\shortstack{Claude 3.5 Sonnet\\\\(2024-10-22)}} & \multirow[c]{4}{*}{ASCII} & \multirow[c]{2}{*}{False} & False                     & -0.98 \\
                                                                        &                           &                           & True                      & -0.99 \\
                                                                        &                           & \multirow[c]{2}{*}{True}  & False                     & \textbf{-0.88} \\
                                                                        &                           &                           & True                      & -0.93 \\
    \cmidrule{2-5}                                                      
                                                                        & \multirow[c]{4}{*}{FEN}   & \multirow[c]{2}{*}{False} & False                     & -0.96 \\
                                                                        &                           &                           & True                      & -0.99 \\
                                                                        &                           & \multirow[c]{2}{*}{True}  & False                     & -0.96 \\
                                                                        &                           &                           & True                      & \textbf{-0.95} \\
    \cmidrule{2-5}                                                      
                                                                        & \multirow[c]{4}{*}{PGN}   & \multirow[c]{2}{*}{False} & False                     & -0.92 \\
                                                                        &                           &                           & True                      & -0.97 \\
                                                                        &                           & \multirow[c]{2}{*}{True}  & False                     & \textbf{-0.82} \\
                                                                        &                           &                           & True                      & -0.86 \\
    \cmidrule{2-5}                                                      
                                                                        & \multirow[c]{4}{*}{RGB}   & \multirow[c]{2}{*}{False} & False                     & -0.96 \\
                                                                        &                           &                           & True                      & -0.99 \\
                                                                        &                           & \multirow[c]{2}{*}{True}  & False                     & -0.94 \\
                                                                        &                           &                           & True                      & \textbf{-0.91} \\
   \bottomrule
\end{tabular}
}
    \end{center}
    \label{tab:chess-ablation-anthropic}
\end{table}

\begin{table}
    \caption{
        Ablating the use of chain-of-thought prompting and whether or not to show legal actions in the prompt for \flash{}, \pro{}, and \expflash{} on chess.
        For this task, the models almost always profit from having the legal actions in the prompt.
        Chain-of-thought prompting sometimes improves performance, but not across the board.
    }
    \begin{center}
        \resizebox{0.55\textwidth}{!}{\begin{tabular}{@{}llllr@{}}
    \toprule
    \textbf{Model}                              & \textbf{Observation}      & \textbf{Legal Actions}    & \textbf{Chain-of-Thought} & \textbf{Average Score} \\
    \midrule    
    \multirow[c]{17}{*}{Gemini 1.5 Flash}       & \multirow[c]{4}{*}{ASCII} & \multirow[c]{2}{*}{False} & False                     &  -1.00 \\
                                                &                           &                           & True                      &  -1.00 \\
                                                &                           & \multirow[c]{2}{*}{True}  & False                     &  \textbf{-0.98} \\
                                                &                           &                           & True                      &  -0.99 \\
    \cmidrule{2-5}                                                                                                                      
                                                & \multirow[c]{4}{*}{FEN}   & \multirow[c]{2}{*}{False} & False                     &  -0.99 \\
                                                &                           &                           & True                      &  -0.99 \\
                                                &                           & \multirow[c]{2}{*}{True}  & False                     &  -0.99 \\
                                                &                           &                           & True                      &  \textbf{-0.98} \\
    \cmidrule{2-5}                                                                                                                      
                                                & \multirow[c]{4}{*}{PGN}   & \multirow[c]{2}{*}{False} & False                     &  -0.98 \\
                                                &                           &                           & True                      &  -0.99 \\
                                                &                           & \multirow[c]{2}{*}{True}  & False                     &  \textbf{-0.96} \\
                                                &                           &                           & True                      &  -0.97 \\
    \cmidrule{2-5}                                                                                                                      
                                                & \multirow[c]{4}{*}{RGB}   & \multirow[c]{2}{*}{False} & False                     &  -1.00 \\
                                                &                           &                           & True                      &  -1.00 \\
                                                &                           & \multirow[c]{2}{*}{True}  & False                     &  \textbf{-0.97} \\
                                                &                           &                           & True                      &  -1.00 \\
    \midrule    
    \multirow[c]{17}{*}{Gemini 1.5 Pro}         & \multirow[c]{4}{*}{ASCII} & \multirow[c]{2}{*}{False} & False                     & -1.00 \\
                                                &                           &                           & True                      & -0.99 \\
                                                &                           & \multirow[c]{2}{*}{True}  & False                     & \textbf{-0.96} \\
                                                &                           &                           & True                      & -0.97 \\
    \cmidrule{2-5}                                                                                                                       
                                                & \multirow[c]{4}{*}{FEN}   & \multirow[c]{2}{*}{False} & False                     & -1.00 \\
                                                &                           &                           & True                      & -1.00 \\
                                                &                           & \multirow[c]{2}{*}{True}  & False                     & \textbf{-0.95} \\
                                                &                           &                           & True                      & -0.97 \\
    \cmidrule{2-5}                                                                                                                       
                                                & \multirow[c]{4}{*}{PGN}   & \multirow[c]{2}{*}{False} & False                     & -0.93 \\
                                                &                           &                           & True                      & -0.94 \\
                                                &                           & \multirow[c]{2}{*}{True}  & False                     & \textbf{-0.90} \\
                                                &                           &                           & True                      & -0.97 \\
    \cmidrule{2-5}                                                                                                                       
                                                & \multirow[c]{4}{*}{RGB}   & \multirow[c]{2}{*}{False} & False                     & -1.00 \\
                                                &                           &                           & True                      & -0.99 \\
                                                &                           & \multirow[c]{2}{*}{True}  & False                     & -0.99 \\
                                                &                           &                           & True                      & \textbf{-0.97} \\
    \midrule
    \multirow[c]{17}{*}{Gemini 2.0 Flash Exp.}  & \multirow[c]{4}{*}{ASCII} & \multirow[c]{2}{*}{False} & False                     & -1.00 \\
                                                &                           &                           & True                      & -0.99 \\
                                                &                           & \multirow[c]{2}{*}{True}  & False                     & \textbf{-0.93} \\
                                                &                           &                           & True                      & -0.98 \\
    \cmidrule{2-5}                                                                                                                     
                                                & \multirow[c]{4}{*}{FEN}   & \multirow[c]{2}{*}{False} & False                     & -0.95 \\
                                                &                           &                           & True                      & -0.99 \\
                                                &                           & \multirow[c]{2}{*}{True}  & False                     & \textbf{-0.91} \\
                                                &                           &                           & True                      & -0.97 \\
    \cmidrule{2-5}                                                                                                                     
                                                & \multirow[c]{4}{*}{PGN}   & \multirow[c]{2}{*}{False} & False                     & \textbf{-0.87} \\
                                                &                           &                           & True                      & -0.96 \\
                                                &                           & \multirow[c]{2}{*}{True}  & False                     & -0.91 \\
                                                &                           &                           & True                      & -0.96 \\
    \cmidrule{2-5}                                                                                                                     
                                                & \multirow[c]{4}{*}{RGB}   & \multirow[c]{2}{*}{False} & False                     & -0.98 \\
                                                &                           &                           & True                      & -0.99 \\
                                                &                           & \multirow[c]{2}{*}{True}  & False                     & \textbf{-0.95} \\
                                                &                           &                           & True                      & \textbf{-0.95} \\
   \bottomrule
\end{tabular}
}
    \end{center}
    \label{tab:chess-ablation-gemini}
\end{table}

\begin{table}
    \caption{
        Ablating the use of chain-of-thought prompting and whether or not to show legal actions in the prompt for \gpt{}, \mini{}, \preview{}, and \oone{} on chess.
        For this task, the models always profit from having the legal actions in the prompt.
        Chain-of-thought prompting sometimes improves performance, but not across the board.
    }
    \begin{center}
        \resizebox{0.55\textwidth}{!}{\begin{tabular}{@{}llllr@{}}
    \toprule
    \textbf{Model}                  & \textbf{Observation}      & \textbf{Legal Actions}    & \textbf{Chain-of-Thought} & \textbf{Average Score} \\
    \midrule    
    \multirow[c]{17}{*}{GPT-4o}     & \multirow[c]{4}{*}{ASCII} & \multirow[c]{2}{*}{False} & False                     & -1.00 \\
                                    &                           &                           & True                      & -1.00 \\
                                    &                           & \multirow[c]{2}{*}{True}  & False                     & -0.94 \\
                                    &                           &                           & True                      & \textbf{-0.92} \\
    \cmidrule{2-5}                                                                                                         
                                    & \multirow[c]{4}{*}{FEN}   & \multirow[c]{2}{*}{False} & False                     & -0.94 \\
                                    &                           &                           & True                      & -0.98 \\
                                    &                           & \multirow[c]{2}{*}{True}  & False                     & -0.98 \\
                                    &                           &                           & True                      & \textbf{-0.90} \\
    \cmidrule{2-5}                                                                                                         
                                    & \multirow[c]{4}{*}{PGN}   & \multirow[c]{2}{*}{False} & False                     & \textbf{-0.63} \\
                                    &                           &                           & True                      & -0.97 \\
                                    &                           & \multirow[c]{2}{*}{True}  & False                     & -0.88 \\
                                    &                           &                           & True                      & -0.93 \\
    \cmidrule{2-5}                                                                                                         
                                    & \multirow[c]{4}{*}{RGB}   & \multirow[c]{2}{*}{False} & False                     & -0.90 \\
                                    &                           &                           & True                      & -0.99 \\
                                    &                           & \multirow[c]{2}{*}{True}  & False                     & \textbf{-0.84} \\
                                    &                           &                           & True                      & -0.91 \\
    \midrule                                                                                                               
    \multirow[c]{13}{*}{o1-mini}    & \multirow[c]{4}{*}{ASCII} & \multirow[c]{2}{*}{False} & False                     & -0.99 \\
                                    &                           &                           & True                      & -0.99 \\
                                    &                           & \multirow[c]{2}{*}{True}  & False                     & \textbf{-0.96} \\
                                    &                           &                           & True                      & \textbf{-0.96} \\
    \cmidrule{2-5}                                                                                                         
                                    & \multirow[c]{4}{*}{FEN}   & \multirow[c]{2}{*}{False} & False                     & -1.00 \\
                                    &                           &                           & True                      & -1.00 \\
                                    &                           & \multirow[c]{2}{*}{True}  & False                     & \textbf{-0.89} \\
                                    &                           &                           & True                      & -0.92 \\
    \cmidrule{2-5}                                                                                                         
                                    & \multirow[c]{4}{*}{PGN}   & \multirow[c]{2}{*}{False} & False                     & -0.99 \\
                                    &                           &                           & True                      & -0.97 \\
                                    &                           & \multirow[c]{2}{*}{True}  & False                     & -0.84 \\
                                    &                           &                           & True                      & \textbf{-0.83} \\
    \midrule                                                                                                               
    \multirow[c]{13}{*}{o1-preview} & \multirow[c]{4}{*}{ASCII} & \multirow[c]{2}{*}{False} & False                     & -0.99 \\
                                    &                           &                           & True                      & -0.99 \\
                                    &                           & \multirow[c]{2}{*}{True}  & False                     & \textbf{-0.93} \\
                                    &                           &                           & True                      & -0.94 \\
    \cmidrule{2-5}                                                                                                         
                                    & \multirow[c]{4}{*}{FEN}   & \multirow[c]{2}{*}{False} & False                     & -0.99 \\
                                    &                           &                           & True                      & -0.99 \\
                                    &                           & \multirow[c]{2}{*}{True}  & False                     & -0.95 \\
                                    &                           &                           & True                      & \textbf{-0.89} \\
    \cmidrule{2-5}                                                                                                         
                                    & \multirow[c]{4}{*}{PGN}   & \multirow[c]{2}{*}{False} & False                     & -0.98 \\
                                    &                           &                           & True                      & -0.98 \\
                                    &                           & \multirow[c]{2}{*}{True}  & False                     & \textbf{-0.95} \\
                                    &                           &                           & True                      & -0.90 \\
    \midrule    
    \multirow[c]{17}{*}{o1}         & \multirow[c]{4}{*}{ASCII} & \multirow[c]{2}{*}{False} & False                     & -0.98 \\
                                    &                           &                           & True                      & -0.98 \\
                                    &                           & \multirow[c]{2}{*}{True}  & False                     & \textbf{-0.85} \\
                                    &                           &                           & True                      & -0.87 \\
    \cmidrule{2-5}
                                    & \multirow[c]{4}{*}{FEN}   & \multirow[c]{2}{*}{False} & False                     & -0.95 \\
                                    &                           &                           & True                      & -0.89 \\
                                    &                           & \multirow[c]{2}{*}{True}  & False                     & \textbf{-0.82} \\
                                    &                           &                           & True                      & -0.87 \\
    \cmidrule{2-5}
                                    & \multirow[c]{4}{*}{PGN}   & \multirow[c]{2}{*}{False} & False                     & -0.80 \\
                                    &                           &                           & True                      & -0.84 \\
                                    &                           & \multirow[c]{2}{*}{True}  & False                     & -0.85 \\
                                    &                           &                           & True                      & \textbf{-0.79} \\
    \cmidrule{2-5}
                                    & \multirow[c]{4}{*}{RGB}   & \multirow[c]{2}{*}{False} & False                     & -0.95 \\
                                    &                           &                           & True                      & -0.97 \\
                                    &                           & \multirow[c]{2}{*}{True}  & False                     & \textbf{-0.82} \\
                                    &                           &                           & True                      & -0.88 \\
   \bottomrule
\end{tabular}
}
    \end{center}
    \label{tab:chess-ablation-openai}
\end{table}

\begin{table}
    \caption{
        Ablating the use of chain-of-thought prompting for the task of solving crosswords.
        Chain-of-thought prompting sometimes improves performance, but not across all models.
        For this task, showing legal actions is infeasible (as it would include all possible (\ie thousands of) words of the correct length for every slot), so we do not ablate it.
    }
    \begin{center}
        \resizebox{0.55\textwidth}{!}{\begin{tabular}{@{}llllr@{}}
    \toprule
    \textbf{Model}                                                      & \textbf{Observation}      & \textbf{Legal Actions}    & \textbf{Chain-of-Thought} & \textbf{Average Score} \\
    \midrule
    \multirow[c]{2}{*}{\shortstack{Claude 3.5 Sonnet\\\\(2024-10-22)}}  & \multirow[c]{2}{*}{ASCII} & \multirow[c]{2}{*}{False} & False                     & 6.73 \\
                                                                        &                           &                           & True                      & \textbf{6.81} \\
    \midrule
    \multirow[c]{2}{*}{Gemini 1.5 Flash}                                & \multirow[c]{2}{*}{ASCII} & \multirow[c]{2}{*}{False} & False                     & \textbf{1.99} \\
                                                                        &                           &                           & True                      & 1.84 \\
    \midrule                            
    \multirow[c]{2}{*}{Gemini 1.5 Pro}                                  & \multirow[c]{2}{*}{ASCII} & \multirow[c]{2}{*}{False} & False                     & 3.15 \\
                                                                        &                           &                           & True                      & \textbf{5.50} \\
    \midrule
    \multirow[c]{2}{*}{Gemini 2.0 Flash Exp.}                           & \multirow[c]{2}{*}{ASCII} & \multirow[c]{2}{*}{False} & False                     & 3.70 \\
                                                                        &                           &                           & True                      & \textbf{3.98} \\
    \midrule                                
    \multirow[c]{2}{*}{GPT-4o}                                          & \multirow[c]{2}{*}{ASCII} & \multirow[c]{2}{*}{False} & False                     & 5.76 \\
                                                                        &                           &                           & True                      & \textbf{6.67} \\
    \midrule                                
    \multirow[c]{2}{*}{o1-mini}                                         & \multirow[c]{2}{*}{ASCII} & \multirow[c]{2}{*}{False} & False                     & \textbf{4.95} \\
                                                                        &                           &                           & True                      & 4.82 \\
    \midrule                                    
    \multirow[c]{2}{*}{o1-preview}                                      & \multirow[c]{2}{*}{ASCII} & \multirow[c]{2}{*}{False} & False                     & 9.57 \\
                                                                        &                           &                           & True                      & \textbf{9.79} \\
    \midrule                                    
    \multirow[c]{2}{*}{o1}                                              & \multirow[c]{2}{*}{ASCII} & \multirow[c]{2}{*}{False} & False                     & 9.55 \\
                                                                        &                           &                           & True                      & \textbf{9.56} \\
    \bottomrule
\end{tabular}
}
    \end{center}
    \label{tab:crossword-ablation}
\end{table}

\begin{table}
    \caption{
        Ablating the use of chain-of-thought prompting and whether or not to show legal actions in the prompt for the task of simulating a cheetah from the DM Control Suite.
        For this task, most models do not benefit from chain-of-thought prompting or showing the legal actions in the prompt (note that there are infinitely many legal actions which we represent with the following string: \mintinline[mathescape,fontsize=\ificlr\scriptsize\else\footnotesize\fi]{python}{'A comma-separated list (enclosed by square brackets) of 6 values between -1 and 1.'}).
    }
    \begin{center}
        \resizebox{0.55\textwidth}{!}{\begin{tabular}{@{}llllr@{}}
    \toprule
    \textbf{Model}                                                      & \textbf{Observation}              & \textbf{Legal Actions}    & \textbf{Chain-of-Thought} & \textbf{Average Score} \\
    \midrule
    \multirow[c]{4}{*}{\shortstack{Claude 3.5 Sonnet\\\\(2024-10-22)}}  & \multirow[c]{4}{*}{Coordinates}   & \multirow[c]{2}{*}{False} & False                     & 4.70 \\
                                                                        &                                   &                           & True                      & 3.53 \\
                                                                        &                                   & \multirow[c]{2}{*}{True}  & False                     & \textbf{10.20} \\
                                                                        &                                   &                           & True                      & 1.33 \\
    \midrule
    \multirow[c]{4}{*}{Gemini 1.5 Flash}                                & \multirow[c]{4}{*}{Coordinates}   & \multirow[c]{2}{*}{False} & False                     & \textbf{3.93} \\
                                                                        &                                   &                           & True                      & 1.89 \\
                                                                        &                                   & \multirow[c]{2}{*}{True}  & False                     & 0.39 \\
                                                                        &                                   &                           & True                      & 2.31 \\
    \midrule                            
    \multirow[c]{4}{*}{Gemini 1.5 Pro}                                  & \multirow[c]{4}{*}{Coordinates}   & \multirow[c]{2}{*}{False} & False                     & \textbf{18.00} \\
                                                                        &                                   &                           & True                      & 1.26 \\
                                                                        &                                   & \multirow[c]{2}{*}{True}  & False                     & 7.69 \\
                                                                        &                                   &                           & True                      & 2.44 \\
    \midrule
    \multirow[c]{4}{*}{Gemini 2.0 Flash Exp.}                           & \multirow[c]{4}{*}{Coordinates}   & \multirow[c]{2}{*}{False} & False                     & \textbf{15.96} \\
                                                                        &                                   &                           & True                      & 1.91  \\
                                                                        &                                   & \multirow[c]{2}{*}{True}  & False                     & 13.53 \\
                                                                        &                                   &                           & True                      & 1.85  \\
    \midrule                                
    \multirow[c]{4}{*}{GPT-4o}                                          & \multirow[c]{4}{*}{Coordinates}   & \multirow[c]{2}{*}{False} & False                     & \textbf{11.03} \\
                                                                        &                                   &                           & True                      & 0.72 \\
                                                                        &                                   & \multirow[c]{2}{*}{True}  & False                     & 1.00 \\
                                                                        &                                   &                           & True                      & 0.92 \\
    \midrule                                        
    \multirow[c]{4}{*}{o1-mini}                                         & \multirow[c]{4}{*}{Coordinates}   & \multirow[c]{2}{*}{False} & False                     & \textbf{0.80} \\
                                                                        &                                   &                           & True                      & 0.46 \\
                                                                        &                                   & \multirow[c]{2}{*}{True}  & False                     & 0.78 \\
                                                                        &                                   &                           & True                      & 0.44 \\
    \midrule                                
    \multirow[c]{4}{*}{o1-preview}                                      & \multirow[c]{4}{*}{Coordinates}   & \multirow[c]{2}{*}{False} & False                     & 3.16 \\
                                                                        &                                   &                           & True                      & 1.04 \\
                                                                        &                                   & \multirow[c]{2}{*}{True}  & False                     & \textbf{4.08} \\
                                                                        &                                   &                           & True                      & 1.88 \\
    \midrule                                
    \multirow[c]{4}{*}{o1}                                              & \multirow[c]{4}{*}{Coordinates}   & \multirow[c]{2}{*}{False} & False                     & 2.08 \\
                                                                        &                                   &                           & True                      & \textbf{2.14} \\
                                                                        &                                   & \multirow[c]{2}{*}{True}  & False                     & 2.06 \\
                                                                        &                                   &                           & True                      & 1.84 \\
   \bottomrule
\end{tabular}
}
    \end{center}
    \label{tab:dm-control-ablation}
\end{table}

\begin{table}
    \caption{
        Ablating the use of chain-of-thought prompting and whether or not to show legal actions in the prompt for \claude{} on the grid world navigation task.
        For ASCII and coordinate observations, the performance is largely independent of the ablated settings.
        For RGB images, showing the legal actions and using chain-of-thought prompting performs best.
    }
    \begin{center}
        \resizebox{0.55\textwidth}{!}{\begin{tabular}{@{}llllr@{}}
    \toprule
    \textbf{Model}                                                      & \textbf{Observation}              & \textbf{Legal Actions}    & \textbf{Chain-of-Thought} & \textbf{Average Score} \\
    \midrule        
    \multirow[c]{13}{*}{\shortstack{Claude 3.5 Sonnet\\\\(2024-10-22)}} & \multirow[c]{4}{*}{ASCII}         & \multirow[c]{2}{*}{False} & False                     & \textbf{1.00} \\
                                                                        &                                   &                           & True                      & \textbf{1.00} \\
                                                                        &                                   & \multirow[c]{2}{*}{True}  & False                     & \textbf{1.00} \\
                                                                        &                                   &                           & True                      & \textbf{1.00} \\
    \cmidrule{2-5}                                                                                                                                                     
                                                                        & \multirow[c]{4}{*}{Coordinates}   & \multirow[c]{2}{*}{False} & False                     & 0.96 \\
                                                                        &                                   &                           & True                      & \textbf{1.00} \\
                                                                        &                                   & \multirow[c]{2}{*}{True}  & False                     & 0.99 \\
                                                                        &                                   &                           & True                      & 0.99 \\
    \cmidrule{2-5}                                                                                                                                                     
                                                                        & \multirow[c]{4}{*}{RGB}           & \multirow[c]{2}{*}{False} & False                     & 0.19 \\
                                                                        &                                   &                           & True                      & 0.40 \\
                                                                        &                                   & \multirow[c]{2}{*}{True}  & False                     & 0.21 \\
                                                                        &                                   &                           & True                      & \textbf{0.53} \\
   \bottomrule
\end{tabular}
}
    \end{center}
    \label{tab:grid-world-ablation-anthropic}
\end{table}

\begin{table}
    \caption{
        Ablating the use of chain-of-thought prompting and whether or not to show legal actions in the prompt for \flash{}, \pro{}, and \expflash{} on the grid world navigation task.
        The best-performing setting is model and observation dependent.
    }
    \begin{center}
        \resizebox{0.55\textwidth}{!}{\begin{tabular}{@{}llllr@{}}
    \toprule
    \textbf{Model}                              & \textbf{Observation}              & \textbf{Legal Actions}    & \textbf{Chain-of-Thought} & \textbf{Average Score} \\
    \midrule
    \multirow[c]{13}{*}{Gemini 1.5 Flash}       & \multirow[c]{4}{*}{ASCII}         & \multirow[c]{2}{*}{False} & False                     & 0.09 \\
                                                &                                   &                           & True                      & 0.08 \\
                                                &                                   & \multirow[c]{2}{*}{True}  & False                     & 0.11 \\
                                                &                                   &                           & True                      & \textbf{0.15} \\
    \cmidrule{2-5}                                                                                                                             
                                                & \multirow[c]{4}{*}{Coordinates}   & \multirow[c]{2}{*}{False} & False                     & \textbf{0.21} \\
                                                &                                   &                           & True                      & 0.10 \\
                                                &                                   & \multirow[c]{2}{*}{True}  & False                     & 0.14 \\
                                                &                                   &                           & True                      & 0.08 \\
    \cmidrule{2-5}                                                                                                                             
                                                & \multirow[c]{4}{*}{RGB}           & \multirow[c]{2}{*}{False} & False                     & \textbf{0.16} \\
                                                &                                   &                           & True                      & 0.12 \\
                                                &                                   & \multirow[c]{2}{*}{True}  & False                     & 0.14 \\
                                                &                                   &                           & True                      & 0.12 \\
    \midrule
    \multirow[c]{13}{*}{Gemini 1.5 Pro}         & \multirow[c]{4}{*}{ASCII}         & \multirow[c]{2}{*}{False} & False                     & \textbf{0.51} \\
                                                &                                   &                           & True                      & 0.25 \\
                                                &                                   & \multirow[c]{2}{*}{True}  & False                     & \textbf{0.51} \\
                                                &                                   &                           & True                      & 0.36 \\
    \cmidrule{2-5}                                                                                                                             
                                                & \multirow[c]{4}{*}{Coordinates}   & \multirow[c]{2}{*}{False} & False                     & \textbf{0.27} \\
                                                &                                   &                           & True                      & 0.13 \\
                                                &                                   & \multirow[c]{2}{*}{True}  & False                     & 0.16 \\
                                                &                                   &                           & True                      & 0.12 \\
    \cmidrule{2-5}                                                                                                                             
                                                & \multirow[c]{4}{*}{RGB}           & \multirow[c]{2}{*}{False} & False                     & 0.17 \\
                                                &                                   &                           & True                      & \textbf{0.25} \\
                                                &                                   & \multirow[c]{2}{*}{True}  & False                     & 0.24 \\
                                                &                                   &                           & True                      & 0.24 \\
    \midrule
    \multirow[c]{13}{*}{Gemini 2.0 Flash Exp.}  & \multirow[c]{4}{*}{ASCII}         & \multirow[c]{2}{*}{False} & False                     & 0.27 \\
                                                &                                   &                           & True                      & \textbf{0.41} \\
                                                &                                   & \multirow[c]{2}{*}{True}  & False                     & 0.25 \\
                                                &                                   &                           & True                      & 0.39 \\
    \cmidrule{2-5}                                                                                                                             
                                                & \multirow[c]{4}{*}{Coordinates}   & \multirow[c]{2}{*}{False} & False                     & 0.22 \\
                                                &                                   &                           & True                      & 0.22 \\
                                                &                                   & \multirow[c]{2}{*}{True}  & False                     & \textbf{0.30} \\
                                                &                                   &                           & True                      & 0.16 \\
    \cmidrule{2-5}                                                                                                                             
                                                & \multirow[c]{4}{*}{RGB}           & \multirow[c]{2}{*}{False} & False                     & 0.18 \\
                                                &                                   &                           & True                      & 0.37 \\
                                                &                                   & \multirow[c]{2}{*}{True}  & False                     & 0.23 \\
                                                &                                   &                           & True                      & \textbf{0.43} \\
                                            
   \bottomrule
\end{tabular}
}
    \end{center}
    \label{tab:grid-world-ablation-gemini}
\end{table}

\begin{table}
    \caption{
        Ablating the use of chain-of-thought prompting and whether or not to show legal actions in the prompt for \gpt{}, \mini{}, \preview{}, and \oone{} on the grid world navigation task.
        The best-performing setting is model and observation dependent.
        Note that \mini{} and \preview{} are text-only and, therefore, cannot process RGB observations.
    }
    \begin{center}
        \resizebox{0.55\textwidth}{!}{\begin{tabular}{@{}llllr@{}}
    \toprule
    \textbf{Model}                  & \textbf{Observation}              & \textbf{Legal Actions}    & \textbf{Chain-of-Thought} & \textbf{Average Score} \\
    \midrule            
    \multirow[c]{13}{*}{GPT-4o}     & \multirow[c]{4}{*}{ASCII}         & \multirow[c]{2}{*}{False} & False                     & 0.56 \\
                                    &                                   &                           & True                      & \textbf{0.57} \\
                                    &                                   & \multirow[c]{2}{*}{True}  & False                     & 0.50 \\
                                    &                                   &                           & True                      & 0.49 \\
    \cmidrule{2-5}                                                                                                                 
                                    & \multirow[c]{4}{*}{Coordinates}   & \multirow[c]{2}{*}{False} & False                     & 0.69 \\
                                    &                                   &                           & True                      & 0.73 \\
                                    &                                   & \multirow[c]{2}{*}{True}  & False                     & \textbf{0.76} \\
                                    &                                   &                           & True                      & 0.72 \\
    \cmidrule{2-5}                                                                                                                 
                                    & \multirow[c]{4}{*}{RGB}           & \multirow[c]{2}{*}{False} & False                     & 0.52 \\
                                    &                                   &                           & True                      & \textbf{0.70} \\
                                    &                                   & \multirow[c]{2}{*}{True}  & False                     & 0.56 \\
                                    &                                   &                           & True                      & 0.61 \\
    \midrule        
    \multirow[c]{8}{*}{o1-mini}     & \multirow[c]{4}{*}{ASCII}         & \multirow[c]{2}{*}{False} & False                     & 0.20 \\
                                    &                                   &                           & True                      & 0.22 \\
                                    &                                   & \multirow[c]{2}{*}{True}  & False                     & 0.21 \\
                                    &                                   &                           & True                      & \textbf{0.32} \\
    \cmidrule{2-5}                                                                                                                 
                                    & \multirow[c]{4}{*}{Coordinates}   & \multirow[c]{2}{*}{False} & False                     & \textbf{0.92} \\
                                    &                                   &                           & True                      & 0.79 \\
                                    &                                   & \multirow[c]{2}{*}{True}  & False                     & 0.87 \\
                                    &                                   &                           & True                      & 0.80 \\
    \midrule        
    \multirow[c]{8}{*}{o1-preview}  & \multirow[c]{4}{*}{ASCII}         & \multirow[c]{2}{*}{False} & False                     & 0.94 \\
                                    &                                   &                           & True                      & \textbf{0.98} \\
                                    &                                   & \multirow[c]{2}{*}{True}  & False                     & 0.97 \\
                                    &                                   &                           & True                      & 0.97 \\
    \cmidrule{2-5}                                                                                                                 
                                    & \multirow[c]{4}{*}{Coordinates}   & \multirow[c]{2}{*}{False} & False                     & 0.99 \\
                                    &                                   &                           & True                      & \textbf{1.00} \\
                                    &                                   & \multirow[c]{2}{*}{True}  & False                     & 0.99 \\
                                    &                                   &                           & True                      & \textbf{1.00} \\
    \midrule        
    \multirow[c]{8}{*}{o1}          & \multirow[c]{4}{*}{ASCII}         & \multirow[c]{2}{*}{False} & False                     & \textbf{1.00} \\
                                    &                                   &                           & True                      & \textbf{1.00} \\
                                    &                                   & \multirow[c]{2}{*}{True}  & False                     & \textbf{1.00} \\
                                    &                                   &                           & True                      & \textbf{1.00} \\
    \cmidrule{2-5}                                                                                                                 
                                    & \multirow[c]{4}{*}{Coordinates}   & \multirow[c]{2}{*}{False} & False                     & 0.99 \\
                                    &                                   &                           & True                      & \textbf{1.00} \\
                                    &                                   & \multirow[c]{2}{*}{True}  & False                     & \textbf{1.00} \\
                                    &                                   &                           & True                      & \textbf{1.00} \\
    \cmidrule{2-5}                                                                                                                  
                                    & \multirow[c]{4}{*}{RGB}           & \multirow[c]{2}{*}{False} & False                     & 0.81 \\
                                    &                                   &                           & True                      & \textbf{0.86} \\
                                    &                                   & \multirow[c]{2}{*}{True}  & False                     & 0.84 \\
                                    &                                   &                           & True                      & 0.84 \\
   \bottomrule
\end{tabular}
}
    \end{center}
    \label{tab:grid-world-ablation-openai}
\end{table}

\begin{table}
    \caption{
        Ablating the use of chain-of-thought prompting and whether or not to show legal actions in the prompt for the game of tic-tac-toe.
        Across almost all models and observations types, using chain-of-thought prompting and showing the legal actions achieves the highest performance -- often by a large margin.
        Note that \mini{} and \preview{} are text-only and, therefore, cannot process RGB observations.
    }
    \begin{center}
        \resizebox{0.55\textwidth}{!}{\begin{tabular}{@{}llllr@{}}
    \toprule
    \textbf{Model}                                                      & \textbf{Observation}      & \textbf{Legal Actions}    & \textbf{Chain-of-Thought} & \textbf{Average Score} \\
    \midrule
    \multirow[c]{8}{*}{\shortstack{Claude 3.5 Sonnet\\\\(2024-10-22)}}  & \multirow[c]{4}{*}{ASCII} & \multirow[c]{2}{*}{False} & False                     & 0.10 \\
                                                                        &                           &                           & True                      & 0.19 \\
                                                                        &                           & \multirow[c]{2}{*}{True}  & False                     & 0.14 \\
                                                                        &                           &                           & True                      & \textbf{0.34} \\
    \cmidrule{2-5}                                                                                                                                             
                                                                        & \multirow[c]{4}{*}{RGB}   & \multirow[c]{2}{*}{False} & False                     & -0.04 \\
                                                                        &                           &                           & True                      & 0.16 \\
                                                                        &                           & \multirow[c]{2}{*}{True}  & False                     & \textbf{0.20} \\
                                                                        &                           &                           & True                      & 0.19 \\
    \midrule                                                                                                                                                   
    \multirow[c]{8}{*}{Gemini 1.5 Flash}                                & \multirow[c]{4}{*}{ASCII} & \multirow[c]{2}{*}{False} & False                     & -0.12 \\
                                                                        &                           &                           & True                      & -0.15 \\
                                                                        &                           & \multirow[c]{2}{*}{True}  & False                     & -0.10 \\
                                                                        &                           &                           & True                      & \textbf{0.00} \\
    \cmidrule{2-5}                                                                                                                                             
                                                                        & \multirow[c]{4}{*}{RGB}   & \multirow[c]{2}{*}{False} & False                     & 0.07 \\
                                                                        &                           &                           & True                      & 0.07 \\
                                                                        &                           & \multirow[c]{2}{*}{True}  & False                     & -0.05 \\
                                                                        &                           &                           & True                      & \textbf{0.10} \\
    \midrule                                                                                                                                                   
    \multirow[c]{8}{*}{Gemini 1.5 Pro}                                  & \multirow[c]{4}{*}{ASCII} & \multirow[c]{2}{*}{False} & False                     & -0.08 \\
                                                                        &                           &                           & True                      & -0.07 \\
                                                                        &                           & \multirow[c]{2}{*}{True}  & False                     & \textbf{-0.02} \\
                                                                        &                           &                           & True                      & -0.05 \\
    \cmidrule{2-5}                                                                                                                                             
                                                                        & \multirow[c]{4}{*}{RGB}   & \multirow[c]{2}{*}{False} & False                     & \textbf{0.08} \\
                                                                        &                           &                           & True                      & 0.06 \\
                                                                        &                           & \multirow[c]{2}{*}{True}  & False                     & -0.06 \\
                                                                        &                           &                           & True                      & 0.01 \\
    \midrule                                                                                                                                                   
    \multirow[c]{8}{*}{Gemini 2.0 Flash Exp.}                           & \multirow[c]{4}{*}{ASCII} & \multirow[c]{2}{*}{False} & False                     & 0.06 \\
                                                                        &                           &                           & True                      & 0.05 \\
                                                                        &                           & \multirow[c]{2}{*}{True}  & False                     & 0.10 \\
                                                                        &                           &                           & True                      & \textbf{0.18} \\
    \cmidrule{2-5}                                                                                                                                             
                                                                        & \multirow[c]{4}{*}{RGB}   & \multirow[c]{2}{*}{False} & False                     & -0.07 \\
                                                                        &                           &                           & True                      & \textbf{0.07} \\
                                                                        &                           & \multirow[c]{2}{*}{True}  & False                     & 0.06 \\
                                                                        &                           &                           & True                      & 0.04 \\
    \midrule                                                                                                                                                   
    \multirow[c]{8}{*}{GPT-4o}                                          & \multirow[c]{4}{*}{ASCII} & \multirow[c]{2}{*}{False} & False                     & 0.01 \\
                                                                        &                           &                           & True                      & 0.15 \\
                                                                        &                           & \multirow[c]{2}{*}{True}  & False                     & 0.11 \\
                                                                        &                           &                           & True                      & \textbf{0.20} \\
    \cmidrule{2-5}                                                                                                                                             
                                                                        & \multirow[c]{4}{*}{RGB}   & \multirow[c]{2}{*}{False} & False                     & 0.00 \\
                                                                        &                           &                           & True                      & 0.10 \\
                                                                        &                           & \multirow[c]{2}{*}{True}  & False                     & 0.15 \\
                                                                        &                           &                           & True                      & \textbf{0.17} \\
    \midrule                                                                                                                                                   
    \multirow[c]{4}{*}{o1-mini}                                         & \multirow[c]{4}{*}{ASCII} & \multirow[c]{2}{*}{False} & False                     & 0.25 \\
                                                                        &                           &                           & True                      & 0.26 \\
                                                                        &                           & \multirow[c]{2}{*}{True}  & False                     & 0.32 \\
                                                                        &                           &                           & True                      & \textbf{0.45} \\
    \midrule                                                                                                                                                   
    \multirow[c]{4}{*}{o1-preview}                                      & \multirow[c]{4}{*}{ASCII} & \multirow[c]{2}{*}{False} & False                     & 0.21 \\
                                                                        &                           &                           & True                      & 0.24 \\
                                                                        &                           & \multirow[c]{2}{*}{True}  & False                     & 0.35 \\
                                                                        &                           &                           & True                      & \textbf{0.36} \\
    \midrule                                                                                                                                                   
    \multirow[c]{8}{*}{o1}                                              & \multirow[c]{4}{*}{ASCII} & \multirow[c]{2}{*}{False} & False                     & 0.58 \\
                                                                        &                           &                           & True                      & 0.58 \\
                                                                        &                           & \multirow[c]{2}{*}{True}  & False                     & \textbf{0.83} \\
                                                                        &                           &                           & True                      & 0.66 \\
    \cmidrule{2-5}                                                                                                                                             
                                                                        & \multirow[c]{4}{*}{RGB}   & \multirow[c]{2}{*}{False} & False                     & 0.38 \\
                                                                        &                           &                           & True                      & 0.48 \\
                                                                        &                           & \multirow[c]{2}{*}{True}  & False                     & 0.52 \\
                                                                        &                           &                           & True                      & \textbf{0.60} \\
   \bottomrule                                                                                                                                                   
\end{tabular}
}
    \end{center}
    \label{tab:tic-tac-toe-ablation}
\end{table}

\subsubsection{Including Past Actions}
\label{app:additional-results:past-actions-ablation}

Over the course of our experimental investigation of frontier models' performance in dynamic environments, we noticed that these models have a tendency of getting stuck into repeating their previous action.
We therefore conduct an ablation where we do not show the past actions of the evaluation trajectory in the prompt (the past observations remain in the prompt, as do the observations and actions of the demonstration episode).
\cref{tab:past-actions-ablation-anthropic,tab:past-actions-ablation-gemini-flash,tab:past-actions-ablation-gemini-pro,tab:past-actions-ablation-gpt-4o,tab:past-actions-ablation-o1} show the performance for all frontier models on the grid world navigation task with $1$ expert demonstration episode with and without the past actions in the prompt.
In general, all models benefit from having the full history in the prompt (\ie both the previous observations and actions), even if that allows them to repeat their previous action more easily (in most environments they can still infer the previous action from the two past observations).
The only exception is \claude{} (see \cref{tab:past-actions-ablation-anthropic}), which does not show a clear trend in terms of including the past actions.
Given these results, we decide to always include the past actions in the prompt for all models and environments in all our other experiments.

\begin{table}
    \caption{
        Ablating whether to include the past actions of the evaluation trajectory in the prompt (the actions of the demonstration episodes are always included) for \claude{} on our grid world navigation task.
        Models are prone to repeating the previous action, so omitting it from the prompt could alleviate this problem.
        For ASCII observations, \claude{} is mostly indifferent to having the actions in the prompt.
        For coordinate observations, \claude{} performs better with the actions in the prompt, while for RGB observations, the past actions deteriorate performance.
    }
    \begin{center}
        \resizebox{0.65\textwidth}{!}{\begin{tabular}{@{}lllllr@{}}
    \toprule
    \textbf{Model}                                                      & \textbf{Observation}            & \textbf{Legal Actions}    & \textbf{Chain-of-Thought} & \textbf{Past Actions} & \textbf{Average Score} \\
    \midrule
    \multirow[c]{28}{*}{\shortstack{Claude 3.5 Sonnet\\\\(2024-10-22)}} & \multirow[c]{9}{*}{ASCII}       & \multirow[c]{4}{*}{False} & \multirow[c]{2}{*}{False} & False                 & 0.94 \\
                                                                        &                                 &                           &                           & \textbf{True}         & \textbf{1.00} \\
    \cmidrule{4-6}              
                                                                        &                                 &                           & \multirow[c]{2}{*}{True}  & \textbf{False}        & \textbf{1.00} \\
                                                                        &                                 &                           &                           & \textbf{True}         & \textbf{1.00} \\
    \cmidrule{3-6}              
                                                                        &                                 & \multirow[c]{4}{*}{True}  & \multirow[c]{2}{*}{False} & \textbf{False}        & \textbf{1.00} \\
                                                                        &                                 &                           &                           & \textbf{True}         & \textbf{1.00} \\
    \cmidrule{4-6}              
                                                                        &                                 &                           & \multirow[c]{2}{*}{True}  & \textbf{False}        & \textbf{1.00} \\
                                                                        &                                 &                           &                           & \textbf{True}         & \textbf{1.00} \\
    \cmidrule{2-6}              
                                                                        & \multirow[c]{9}{*}{Coordinates} & \multirow[c]{4}{*}{False} & \multirow[c]{2}{*}{False} & False                 & 0.77 \\
                                                                        &                                 &                           &                           & \textbf{True}         & \textbf{0.96} \\
    \cmidrule{4-6}              
                                                                        &                                 &                           & \multirow[c]{2}{*}{True}  & False                 & 0.84 \\
                                                                        &                                 &                           &                           & \textbf{True}         & \textbf{1.00} \\
    \cmidrule{3-6}              
                                                                        &                                 & \multirow[c]{4}{*}{True}  & \multirow[c]{2}{*}{False} & False                 & 0.77 \\
                                                                        &                                 &                           &                           & \textbf{True}         & \textbf{0.99} \\
    \cmidrule{4-6}              
                                                                        &                                 &                           & \multirow[c]{2}{*}{True}  & False                 & 0.78 \\
                                                                        &                                 &                           &                           & \textbf{True}         & \textbf{0.99} \\
    \cmidrule{2-6}              
                                                                        & \multirow[c]{9}{*}{RGB}         & \multirow[c]{4}{*}{False} & \multirow[c]{2}{*}{False} & \textbf{False}        & \textbf{0.26} \\
                                                                        &                                 &                           &                           & True                  & 0.19 \\
    \cmidrule{4-6}              
                                                                        &                                 &                           & \multirow[c]{2}{*}{True}  & \textbf{False}        & \textbf{0.47} \\
                                                                        &                                 &                           &                           & True                  & 0.40 \\
    \cmidrule{3-6}              
                                                                        &                                 & \multirow[c]{4}{*}{True}  & \multirow[c]{2}{*}{False} & \textbf{False}        & \textbf{0.37} \\
                                                                        &                                 &                           &                           & True                  & 0.21 \\
    \cmidrule{4-6}              
                                                                        &                                 &                           & \multirow[c]{2}{*}{True}  & \textbf{False}        & \textbf{0.68} \\
                                                                        &                                 &                           &                           & True                  & 0.53 \\
    \bottomrule
\end{tabular}
}
    \end{center}
    \label{tab:past-actions-ablation-anthropic}
\end{table}

\begin{table}
    \caption{
        Ablating whether to include the past actions of the evaluation trajectory in the prompt (the actions of the demonstration episodes are always included) for \flash{} on our grid world navigation task.
        Models are prone to repeating the previous action, so omitting it from the prompt could alleviate this problem.
        In this case, including the full history of the current trajectory, \ie both the observations and the actions, almost always improves the performance of \flash{}.
    }
    \begin{center}
        \resizebox{0.65\textwidth}{!}{\begin{tabular}{@{}lllllr@{}}
    \toprule
    \textbf{Model}                          & \textbf{Observation}              & \textbf{Legal Actions}    & \textbf{Chain-of-Thought} & \textbf{Past Actions} & \textbf{Average Score} \\
    \midrule
    \multirow[c]{28}{*}{Gemini 1.5 Flash}   & \multirow[c]{9}{*}{ASCII}         & \multirow[c]{4}{*}{False} & \multirow[c]{2}{*}{False} & False                 & 0.07 \\
                                            &                                   &                           &                           & \textbf{True}         & \textbf{0.09} \\
    \cmidrule{4-6}                  
                                            &                                   &                           & \multirow[c]{2}{*}{True}  & \textbf{False}        & \textbf{0.12} \\
                                            &                                   &                           &                           & True                  & 0.08 \\
    \cmidrule{3-6}                  
                                            &                                   & \multirow[c]{4}{*}{True}  & \multirow[c]{2}{*}{False} & False                 & 0.09 \\
                                            &                                   &                           &                           & \textbf{True}         & \textbf{0.11} \\
    \cmidrule{4-6}                  
                                            &                                   &                           & \multirow[c]{2}{*}{True}  & False                 & 0.10 \\
                                            &                                   &                           &                           & \textbf{True}         & \textbf{0.15} \\
    \cmidrule{2-6}              
                                            & \multirow[c]{9}{*}{Coordinates}   & \multirow[c]{4}{*}{False} & \multirow[c]{2}{*}{False} & False                 & 0.08 \\
                                            &                                   &                           &                           & \textbf{True}         & \textbf{0.21} \\
    \cmidrule{4-6}                  
                                            &                                   &                           & \multirow[c]{2}{*}{True}  & False                 & 0.07 \\
                                            &                                   &                           &                           & \textbf{True}         & \textbf{0.10} \\
    \cmidrule{3-6}                  
                                            &                                   & \multirow[c]{4}{*}{True}  & \multirow[c]{2}{*}{False} & False                 & 0.10 \\
                                            &                                   &                           &                           & \textbf{True}         & \textbf{0.14} \\
    \cmidrule{4-6}                  
                                            &                                   &                           & \multirow[c]{2}{*}{True}  & \textbf{False}        & \textbf{0.08} \\
                                            &                                   &                           &                           & \textbf{True}         & \textbf{0.08} \\
    \cmidrule{2-6}                  
                                            & \multirow[c]{9}{*}{RGB}           & \multirow[c]{4}{*}{False} & \multirow[c]{2}{*}{False} & False                 & 0.11 \\
                                            &                                   &                           &                           & \textbf{True}         & \textbf{0.16} \\
    \cmidrule{4-6}                  
                                            &                                   &                           & \multirow[c]{2}{*}{True}  & False                 & 0.08 \\
                                            &                                   &                           &                           & \textbf{True}         & \textbf{0.12} \\
    \cmidrule{3-6}                  
                                            &                                   & \multirow[c]{4}{*}{True}  & \multirow[c]{2}{*}{False} & False                 & 0.10 \\
                                            &                                   &                           &                           & \textbf{True}         & \textbf{0.14} \\
    \cmidrule{4-6}                  
                                            &                                   &                           & \multirow[c]{2}{*}{True}  & False                 & 0.09 \\
                                            &                                   &                           &                           & \textbf{True}         & \textbf{0.12} \\
   \bottomrule
\end{tabular}
}
    \end{center}
    \label{tab:past-actions-ablation-gemini-flash}
\end{table}

\begin{table}
    \caption{
        Ablating whether to include the past actions of the evaluation trajectory in the prompt (the actions of the demonstration episodes are always included) for \pro{} on our grid world navigation task.
        Models are prone to repeating the previous action, so omitting it from the prompt could alleviate this problem.
        Without exception, including the past actions in the prompt always improves the performance of \pro{}.
    }
    \begin{center}
        \resizebox{0.65\textwidth}{!}{\begin{tabular}{@{}lllllr@{}}
    \toprule
    \textbf{Model}                          & \textbf{Observation}              & \textbf{Legal Actions}    & \textbf{Chain-of-Thought} & \textbf{Past Actions} & \textbf{Average Score} \\
    \midrule
    \multirow[c]{28}{*}{Gemini 1.5 Pro}     & \multirow[c]{9}{*}{ASCII}         & \multirow[c]{4}{*}{False} & \multirow[c]{2}{*}{False} & False                 & 0.41 \\
                                            &                                   &                           &                           & \textbf{True}         & \textbf{0.51} \\
    \cmidrule{4-6}                  
                                            &                                   &                           & \multirow[c]{2}{*}{True}  & False                 & 0.22 \\
                                            &                                   &                           &                           & \textbf{True}         & \textbf{0.25} \\
    \cmidrule{3-6}                  
                                            &                                   & \multirow[c]{4}{*}{True}  & \multirow[c]{2}{*}{False} & False                 & 0.34 \\
                                            &                                   &                           &                           & \textbf{True}         & \textbf{0.51} \\
    \cmidrule{4-6}                  
                                            &                                   &                           & \multirow[c]{2}{*}{True}  & False                 & 0.30 \\
                                            &                                   &                           &                           & \textbf{True}         & \textbf{0.36} \\
    \cmidrule{2-6}                  
                                            & \multirow[c]{9}{*}{Coordinates}   & \multirow[c]{4}{*}{False} & \multirow[c]{2}{*}{False} & False                 & 0.07 \\
                                            &                                   &                           &                           & \textbf{True}         & \textbf{0.27} \\
    \cmidrule{4-6}                  
                                            &                                   &                           & \multirow[c]{2}{*}{True}  & False                 & 0.06 \\
                                            &                                   &                           &                           & \textbf{True}         & \textbf{0.13} \\
    \cmidrule{3-6}                  
                                            &                                   & \multirow[c]{4}{*}{True}  & \multirow[c]{2}{*}{False} & False                 & 0.02 \\
                                            &                                   &                           &                           & \textbf{True}         & \textbf{0.16} \\
    \cmidrule{4-6}                  
                                            &                                   &                           & \multirow[c]{2}{*}{True}  & False                 & 0.08 \\
                                            &                                   &                           &                           & \textbf{True}         & \textbf{0.12} \\
    \cmidrule{2-6}                  
                                            & \multirow[c]{9}{*}{RGB}           & \multirow[c]{4}{*}{False} & \multirow[c]{2}{*}{False} & False                 & 0.14 \\
                                            &                                   &                           &                           & \textbf{True}         & \textbf{0.17} \\
    \cmidrule{4-6}                  
                                            &                                   &                           & \multirow[c]{2}{*}{True}  & False                 & 0.14 \\
                                            &                                   &                           &                           & \textbf{True}         & \textbf{0.25} \\
    \cmidrule{3-6}                  
                                            &                                   & \multirow[c]{4}{*}{True}  & \multirow[c]{2}{*}{False} & False                 & 0.16 \\
                                            &                                   &                           &                           & \textbf{True}         & \textbf{0.24} \\
    \cmidrule{4-6}                  
                                            &                                   &                           & \multirow[c]{2}{*}{True}  & False                 & 0.15 \\
                                            &                                   &                           &                           & \textbf{True}         & \textbf{0.24} \\
    \bottomrule
\end{tabular}
}
    \end{center}
    \label{tab:past-actions-ablation-gemini-pro}
\end{table}

\begin{table}
    \caption{
        Ablating whether to include the past actions of the evaluation trajectory in the prompt (the actions of the demonstration episodes are always included) for \expflash{} on our grid world navigation task.
        Models are prone to repeating the previous action, so omitting it from the prompt could alleviate this problem.
        In this case, including the full history of the current trajectory, \ie both the observations and the actions, almost always improves the performance of \expflash{}.
    }
    \begin{center}
        \resizebox{0.65\textwidth}{!}{\begin{tabular}{@{}lllllr@{}}
    \toprule
    \textbf{Model}                              & \textbf{Observation}              & \textbf{Legal Actions}    & \textbf{Chain-of-Thought} & \textbf{Past Actions} & \textbf{Average Score} \\
    \midrule
    \multirow[c]{28}{*}{Gemini 2.0 Flash Exp.}  & \multirow[c]{9}{*}{ASCII}         & \multirow[c]{4}{*}{False} & \multirow[c]{2}{*}{False} & False                 & 0.19 \\
                                                &                                   &                           &                           & \textbf{True}         & \textbf{0.27} \\
    \cmidrule{4-6}                                                                                                                                                     
                                                &                                   &                           & \multirow[c]{2}{*}{True}  & False                 & 0.37 \\
                                                &                                   &                           &                           & \textbf{True}         & \textbf{0.41} \\
    \cmidrule{3-6}                                                                                                                                                     
                                                &                                   & \multirow[c]{4}{*}{True}  & \multirow[c]{2}{*}{False} & False                 & 0.17 \\
                                                &                                   &                           &                           & \textbf{True}         & \textbf{0.25} \\
    \cmidrule{4-6}                                                                                                                                                     
                                                &                                   &                           & \multirow[c]{2}{*}{True}  & False                 & 0.36 \\
                                                &                                   &                           &                           & \textbf{True}         & \textbf{0.39} \\
    \cmidrule{2-6}                                                                                                                                                     
                                                & \multirow[c]{9}{*}{Coordinates}   & \multirow[c]{4}{*}{False} & \multirow[c]{2}{*}{False} & False                 & 0.07 \\
                                                &                                   &                           &                           & \textbf{True}         & \textbf{0.22} \\
    \cmidrule{4-6}                                                                                                                                                     
                                                &                                   &                           & \multirow[c]{2}{*}{True}  & False                 & 0.08 \\
                                                &                                   &                           &                           & \textbf{True}         & \textbf{0.22} \\
    \cmidrule{3-6}                                                                                                                                                     
                                                &                                   & \multirow[c]{4}{*}{True}  & \multirow[c]{2}{*}{False} & False                 & 0.06 \\
                                                &                                   &                           &                           & \textbf{True}         & \textbf{0.30} \\
    \cmidrule{4-6}                                                                                                                                                     
                                                &                                   &                           & \multirow[c]{2}{*}{True}  & False                 & 0.05 \\
                                                &                                   &                           &                           & \textbf{True}         & \textbf{0.16} \\
    \cmidrule{2-6}                                                                                                                                                     
                                                & \multirow[c]{9}{*}{RGB}           & \multirow[c]{4}{*}{False} & \multirow[c]{2}{*}{False} & \textbf{False}        & \textbf{0.25} \\
                                                &                                   &                           &                           & True                  & 0.18 \\
    \cmidrule{4-6}                                                                                                                                                     
                                                &                                   &                           & \multirow[c]{2}{*}{True}  & False                 & 0.36 \\
                                                &                                   &                           &                           & \textbf{True}         & \textbf{0.37} \\
    \cmidrule{3-6}                                                                                                                                                     
                                                &                                   & \multirow[c]{4}{*}{True}  & \multirow[c]{2}{*}{False} & \textbf{False}        & \textbf{0.30} \\
                                                &                                   &                           &                           & True                  & 0.23 \\
    \cmidrule{4-6}                                                                                                                                                     
                                                &                                   &                           & \multirow[c]{2}{*}{True}  & False                 & 0.38 \\
                                                &                                   &                           &                           & \textbf{True}         & \textbf{0.43} \\
   \bottomrule
\end{tabular}
}
    \end{center}
    \label{tab:past-actions-ablation-gemini-flash-exp}
\end{table}

\begin{table}
    \caption{
        Ablating whether to include the past actions of the evaluation trajectory in the prompt (the actions of the demonstration episodes are always included) for \gpt{} on our grid world navigation task.
        Models are prone to repeating the previous action, so omitting it from the prompt could alleviate this problem.
        Including the past actions in the prompt always achieves the highest performance.
    }
    \begin{center}
        \resizebox{0.65\textwidth}{!}{\begin{tabular}{@{}lllllr@{}}
    \toprule
    \textbf{Model}              & \textbf{Observation}              & \textbf{Legal Actions}    & \textbf{Chain-of-Thought} & \textbf{Past Actions} & \textbf{Average Score} \\
    \midrule
    \multirow[c]{28}{*}{GPT-4o} & \multirow[c]{9}{*}{ASCII}         & \multirow[c]{4}{*}{False} & \multirow[c]{2}{*}{False} & False                 & 0.28 \\
                                &                                   &                           &                           & \textbf{True}         & \textbf{0.56} \\
    \cmidrule{4-6}                                  
                                &                                   &                           & \multirow[c]{2}{*}{True}  & False                 & 0.47 \\
                                &                                   &                           &                           & \textbf{True}         & \textbf{0.57} \\
    \cmidrule{3-6}                                  
                                &                                   & \multirow[c]{4}{*}{True}  & \multirow[c]{2}{*}{False} & False                 & 0.36 \\
                                &                                   &                           &                           & \textbf{True}         & \textbf{0.50} \\
    \cmidrule{4-6}                                  
                                &                                   &                           & \multirow[c]{2}{*}{True}  & False                 & 0.43 \\
                                &                                   &                           &                           & \textbf{True}         & \textbf{0.49} \\
    \cmidrule{2-6}                                  
                                & \multirow[c]{9}{*}{Coordinates}   & \multirow[c]{4}{*}{False} & \multirow[c]{2}{*}{False} & False                 & 0.28 \\
                                &                                   &                           &                           & \textbf{True}         & \textbf{0.69} \\
    \cmidrule{4-6}                                  
                                &                                   &                           & \multirow[c]{2}{*}{True}  & False                 & 0.46 \\
                                &                                   &                           &                           & \textbf{True}         & \textbf{0.73} \\
    \cmidrule{3-6}                                  
                                &                                   & \multirow[c]{4}{*}{True}  & \multirow[c]{2}{*}{False} & False                 & 0.21 \\
                                &                                   &                           &                           & \textbf{True}         & \textbf{0.76} \\
    \cmidrule{4-6}                                  
                                &                                   &                           & \multirow[c]{2}{*}{True}  & False                 & 0.48 \\
                                &                                   &                           &                           & \textbf{True}         & \textbf{0.72} \\
    \cmidrule{2-6}                                  
                                & \multirow[c]{9}{*}{RGB}           & \multirow[c]{4}{*}{False} & \multirow[c]{2}{*}{False} & False                 & 0.47 \\
                                &                                   &                           &                           & \textbf{True}         & \textbf{0.52} \\
    \cmidrule{4-6}                                  
                                &                                   &                           & \multirow[c]{2}{*}{True}  & False                 & 0.51 \\
                                &                                   &                           &                           & \textbf{True}         & \textbf{0.70} \\
    \cmidrule{3-6}                                  
                                &                                   & \multirow[c]{4}{*}{True}  & \multirow[c]{2}{*}{False} & False                 & 0.51 \\
                                &                                   &                           &                           & \textbf{True}         & \textbf{0.56} \\
    \cmidrule{4-6}                                  
                                &                                   &                           & \multirow[c]{2}{*}{True}  & \textbf{False}        & \textbf{0.61} \\
                                &                                   &                           &                           & \textbf{True}         & \textbf{0.61} \\
   \bottomrule
\end{tabular}
}
    \end{center}
    \label{tab:past-actions-ablation-gpt-4o}
\end{table}

\begin{table}
    \caption{
        Ablating whether to include the past actions of the evaluation trajectory in the prompt (the actions of the demonstration episodes are always included) for \mini{}, \preview{}, and \oone{} on our grid world navigation task.
        Models are prone to repeating the previous action, so omitting it from the prompt could alleviate this problem.
        In the vast majority of cases, including the past actions in the evaluation prompt results in the best performance for \mini{}, \preview{}, and \oone{}.
        Note that \mini{} and \preview{} are text-only models and, therefore, cannot process RGB images.
    }
    \begin{center}
        \resizebox{0.6\textwidth}{!}{\begin{tabular}{@{}lllllr@{}}
    \toprule
    \textbf{Model}                                      & \textbf{Observation}              & \textbf{Legal Actions}    & \textbf{Chain-of-Thought} & \textbf{Past Actions} & \textbf{Average Score} \\
    \midrule    
    \multirow[c]{18}{*}{\shortstack{\\\\\\o1-mini}}     & \multirow[c]{9}{*}{ASCII}         & \multirow[c]{4}{*}{False} & \multirow[c]{2}{*}{False} & False                 & 0.18 \\
                                                        &                                   &                           &                           & \textbf{True}         & \textbf{0.20} \\
    \cmidrule{4-6}                                      
                                                        &                                   &                           & \multirow[c]{2}{*}{True}  & \textbf{False}        & \textbf{0.22} \\
                                                        &                                   &                           &                           & True                  & 0.15 \\
    \cmidrule{3-6}                                      
                                                        &                                   & \multirow[c]{4}{*}{True}  & \multirow[c]{2}{*}{False} & \textbf{False}        & \textbf{0.24} \\
                                                        &                                   &                           &                           & True                  & 0.23 \\
    \cmidrule{4-6}                                      
                                                        &                                   &                           & \multirow[c]{2}{*}{True}  & False                 & 0.29 \\
                                                        &                                   &                           &                           & \textbf{True}         & \textbf{0.35} \\
    \cmidrule{2-6}                                      
                                                        & \multirow[c]{9}{*}{Coordinates}   & \multirow[c]{4}{*}{False} & \multirow[c]{2}{*}{False} & False                 & 0.41 \\
                                                        &                                   &                           &                           & \textbf{True}         & \textbf{0.68} \\
    \cmidrule{4-6}                                      
                                                        &                                   &                           & \multirow[c]{2}{*}{True}  & False                 & 0.40 \\
                                                        &                                   &                           &                           & \textbf{True}         & \textbf{0.72} \\
    \cmidrule{3-6}                                      
                                                        &                                   & \multirow[c]{4}{*}{True}  & \multirow[c]{2}{*}{False} & False                 & 0.44 \\
                                                        &                                   &                           &                           & \textbf{True}         & \textbf{0.77} \\
    \cmidrule{4-6}                                      
                                                        &                                   &                           & \multirow[c]{2}{*}{True}  & False                 & 0.44 \\
                                                        &                                   &                           &                           & \textbf{True}         & \textbf{0.71} \\
    \midrule
    \multirow[c]{18}{*}{\shortstack{\\\\\\o1-preview}}  & \multirow[c]{9}{*}{ASCII}         & \multirow[c]{4}{*}{False} & \multirow[c]{2}{*}{False} & False                 & 0.23 \\
                                                        &                                   &                           &                           & \textbf{True}         & \textbf{0.38} \\
    \cmidrule{4-6}                                      
                                                        &                                   &                           & \multirow[c]{2}{*}{True}  & False                 & 0.31 \\
                                                        &                                   &                           &                           & \textbf{True}         & \textbf{0.42} \\
    \cmidrule{3-6}                                      
                                                        &                                   & \multirow[c]{4}{*}{True}  & \multirow[c]{2}{*}{False} & False                 & 0.30 \\
                                                        &                                   &                           &                           & \textbf{True}         & \textbf{0.42} \\
    \cmidrule{4-6}                                      
                                                        &                                   &                           & \multirow[c]{2}{*}{True}  & False                 & 0.27 \\
                                                        &                                   &                           &                           & \textbf{True}         & \textbf{0.40} \\
    \cmidrule{2-6}                                      
                                                        & \multirow[c]{9}{*}{Coordinates}   & \multirow[c]{4}{*}{False} & \multirow[c]{2}{*}{False} & False                 & 0.10 \\
                                                        &                                   &                           &                           & \textbf{True}         & \textbf{0.13} \\
    \cmidrule{4-6}                                      
                                                        &                                   &                           & \multirow[c]{2}{*}{True}  & \textbf{False}        & \textbf{0.14} \\
                                                        &                                   &                           &                           & True                  & 0.13 \\
    \cmidrule{3-6}                                      
                                                        &                                   & \multirow[c]{4}{*}{True}  & \multirow[c]{2}{*}{False} & False                 & 0.11 \\
                                                        &                                   &                           &                           & \textbf{True}         & \textbf{0.16} \\
    \cmidrule{4-6}                                      
                                                        &                                   &                           & \multirow[c]{2}{*}{True}  & False                 & 0.10 \\
                                                        &                                   &                           &                           & \textbf{True}         & \textbf{0.13} \\
    \midrule    
    \multirow[c]{28}{*}{\shortstack{\\\\\\o1-mini}}     & \multirow[c]{9}{*}{ASCII}         & \multirow[c]{4}{*}{False} & \multirow[c]{2}{*}{False} & \textbf{False}        & \textbf{1.00} \\
                                                        &                                   &                           &                           & \textbf{True}         & \textbf{1.00} \\
    \cmidrule{4-6}                                                                                                                                                             
                                                        &                                   &                           & \multirow[c]{2}{*}{True}  & \textbf{False}        & \textbf{1.00} \\
                                                        &                                   &                           &                           & \textbf{True}         & \textbf{1.00} \\
    \cmidrule{3-6}                                                                                                                                                             
                                                        &                                   & \multirow[c]{4}{*}{True}  & \multirow[c]{2}{*}{False} & \textbf{False}        & \textbf{1.00} \\
                                                        &                                   &                           &                           & \textbf{True}         & \textbf{1.00} \\
    \cmidrule{4-6}                                                                                                                                                             
                                                        &                                   &                           & \multirow[c]{2}{*}{True}  & \textbf{False}        & \textbf{1.00} \\
                                                        &                                   &                           &                           & \textbf{True}         & \textbf{1.00} \\
    \cmidrule{2-6}                                                                                                                                                             
                                                        & \multirow[c]{9}{*}{Coordinates}   & \multirow[c]{4}{*}{False} & \multirow[c]{2}{*}{False} & False                 & 0.98 \\
                                                        &                                   &                           &                           & \textbf{True}         & \textbf{0.99} \\
    \cmidrule{4-6}                                                                                                                                                             
                                                        &                                   &                           & \multirow[c]{2}{*}{True}  & False                 & 0.98 \\
                                                        &                                   &                           &                           & \textbf{True}         & \textbf{1.00} \\
    \cmidrule{3-6}                                                                                                                                                             
                                                        &                                   & \multirow[c]{4}{*}{True}  & \multirow[c]{2}{*}{False} & False                 & 0.98 \\
                                                        &                                   &                           &                           & \textbf{True}         & \textbf{1.00} \\
    \cmidrule{4-6}                                                                                                                                                             
                                                        &                                   &                           & \multirow[c]{2}{*}{True}  & False                 & 0.99 \\
                                                        &                                   &                           &                           & \textbf{True}         & \textbf{1.00} \\
    \cmidrule{2-6}                                                                                                                                                             
                                                        & \multirow[c]{9}{*}{RGB}           & \multirow[c]{4}{*}{False} & \multirow[c]{2}{*}{False} & False                 & 0.60 \\
                                                        &                                   &                           &                           & \textbf{True}         & \textbf{0.81} \\
    \cmidrule{4-6}                                                                                                                                                             
                                                        &                                   &                           & \multirow[c]{2}{*}{True}  & False                 & 0.59 \\
                                                        &                                   &                           &                           & \textbf{True}         & \textbf{0.86} \\
    \cmidrule{3-6}                                                                                                                                                             
                                                        &                                   & \multirow[c]{4}{*}{True}  & \multirow[c]{2}{*}{False} & False                 & 0.65 \\
                                                        &                                   &                           &                           & \textbf{True}         & \textbf{0.84} \\
    \cmidrule{4-6}                                                                                                                                                             
                                                        &                                   &                           & \multirow[c]{2}{*}{True}  & False                 & 0.58 \\
                                                        &                                   &                           &                           & \textbf{True}         & \textbf{0.84} \\
    \bottomrule
\end{tabular}
}
    \end{center}
    \label{tab:past-actions-ablation-o1}
\end{table}

\end{document}